%% file: aaai2026.tex
\documentclass[letterpaper]{article} 
\usepackage{aaai2026}  
\usepackage{times}  
\usepackage{helvet}  
\usepackage{courier}  
\usepackage[hyphens]{url}  
\usepackage{graphicx} 
\usepackage{natbib}  
\usepackage{caption} 
\usepackage{algorithm}
\usepackage{algorithmic}

\usepackage{newfloat}
\usepackage{listings}
\DeclareCaptionStyle{ruled}{labelfont=normalfont,labelsep=colon,strut=off} 
\floatstyle{ruled}
\newfloat{listing}{tb}{lst}{}
\floatname{listing}{Listing}
\newcommand{\myfigref}[1]{Figure~\ref{#1}}
\newcommand{\tabref}[1]{Table~\ref{#1}}
\newcommand{\eqref}[1]{Eq.~\ref{#1}}

\title{HyperSHAP: Shapley Values and Interactions for Explaining\\ Hyperparameter Optimization}
\author {
    Marcel Wever\textsuperscript{\rm 1},
    Maximilian Muschalik\textsuperscript{\rm 2},
    Fabian Fumagalli\textsuperscript{\rm 2},
    Marius Lindauer\textsuperscript{\rm 1}
}
\affiliations {
    \textsuperscript{\rm 1}L3S Research Center, Leibniz University Hannover, Hannover, Germany\\
    \textsuperscript{\rm 2}MCML, LMU Munich, Munich, Germany\\
    m.wever@ai.uni-hannover.de, maximilian.muschalik@ifi.lmu.de, f.fumagalli@lmu.de, m.lindauer@ai.uni-hannover.de
}

\usepackage{bibentry}

\input{settings}
\input{project_commands}

\begin{document}

\maketitle

\begin{abstract}
Hyperparameter optimization (HPO) is a crucial step in achieving strong predictive performance.
Yet, the impact of individual hyperparameters on model generalization is highly context-dependent, prohibiting a one-size-fits-all solution and requiring opaque HPO methods to find optimal configurations.
However, the black-box nature of most HPO methods undermines user trust and discourages adoption. 
To address this, we propose a game-theoretic explainability framework for HPO based on Shapley values and interactions.
Our approach provides an additive decomposition of a performance measure across hyperparameters, enabling local and global explanations of hyperparameters' contributions and their interactions. 
The framework, named \tool, offers insights into \emph{ablation} studies, the \emph{tunability} of learning algorithms, and \emph{optimizer behavior} across different hyperparameter spaces.
We demonstrate \tool's capabilities on various HPO benchmarks to analyze the interaction structure of the corresponding HPO problems, demonstrating its broad applicability and actionable insights for improving HPO.
\end{abstract}

\begin{links}
     \link{Code \& Appx.}{https://github.com/automl/HyperSHAP}
\end{links}

\section{Introduction}

Hyperparameter optimization (HPO) is an important step in the design process of machine learning (ML) applications to achieve strong performance for a given dataset and performance measure \citep{snoek-icml14a,DBLP:journals/widm/BischlBLPRCTUBBDL23}. Especially, this is true for deep learning, where hyperparameters describe the architecture and steer the learning behavior \citep{zimmer-tpami21a}. Also, for generative AI and fine-tuning of foundation models, HPO is key for achieving the best results \citep{DBLP:conf/acl/YinCSJCL20,llmFinetuningHPO,DBLP:conf/automl/00010A23}.

Hyperparameters affect the generalization performance of models in varied ways, with some having a more significant impact on tuning than others~\citep{Bergstra-jair12a,fANOVA,zimmer-tpami21a}. The impact of hyperparameters on performance is highly context-dependent, varying with the dataset characteristics (e.g., size, noise level) and the specific performance measure being optimized (e.g., accuracy, F1)~\citep{Bergstra-jair12a,DBLP:conf/kdd/RijnH18}.

This complexity makes HPO particularly challenging, requiring opaque HPO methods to find optimal configurations within large search spaces \cite{feurer-nips15a}.
Yet, even with an optimized configuration, understanding \textit{why it outperforms others} remains difficult due to intricate effects and interactions among hyperparameters.

Despite their potential, HPO methods remain underused by domain experts, ML practitioners, and ML researchers
\cite{Lee2020,Hasebrook2023,Simon2023}. This limited adoption is partly due to their rigidity and poor adaptability to special cases, but also to a lack of interpretability \cite{Wang2019a,Drozdal2020}.
The latter is a key requirement among HPO users \cite{Wang2019a,Xin2021,Hasebrook2023,Sun2023}, and its absence has even led to a shift to manual tuning in high-stakes applications \cite{Xin2021}. For ML researchers, explanations are crucial to understand the contribution of individual components and retain control over model behavior. HPO researchers rely on such insights to analyze method performance and behavior.
Prior work on hyperparameter importance and effects \citep{fANOVA,DBLP:conf/nips/MoosbauerHCLB21,DBLP:conf/automl/SegelGTBL23,DBLP:conf/ijcai/WatanabeBH23,moo-fANOVA} highlights the need to close interpretability gaps to build trust and foster effective collaboration between HPO tools and ML practitioners~\citep{LindauerKKMT0HF24}. A complementary view is offered by \textit{tunability} \cite{tunability}, measuring performance gains over defaults to guide whether and what to tune. Yet, explanation methods tailored to tunability remain scarce.

\begin{figure*}[t]
    \begin{minipage}[c]{\linewidth}
    \centering
    \includegraphics[width=\textwidth]{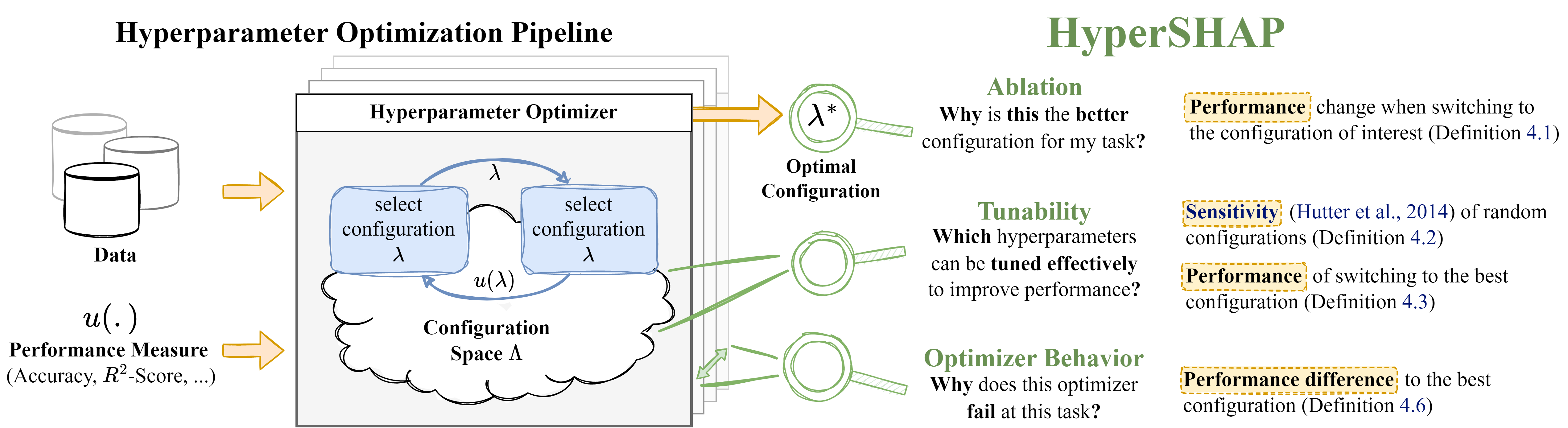}
    \end{minipage}
    \caption{Game-theoretic explanations as defined with \tool analyze hyperparameter values, hyperparameter spaces, and optimizers. \tool can be used for data-specific explanations or across datasets.}
    \label{fig_intro_illustration}
\end{figure*}

\paragraph{Contribution}

We formalize \tool, a novel post-hoc HPO-explanation framework:
\begin{enumerate}[itemsep=0em,topsep=-0.25em] 
    \item[\textbf{(1)}] We propose \tool as a collection of \numgames explanation games and interpret them using Shapley values and interactions for specific configurations, 
    hyperparameter spaces, 
    and optimizers.
    \item[\textbf{(2)}] We showcase how \tool can be employed for tackling various explanation tasks.
    \item[\textbf{(3)}] Comparing to fANOVA, we find that \tool's explanations are more actionable to select subsets of hyperparameters for tuning.
    \item[\textbf{(4)}] We provide a publicly available reference implementation of \tool via GitHub.
\end{enumerate}

\section{Related Work}\label{sec:related-work}

Hyperparameter importance (HPI) has gained significant attention in machine learning due to its crucial role in justifying the need for HPO \citep{DBLP:conf/gecco/PushakH20a,DBLP:journals/telo/PushakH22,DBLP:conf/ppsn/SchneiderSPBTK22}, whereas tunability quantifies how much certain hyperparameters can be tuned for specific tasks~\citep{tunability}.
A variety of approaches have been developed to assess how different hyperparameters affect the performance of resulting models, ranging from simple (surrogate-based) ablations \citep{Fawcett-jh16a,DBLP:conf/aaai/BiedenkappLEHFH17} to sensitivity analyses and eliciting interactions between hyperparameters based on fANOVA \citep{fANOVA,DBLP:conf/kdd/RijnH18,elShawiTuneOrNotTune,DBLP:conf/ijcai/WatanabeBH23}. 
In this work, we propose a novel approach to quantifying HPI using Shapley values, with a particular focus on capturing interactions between hyperparameters through Shapley interaction indices. We focus on quantifying interactions since prior works \citep{zimmer-tpami21a,DBLP:journals/telo/PushakH22,DBLP:journals/jscic/NovelloPLC23} noticed that interaction is occasionally comparably low, which could serve as a foundation for a new generation of HPO methods that do not assume interactions to be omnipresent.

Beyond quantifying HPI, to better understand the impact of hyperparameters and the tuning behavior of hyperparameter optimizers, other approaches have been proposed, such as algorithm footprints \citep{DBLP:conf/cec/Smith-MilesT12}, partial dependence plots for hyperparameter effects~\cite{DBLP:conf/nips/MoosbauerHCLB21} or deriving symbolic explanations \citep{DBLP:conf/automl/SegelGTBL23}, providing an interpretable model for estimating the performance of a learner from its hyperparameters.

\section{Hyperparameter Optimization}
Hyperparameter optimization (HPO) is concerned with the problem of finding the most suitable \textit{hyperparameter configuration} of a learner for a given task, typically consisting of some labeled dataset $D$ and some performance measure $\perf$ quantifying the usefulness \citep{DBLP:journals/widm/BischlBLPRCTUBBDL23}.
To put it formally, let $\cX$ be an instance space and $\cY$ a label space and suppose $x\in \cX$ are (non-deterministically) associated with labels $y \in \cY$ via a joint probability distribution $\mathbb{P}$.

Then, a dataset $D=\{ (x^{(k)}, y^{(k)}) \}_{k=1}^N \subset \cX \times \cY$ is a sample from that probability distribution.
Furthermore, a predictive performance measure $\perf: \cY \times P(\cY) \rightarrow \R$ is a function mapping tuples consisting of a label and a probability distribution over the label space to the reals.
Given a configuration $\conf \in \confs$, a learner parameterized with $\conf$ maps datasets~$D$ from the dataset space $\mathbb{D}$ to a corresponding hypothesis $h_{\conf, D} \in \mathcal{H} := \{ h \mid h: \cX \rightarrow P(\cY) \}$.

As a configuration $\conf\in\confs$ typically affects the hypothesis space $\mathcal{H}$ and the learning behavior, it needs to be tuned to the given dataset and performance measure. The task of HPO is then to find a configuration yielding a hypothesis that generalizes well beyond the data used for training.
For a dataset $D\in\mathbb{D}$, the following optimization problem needs to be solved:
$
\conf^\ast \in \underset{\conf\in\confs}{\arg\max} \int_{(x, y)\sim \mathbb{P}} \perf \big(y, h_{\conf, D}(x)\big).
$
As the true generalization performance is intractable, it is estimated by splitting the given dataset $D$ into training $D_{T}$ and validation data $D_{V}$. Accordingly, we obtain $\conf^\ast \in \underset{\conf\in\confs}{\arg\max}\, \val(\conf, D)$, with
\[
\val(\conf, D) := 
\mathbb{E}_{(D_{T}, D_{V}) \sim D} \sum_{(x, y) \in D_{V}} \frac{\perf \big(y, h_{\conf, D_{T}}(x)\big)}{|D_V|} .
\]

Na{\"i}vely, HPO can be approached by discretizing the domains of hyperparameters and conducting a grid search or by a random search \citep{Bergstra-jair12a}. More commonly, state-of-the-art methods often leverage Bayesian optimization and multi-fidelity optimization for higher efficiency and effectiveness \citep{DBLP:journals/widm/BischlBLPRCTUBBDL23}.

\section{Explainable AI and Game Theory}\label{sec:background-shapley}

Within the field of eXplainable AI (XAI), cooperative game theory has been widely applied to assign contributions to entities, such as features or data points for a given task \citep{DBLP:conf/ijcai/RozemberczkiWBY22}.
Most prominently, it is used to interpret predictions of black-box models using feature attributions \citep{Lundberg.2017} and the \gls*{SV} \citep{Shapley.1953}.
\glspl*{SI} \citep{Grabisch.1999} extend the \gls*{SV} by additionally assigning contributions to groups of entities, which reveal \emph{synergies and redundancies}.

Such feature interactions reveal additive structures essential for understanding complex predictions \citep{Sundararajan.2020}.
Explanations consist of two components \citep{Fumagalli.2024a}: (1) an \emph{explanation game} $\nu: 2^{\mathcal{N}} \to \mathbb{R}$, a set function over feature subsets of the $n$ features of interest indexed by $\mathcal N = \{1,\dots,n\}$ that evaluates properties such as prediction or performance; (2) \emph{interpretable} main and interaction effects derived from the \gls*{SV} and \glspl*{SI}.
Analogously, the next section defines explanation games over hyperparameter ablations in $\val$ , using the \gls*{SV} and \glspl*{SI} to quantify tunability.

\paragraph{Explanation Games via Feature Imputations.}
Given the prediction of a black box model $f: \mathbb{R}^n \to \mathbb{R}$ and an instance $\mathbf{x} \in \mathbb{R}^n$, \emph{baseline imputation} with $b \in \mathbb{R}^n$ for a coalition $S \subseteq \mathcal N$ is given by $\oplus_S: \mathbb{R}^n \times \mathbb{R}^n \to \mathbb{R}^n$ as
\begin{align*}
    \nu_{\mathbf{x}}^{(b)}(S) := f(\mathbf{x} \oplus_S \mathbf{b}) \text{ with } \mathbf{x} \oplus_S \mathbf{b} := \begin{cases}
        x_i, &\text{ if } i \in S \, \, , \\
        b_i, &\text{ if } i \notin S \, \, .
    \end{cases}
\end{align*}

Baseline imputation is highly sensitive to the chosen baseline \citep{sturmfels2020visualizing}. \emph{Marginal} and \emph{conditional} imputation extend this by averaging over randomized baselines \cite{Sundararajan.2020b}: 
$\nu_{\mathbf{x}}^{(p)}(S) := \mathbb{E}_{\mathbf{b} \sim p(\mathbf{b})}[f(\mathbf{x} \oplus_S \mathbf{b})],$
where $p(\mathbf{b})$ is the marginal or conditional feature distribution.
The imputed predictions define \emph{local} games for explaining individual predictions, while \emph{global} games capture aggregate properties, e.g., variance or performance.
As such, explanations increasingly reflect the underlying distribution $p$ \citep{Fumagalli.2024a}.

\paragraph{Shapley Value (SV) and Shapley Interaction (SI).}
An explanation game is additively decomposed by the \glspl*{MI} $m:2^{\mathcal N} \to \mathbb{R}$ \citep{Muschalik.2024}, i.e. the Möbius transform \citep{rota1964foundations}, for $T \subseteq \mathcal N$ as

\begin{align*}
     \nu(T) = \sum_{S \subseteq T} m(S) \text{ with } m(S) := \sum_{L \subseteq S} (-1)^{\vert S \vert - \vert L \vert} \nu(L) \, \, .
\end{align*}
The \glspl*{MI} capture \emph{pure main and interaction effects} but contain $2^n$ non-trivial components, too many for practical interpretation in ML applications \citep{Muschalik.2024}.
To reduce this complexity, the \gls*{SV} and \glspl*{SI} summarize the \glspl*{MI} into \emph{interpretable} effects.
The \gls*{SV} assigns contributions to individuals, is uniquely characterized, and satisfies four axioms: linearity, symmetry, dummy, and efficiency.
The \gls*{SV} summarizes the \glspl*{MI} distributing each \gls*{MI} among the involved players via $\phi^{\text{SV}}(i) = \sum_{S \subseteq \mathcal{N}: i \in S} \frac{1}{\vert S \vert} m(S)$ for all $i \in \mathcal{N}$.
Yet, the \gls*{SV} does not uncover interactions.
Given an \emph{explanation order} $k \in \{1,\dots,n\}$, the \glspl*{SI} $\Phi_k$ extend the \gls*{SV} to assign contributions to subsets of players up to size $k$.
For $k=1$ the \glspl*{SI} yield the \gls*{SV} and the \glspl*{MI} for $k=n$.
Various forms of \glspl*{SI} exist, where positive values indicate synergy and negative values signal redundancy among the involved features.
For instance, the \gls*{FSII} \citep{Tsai.2022} defines the best $k$-additive approximation $\hat{\nu}_k(S) := \sum_{L \subseteq S:\vert L\vert \leq k}{\Phi_k(L)}$ weighted by the Shapley kernel, enabling quantification of interaction strength.
\glspl{SI} thus offer a flexible trade-off between expressivity and complexity, a framework we now adapt to HPO.

\section{Explaining Hyperparameter Optimization}\label{sec:hpi-games}

Explanations in HPO are needed at multiple levels, from individual configurations to qualitative comparisons of HPO tools. 
Here, we consider four areas, dubbed \ablation, \sensitivity, \dstunability, and \opthabit.
We begin with \ablation as the foundation of \tool, extend it to \sensitivity (showing links to fANOVA by~\citet{fANOVA}), and compare it theoretically to \dstunability.
\dstunability then serves to uncover \opthabit. We conclude with practical considerations of \tool.
Let $\mathcal N$ denote the set of hyperparameters; we quantify main and interaction effects based on the \gls*{SV} and \glspl*{SI} of the explanation games.
Proofs are deferred to the appendix.

\subsection{Ablation of Hyperparameter Configurations}\label{sec:ablation}

One common approach to explaining HPO results is to compare a configuration of interest, $\conf^\ast$, to a reference configuration $\conf^0$, typically a library default or a tuned default that has performed well on prior tasks. The configuration $\conf^\ast$ may stem from HPO or be manually selected. The key question is how changes in $\conf^\ast$ impact performance relative to $\conf^0$. To investigate this, we can incrementally modify $\conf^0$ by replacing its hyperparameter values with those from $\conf^\ast$, one at a time; a process known as \emph{ablation}, widely used in empirical ML research \cite{cohen-aimag1988,rendsburg-icml2020,herrmann-icml2024}.

HPO-ablation studies were proposed by \citet{Fawcett-jh16a} but limited to sequential single-hyperparameter ablation paths, ignoring interactions.
Instead, we form an explanation game for ablation using \emph{all possible subsets}, which allows us to capture interactions.

\begin{definition}[\ablation Game]
The \ablation explanation game $\nu_{G_A}: 2^{\mathcal N} \to \mathbb{R}$ is defined as a tuple $G_A := (\conf^0, \conf^\ast, D, \perf),$
consisting of a \emph{baseline (default)} configuration $\conf^0$,
a target configuration $\conf^\ast$, a dataset $D$, and a measure $u$.
Given a coalition $S\subseteq \mathcal{N}$, we construct an intermediate configuration with $\oplus_S: \confs \times \confs \to \confs$ as
\[
\conf^* \oplus_S \conf^0 := \begin{cases}
\lambda^\ast_i, & \text{ if } i \in S,\\
\lambda^0_i, & \text{ else},
\end{cases}
\]
and evaluate its value via
$\nu_{G_A}(S) := \val(\conf^\ast \oplus_S \conf^0, D) \,\, .$
\end{definition}

The \ablation game quantifies the worth of a coalition based on the comparison with a baseline configuration $\conf^0$.
In XAI terminology, this approach is known as \emph{baseline imputation}. 
Natural extensions of the \ablation game capture these ablations with respect to a distribution $\conf^0 \sim p^0(\conf^0)$ over configuration space $\confs$ as
$\mathbb{E}_{\conf^0 \sim p^0(\conf^0)}[ \val(\conf^\ast \oplus_S \conf^0, D)]
,$
which relates to the \emph{marginal performance}~\cite{fANOVA}.
In XAI terminology, it is further distinguished between distributions $p(\conf^0)$ that either depend (conditional) or do not depend (marginal) on the target configuration $\conf^\ast$.
Baseline imputation is often chosen for efficiency and is also argued to have desirable properties \citep{Sundararajan.2020b}.
Still, the choice of baseline strongly influences the explanation \citep{sturmfels2020visualizing}.
We typically use a \emph{default} configuration~\cite{AnastacioH20} here, though our methodology readily extends to probabilistic baselines.

\subsection{Sensitivity and Tunability of Learners}\label{sec:sensitivity-tunability}
Zooming out from a specific configuration, we can ask to what extent it is worthwhile to tune hyperparameters. 
In the literature, this question has been connected to the term of \textit{tunability} \citep{tunability}. 
\dstunability aims to quantify how much performance improvements can be obtained by tuning a learner, comparing against a baseline configuration, e.g., a configuration that is known to work well across various datasets~\citep{DBLP:conf/gecco/PushakH20a}.
In this context, we are interested in the importance of tuning specific hyperparameters.
A classical tool to quantify variable importance is \emph{sensitivity analysis} \citep{Owen_2013}, measuring the variance induced by the variables and decomposing their contributions into main and interaction effects.

\begin{definition}[\sensitivity Game]
The \sensitivity game $\nu_{G_V}: 2^{\mathcal N} \to \mathbb{R}$ is defined as a tuple $G_V := (\conf^0, \confs, p^*, D, \perf),$
consisting of a \emph{baseline} configuration $\conf^0$, a configuration space of interest $\confs$ equipped with a probability distribution $p^*$, a dataset $D$, and a performance measure $u$. 
The value function is given by
\[
    \nu_{G_V}(S) := \mathbb{V}_{\conf \sim p^\ast(\conf)}[\val(\conf \oplus_S \conf^0, D)] \,\, .
\] 
\end{definition}

A large value of a coalition $S \subseteq \mathcal N$ in the \sensitivity game indicates that these hyperparameters are important to be set to the right value.
\citet{fANOVA} implicitly rely on the \sensitivity game and compute the fANOVA decomposition, quantifying \emph{pure} main and interaction effects.
In game theory, this corresponds to the \glspl*{MI} of the \sensitivity game, which can be summarized using the \gls*{SV} and \glspl*{SI} \citep{Fumagalli.2024a}.

While sensitivity analysis is a suitable tool in XAI, it has some drawbacks for measuring tunability \cite{tunability}.
First, as illustrated below, the total variance being decomposed $\nu_{G_V}(\mathcal N)$ highly depends on the chosen probability distribution $p^*$ and the configuration space $\confs$.
Moreover, it does not reflect the performance increase expected when tuning all hyperparameters, but variations (in any direction).
Second, for a coalition of hyperparameters $S \subseteq \mathcal N$, we expect that the coalition's worth (performance) increases when tuning additional hyperparameters, i.e., $\nu(S) \leq  \nu(T)$, if $S \subseteq T$.
This property is known as \emph{monotonicity} \citep{Fujimoto.2006}, but \emph{does not} hold in general for the \sensitivity game $\nu_{G_V}$.
For a simple example, we refer to the appendix.
Based on \citet{tunability}, we define an explanation game for tunability that exhibits monotonicity:

\begin{definition}[\dstunability Game]\label{def:tunability}
The \dstunability game is defined by a tuple $G_T = (\conf^0, \confs, D, \perf),$
consisting of a baseline configuration $\conf^0 \in \confs$, a configuration space $\confs$, a dataset $D$, and a measure $u$. The value function is given by
\[
\nu_{G_T}(S) := \max_{\conf \in \confs}
\,\val(\conf \oplus_S \conf^0, D) \,\, .
\]
\end{definition}

The \dstunability game directly measures the performance obtained from tuning the hyperparameters of a coalition $S$ while leaving the remaining hyperparameters at the default value $\conf^0$.
The \dstunability game is monotone, which yields the following lemma.

\begin{proposition}
    The \dstunability game yields non-negative \glspl*{SV} and non-negative \emph{pure} individual (main) effects obtained from functional ANOVA via the \glspl*{MI}.
\end{proposition}

While the \emph{main effects} obtained from the \dstunability game are non-negative, interactions can still be negative, indicating redundancies of the involved hyperparameters.

\paragraph{Comparing Tunability vs. Sensitivity.}
We now showcase the different results of the \dstunability game vs. the \sensitivity game using an educational example.
We consider a two-dimensional configuration space $\confs := \Lambda_1 \times \Lambda_2$ with discrete configurations $\Lambda_1 := \{0,1\}$ and $\Lambda_2:= \{0,\dots,m\}$ for $m > 1$.
The optimal configuration is defined as $\conf^\ast := (1,m)$, and the performance is quantified by $\val(\conf,D) := \mathbf{1}_{\lambda_1=\lambda_1^\ast}+\mathbf{1}_{\lambda_2 = \lambda^\ast_2}$, where $\mathbf{1}$ is the indicator function.
That is, we observe an increase of performance of $1$ for each of the hyperparameters set to the optimal configuration $\conf^\ast$.
Lastly, we set the configuration baseline to $\conf^0 := (0,0)$ or $\conf^0 := \conf^\ast$.
Intuitively, we expect that both hyperparameters obtain similar importance scores, since they both contribute equally to the optimal performance $\val(\conf^\ast, D) = 2$.
Moreover, if the baseline is set to the optimal configuration $\conf^\ast$, we expect the score to reflect that there is no benefit of tuning.
Since the hyperparameters affect the performance independently, we do not expect any interactions.

\begin{table}[t]
    \centering
    \caption{Importance scores for a 2D HPO problem under the \sensitivity and \dstunability games, with baseline set to $(0,0)$ and optimum $\conf^\ast$. \sensitivity assigns lower scores to hyperparameters with larger domains ($\lambda_2)$. Setting $\conf^0 = \conf^*$ reduces the \dstunability scores to 0; \sensitivity is unaffected.}
    \label{tab:synthetic_example}
    \begin{tabular}{cc|cccc}
    \toprule
     \multicolumn{2}{c|}{\textbf{Game}} & \multicolumn{2}{c}{\textbf{Sensitivity}}  & \multicolumn{2}{c}{\textbf{\dstunability}} \\ 
     \multicolumn{2}{c|}{$\conf^0$} & $(0,0)$ & $\conf^\ast$   & $(0,0)$ & $\conf^\ast$ \\ \midrule
      \multirow{3}{*}{\rotatebox{90}{\textbf{Score}}} & $\lambda_1$ & $1/4$ & $1/4$  & 1 & 0 \\
      &$\lambda_2$  & $\frac{m}{(m+1)^2}$ & $\frac{m}{(m+1)^2}$  & 1 & 0 \\
      &$\lambda_1 \times \lambda_2$ & 0 & 0 & 0 & 0 \\\bottomrule
    \end{tabular}
\end{table}

The HPI scores of the \sensitivity and \dstunability game for the example are given by \tabref{tab:synthetic_example}.
Both approaches, \sensitivity and \dstunability, correctly quantify the absence of interaction $\lambda_1 \times \lambda_2$.
As opposed to the \dstunability game, the \sensitivity game assigns smaller scores to the hyperparameter $\lambda_2$ due to the larger domain $\Lambda_2$. 
In fact, the \sensitivity score of $\lambda_2$ roughly decreases with order $m^{-1}$.
Moreover, the \dstunability scores reflect the performance increase and, as expected, distribute the difference between the optimal and the baseline performance properly among the hyperparameters.
In contrast, the \sensitivity scores decompose the overall variance, which depends on $\confs$ and $p^*$.
Lastly, setting the baseline configuration $\conf^0$ to $\conf^\ast$ decreases the \dstunability scores to zero, whereas the \sensitivity scores remain unaffected.
In summary, \sensitivity reflects the variability 
in performance when changing the hyperparameter values, whereas \dstunability reflects the benefit of tuning over the baseline.

\subsection{Optimizer Bias}\label{sec:opthabit}
The \dstunability game aims to explain the importance of hyperparameters being tuned, which can also be used to gain insights into the capabilities of a hyperparameter optimizer.
In particular, by comparing the optimal performance with the empirical performance of a single optimizer, we can uncover biases and pinpoint specific hyperparameters that the optimizer of interest fails to exploit.
We define a hyperparameter optimizer as a function $\mathcal{O}: \mathbb{D} \times 2^{\confs} \rightarrow \confs$, mapping from the space of datasets and a configuration space to a specific configuration. 

\begin{definition}[\opthabit Game] \label{def:opthabit}
    The \opthabit HPI game is defined as a tuple $G_O = (\confs, \conf^0, \mathcal{O}, D, u),$ consisting of a configuration space $\confs$, a baseline $\conf^0$, the hyperparameter optimizer of interest $\mathcal O$, a dataset $D$ and a measure $u$.
    For $S \subseteq \mathcal N$, we define $\confs^S := \{ \conf \oplus_S \conf^0: \conf \in \confs\}$ and
        \begin{align*}
    \nu_{G_0}(S) := \val \Big(\mathcal{O}(D, \confs^S), D \Big)
    - \nu_{G_T}(S) \,\, .
    \end{align*}
\end{definition}

Intuitively, the value function captures how much performance is lost relative to the best known configuration. 
In other words, with the help of Definition~\ref{def:opthabit}, we can pinpoint where the hyperparameter optimizer $\mathcal{O}$ falls short, revealing, for example, whether it struggles to optimize certain hyperparameters or types thereof. The analysis can be conducted via inexpensive surrogate-based HPO benchmarks.

\subsection{Practical Aspects of \tool}\label{sec:hypershap-in-practice}
This section addresses practical aspects of \tool to efficiently approximate the proposed games and generalize them to multiple datasets.

\paragraph{Efficient Approximation.}
Na{\"i}vely, to evaluate a single coalition in Definition~\ref{def:tunability} of the Tunability game, we need to conduct one HPO run.
While this can be costly, we argue that using surrogate models that are, e.g., obtained through Bayesian optimization, can be used to calculate the maximum efficiently. Surrogate models are commonly used in explainability methods for HPO, including fANOVA and related approaches \cite{fANOVA,DBLP:conf/aaai/BiedenkappLEHFH17,DBLP:conf/nips/MoosbauerHCLB21,DBLP:conf/automl/SegelGTBL23}.
For \tool, we can bound the approximation error for the explanations as follows:
\begin{theorem}
For a surrogate model with approximation error $\epsilon$, the approximation error of Shapley values and interactions in \tool is bounded by $2\epsilon$.
\end{theorem}
\paragraph{\opthabit Analysis} To analyze \opthabit, we propose to approximate $\nu_{G_T}$ using a diverse ensemble of optimizers $\mathbb{O}:= \{ \mathcal{O}_i\}$, and choose the best result for $\confs^S$ obtained through any optimizer from $\mathbb{O}$, forming a virtual optimizer, always returning the best-known value.
This virtual best hyperparameter optimizer approximates
\[
\nu_{G_T}(S) \approx \underset{\conf^i = \mathcal{O}_i(D, \confs^S)}{\max} \val(\conf^i, D) \,\, .
\]

\paragraph{Worst Case Analysis.}
In order to identify hyperparameters that should not be mistuned, we can conduct a worst-case analysis with \tool by replacing the $\max$ by a $\min$ operator in Definition~\ref{def:tunability}.

\paragraph{Game Extensions Across Multiple Datasets.} 
In a more general setting, we are interested in explanations across multiple datasets, for which we can extend the previous games naturally as follows:

\begin{definition}[Multi-Dataset Games]
Given a collection of datasets $\mathcal D := \{D_1, \dots, D_M\}$, the corresponding games $\nu^{D_i}_{G}$ for $1 \leq i \leq M$ with $G \in \{G_A,G_V,G_T,G_O\}$, we define its multi-dataset variant with the value function
$\nu^{\mathcal D}_{G}(S) := \bigoplus_{i = 1 }^M \nu^{D_i}_G(S)$,
where $\bigoplus$ denotes an aggregation operator, e.g., the mean or a quantile of the game values obtained for the datasets $D_i$.
\end{definition}

Considering explanations across datasets enables a broader view of the impact of how individual hyperparameters and their interactions affect generalization performance. Aggregating coalition values reveals which hyperparameters are generally worth tuning, rather than just data-specific importance, justifying tuning recommendations or uncovering systematic optimizer biases beyond data-specific effects.

\section{Experiments}\label{sec_experiments}

We evaluate the applicability of \tool across various explanation tasks and benchmarks.
To this end, we rely on four HPO benchmarks: \lcbench \citep{zimmer-tpami21a}, \rbvranger \citep{DBLP:conf/automl/PfistererSMBB22}, \pdone \citep{wang-jmlr24}, and \jahs \citep{bansal-neurips22a}. The implementation is based on \texttt{shapiq} \cite{Muschalik.2024} and (will be) publicly available on GitHub\footnotemark[1]. We provide details regarding the setup, interpreting plots, and more results in the appendix. Generally, positive interactions are colored in red and negative in blue.

\subsection{Insights from Ablation and Tunability}\label{sec_experiments_ablation_tunability}

\begin{figure*}[t]
    \centering
    \begin{minipage}[c]{0.6\linewidth}
    \includegraphics[width=0.92\textwidth]{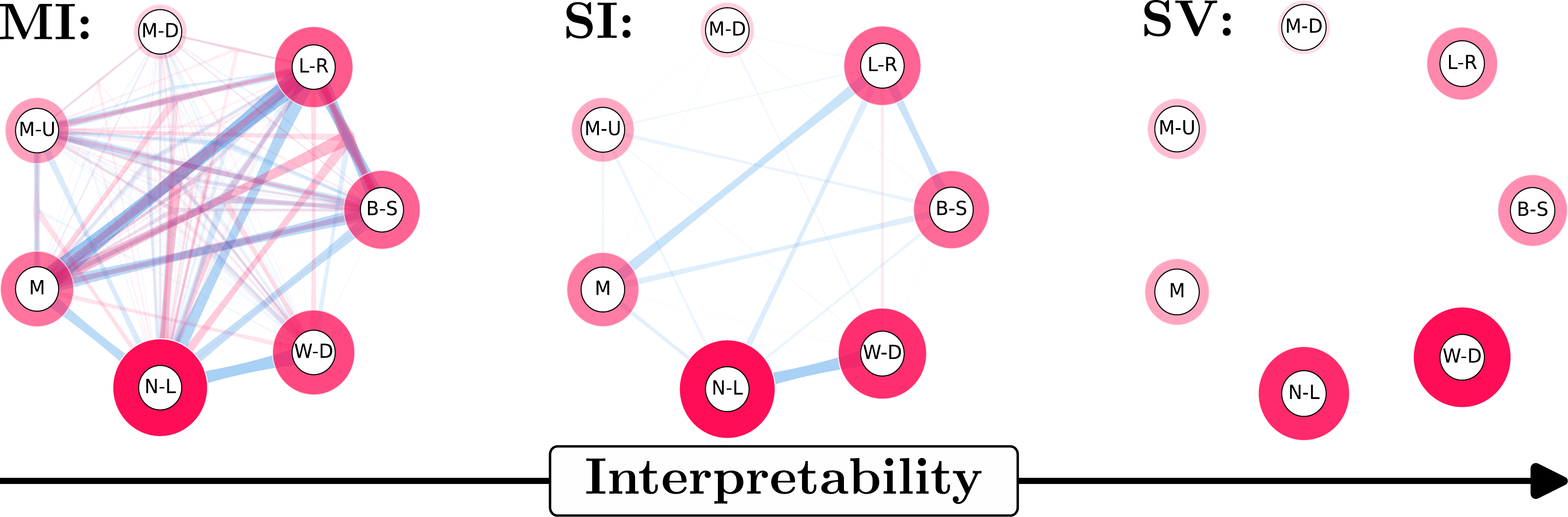}
    \end{minipage}
    \hfill
    \begin{minipage}[c]{0.39\linewidth}
    \includegraphics[width=\linewidth]{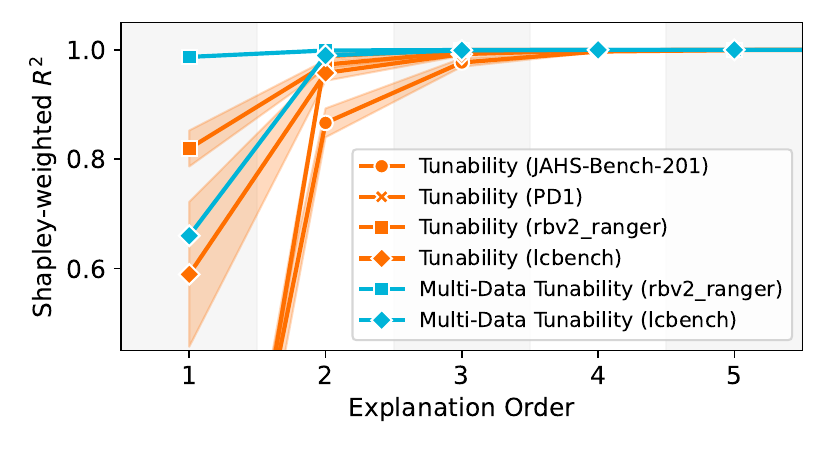}
    \end{minipage}
    \caption{\textbf{Left:} Interaction graphs showing Möbius interactions (MI), second-order Shapley interactions (SI), and Shapley values (SV) where MIs terms are aggregated for interoperability. \textbf{Right:} Faithfulness of lower-order explanations approximating higher-order effects \cite{Muschalik.2024}. An explanation order of 3 already approximates the full game ($R^2 \approx 1$) well.}
    \label{fig_interaction_quantification}
\end{figure*}

\begin{figure}[t]
    \centering
    \begin{minipage}[c]{0.45\columnwidth}
        \includegraphics[width=\textwidth,trim={0 1.1cm 0 0.4cm},clip]{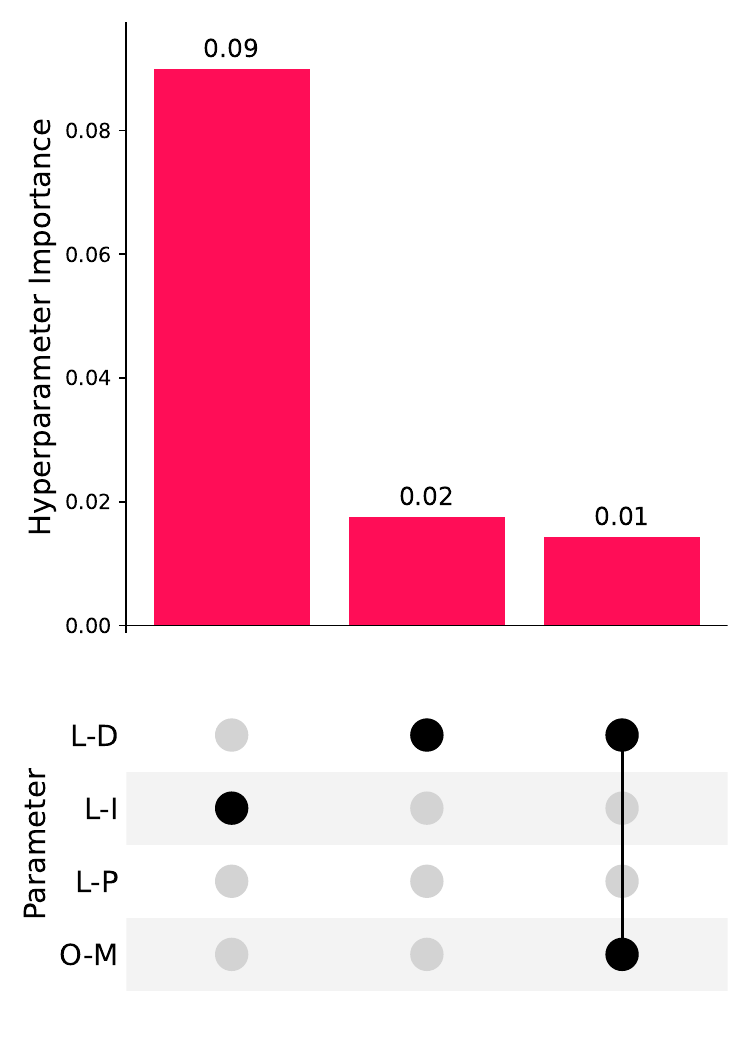}
    \end{minipage}
    \hfill
    \begin{minipage}[c]{0.45\columnwidth}
        \includegraphics[width=\textwidth,trim={0 1.1cm 0 0.4cm},clip]{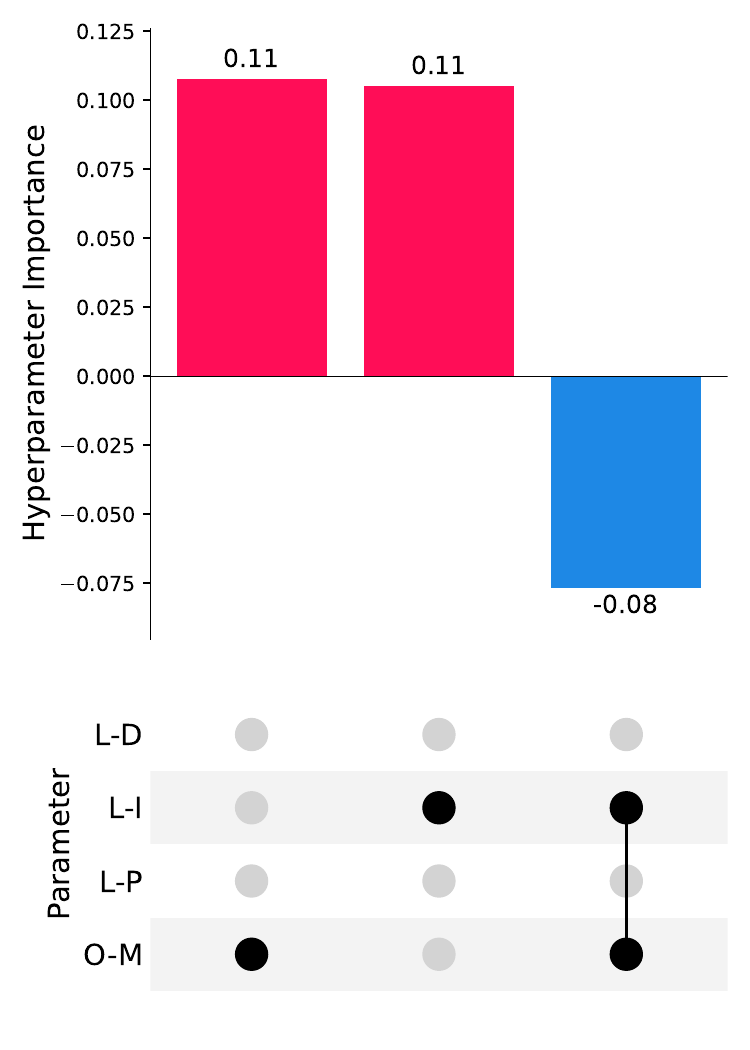}
    \end{minipage}
    \caption{Upset plots for \ablation (left) and \dstunability (right) of~\texttt{lm1b\_transformer} \cite{wang-jmlr24}.}
    \label{fig_exp_ablation_tunability}
\end{figure}

\begin{figure}[t]
    \centering
    \begin{minipage}[c]{0.48\columnwidth}
        \centering
        \includegraphics[width=.75\textwidth]{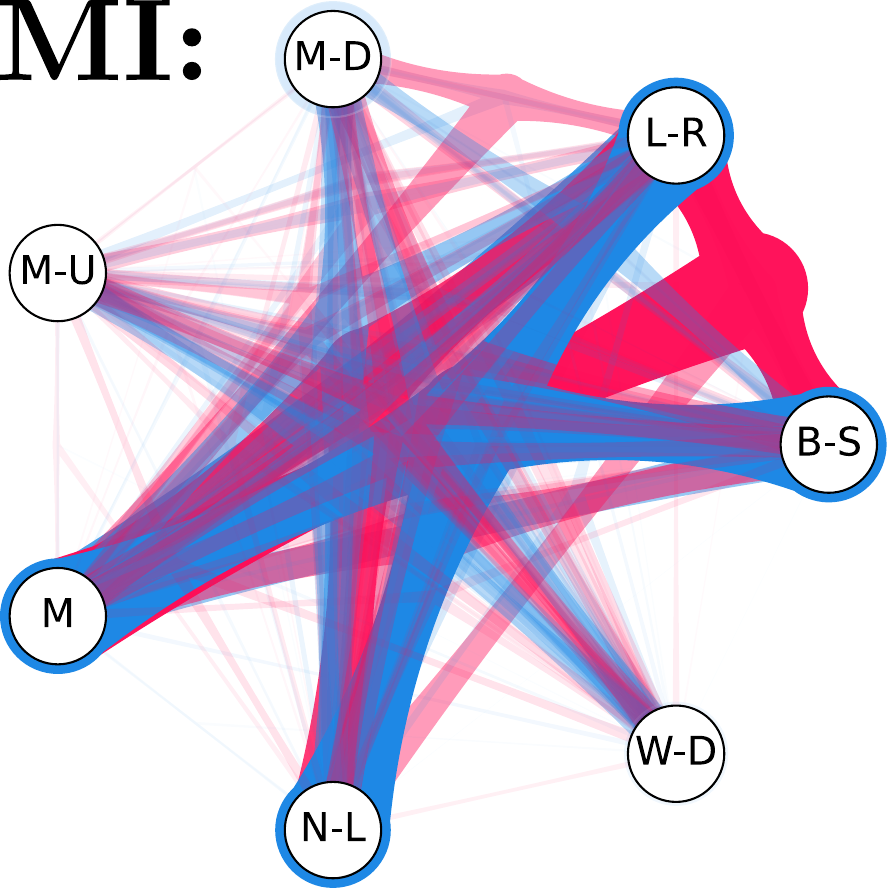}
        \\
        \textbf{a)} Individual Tuning
    \end{minipage}
    \hfill
    \begin{minipage}[c]{0.48\columnwidth}
        \centering
        \includegraphics[width=.75\textwidth]{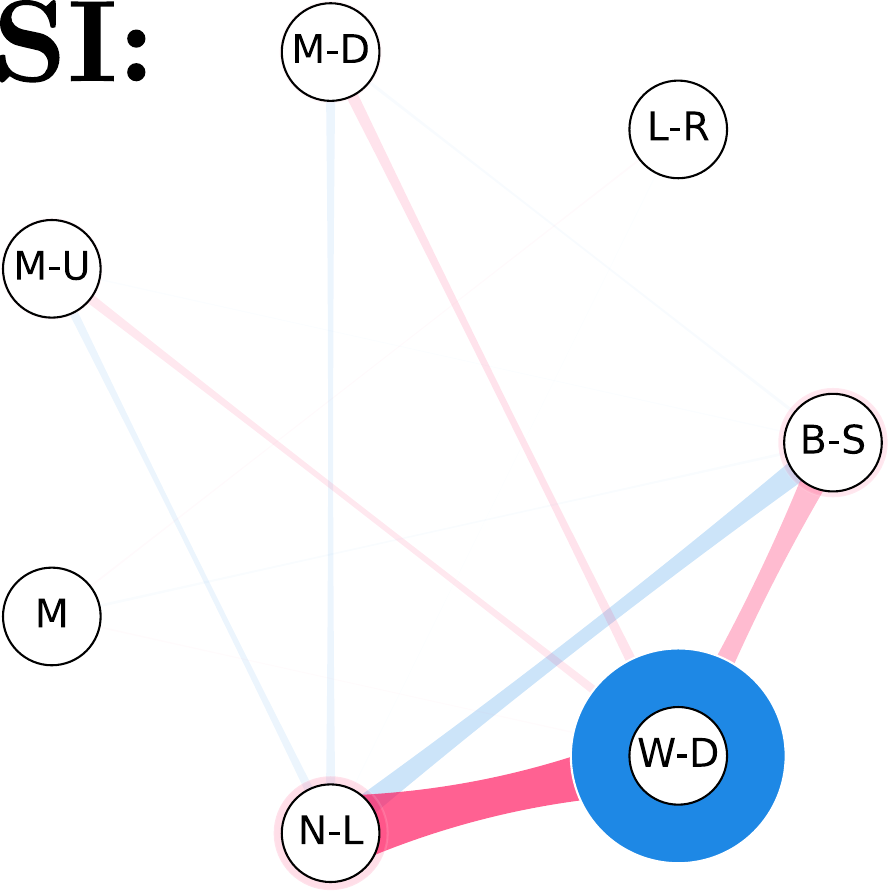}
        \\
        \textbf{b)} W-D not Tuned
    \end{minipage}
    \caption{Interaction graphs showing results for the \opthabit game via Moebius interactions (MI) and Shapley interactions (SI) on dataset ID 3945 of \texttt{lcbench}.}
    \label{fig_optbias_games}
\end{figure}

First, we compare the results of the \ablation and the \dstunability game in terms of hyperparameter importance and interactions (cf.~\myfigref{fig_exp_ablation_tunability}). We retrieve an optimized configuration of \pdone's \texttt{lm1b\_transformer} scenario and explain it with the \ablation game. \tool's explanation shows that the majority of the performance increase is attributed to the initial learning rate (L-I), which is not surprising since it is also intuitively the most important one. However, using \tool to create \dstunability explanations reveals that both hyperparameters, L-I and optimizer momentum (O-M), are of equal importance with a negative interaction. Thus, the optimizer chose to tune L-I over O-M for the configuration in question, even though a similar performance improvement could have been achieved by tuning O-M instead. Hence, \tool can reveal which hyperparameters were subject to optimization via the \ablation game, while the \dstunability game emphasizes the potential contributions of hyperparameters and their interactions.

\subsection{Higher-Order Interactions in HPO}\label{sec:exp-higher-order-interactions}\label{sex_exp_interaction_quantification}

Second, we investigate the interaction structure of HPO problems for individual and across datasets. 
In \myfigref{fig_interaction_quantification}, left (\gls*{MI}), and further in the appendix, 
we observe the presence of many higher-order interactions, which are difficult to interpret.
The \glspl*{SI} (order 2) and \gls*{SV} in \tool summarize the \gls*{MI} into \emph{interpretable} explanations.

\myfigref{fig_interaction_quantification}, right, shows that \glspl*{SI} still \emph{faithfully} capture the overall game behavior, which we measure with a Shapley-weighted loss \cite{Muschalik.2024} and varying explanation order. 
We find that most of the explanatory power is captured by interactions up to the third order, confirming prior research that suggests hyperparameter interactions are typically of lower order \cite{DBLP:conf/gecco/PushakH20a}. 
Interactions beyond the third order contribute little to the overall understanding of the game. 
Thanks to the convenient properties of the \gls*{SV} and \glspl*{SI}, \tool provides a reliable way to capture and fairly summarize higher-order interactions into more interpretable explanations.

\begin{figure*}[t]
    \centering
    \includegraphics[width=.9\textwidth]{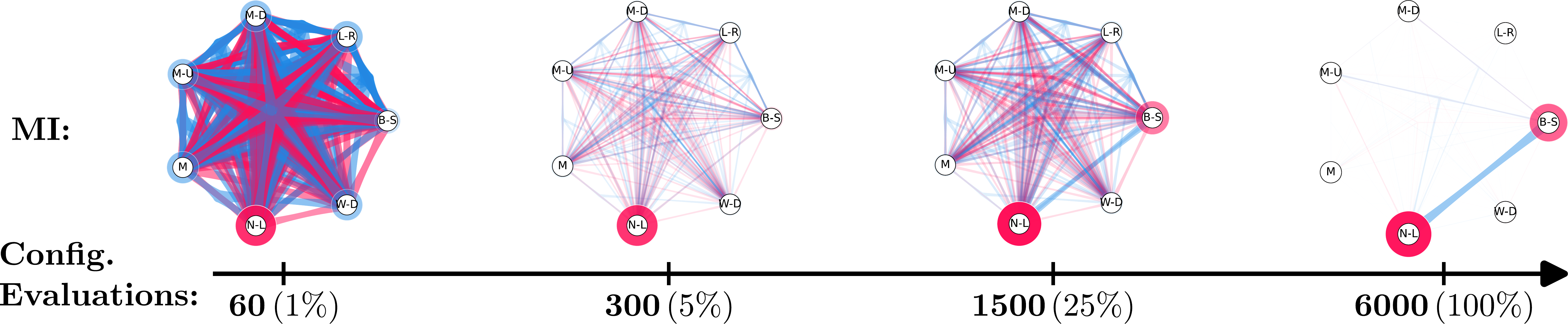}
    \caption{Explaining the surrogate model in SMAC's Bayesian optimization with MIs at 1\%, 5\%, 25\%, and 100\% of the budget. Over time, SMAC notices first the importance of N-L and later B-S.}
    \label{fig_appx_surrogate_explanation}
\end{figure*}

\subsection{Detecting Optimizer Bias}

The third experiment uses the \opthabit game to investigate biases in black-box hyperparameter optimizers. To this end, we create two artificially biased hyperparameter optimizers. The first optimizer tunes each hyperparameter \textit{separately}, ignoring interactions between them, while the second is \textit{not allowed} to tune the most important hyperparameter. The virtual best hyperparameter optimizer is an ensemble of the investigated optimizer and five random search optimizers with a budget of 10,000 samples each. Ideally, a perfect optimizer would show no interactions and no main effects in \tool's \opthabit explanations as the differences for every coalition would be 0.

\cref{fig_optbias_games} shows the \opthabit explanations, i.e., the difference between two \dstunability games, using the optimizer's returned value and the (approximated) maximum, respectively.
Note that main effects in the \opthabit game can only be negative and show the optimizer's inability to properly tune certain hyperparameters. In \cref{fig_optbias_games}\textbf{a}, small main effects, in turn, suggest that the optimizer can effectively tune hyperparameters individually. The presence of both negative and positive interactions, which result from missing out on positive and negative interactions, respectively, shows that it fails to capture interactions. This confirms that the optimizer, which tunes hyperparameters independently, fails to capture their joint synergies. On the other hand, the second optimizer, ignoring the weight decay (W-D) hyperparameter for this particular dataset, clearly demonstrates bias in the interaction graph in \cref{fig_optbias_games}\textbf{b}. The blue main effect for W-D and interactions involving W-D reveal this bias, showing how \tool can help identify such flaws and contribute to the development of more effective HPO methods.

\subsection{Explaining Bayesian Optimization}

Inspired by \citet{LMU}, we use \tool to explain SMAC \cite{lindauer-jmlr22a} -- a state-of-the-art hyperparameter optimizer based on Bayesian optimization -- by analyzing its surrogate model throughout the optimization process. We run SMAC with a total budget of $6\,000$ evaluations and inspect the surrogate model at $1\%, 5\%, 25\%, $ and $100\%$ of the budget. As shown in \cref{fig_appx_surrogate_explanation}, we observe how the model's belief about the main effects of and interactions between hyperparameters evolves over time.

Early in the process, the surrogate model exhibits numerous large higher-order interactions, reflecting high uncertainty and a broad range of plausible interactions. Despite this, it already identifies N-L as an important hyperparameter. As optimization progresses, the model's uncertainty decreases, leading to lower interaction values. By 25\% of the budget, the optimizer uncovers B-S as another important hyperparameter, and the surrogate model briefly broadens its hypothesis about the performance landscape, resulting in increased interaction values. Eventually, as SMAC converges, the model refines its understanding of the performance landscape and interactions that are considered plausible are reduced. This analysis illustrates how \tool can provide insights into the evolving dynamics of HPO processes.

\subsection{Comparison with fANOVA}

\begin{figure}[t]
    \centering
    \begin{minipage}[c]{0.47\columnwidth}
        \includegraphics[width=\textwidth]{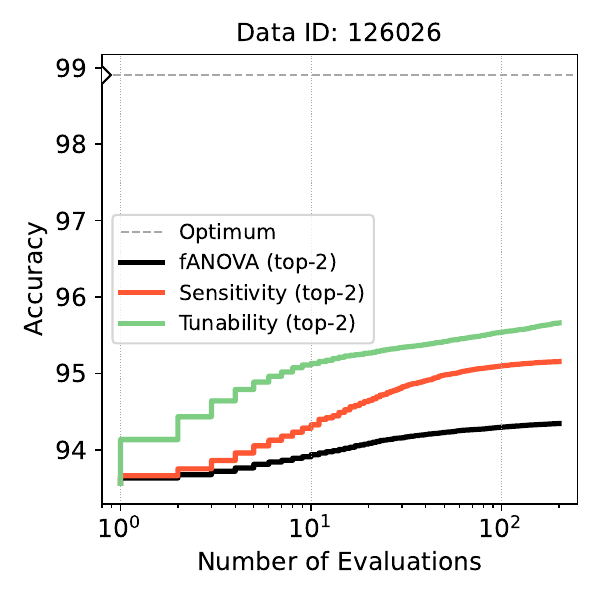}
    \end{minipage}
    \hfill
    \begin{minipage}[c]{0.47\columnwidth}
        \includegraphics[width=\textwidth]{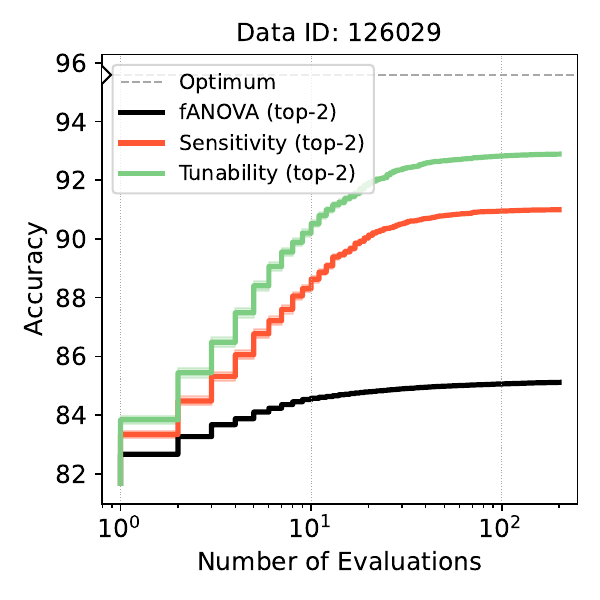}
    \end{minipage}
    \caption{Anytime performance plots of HPO runs involving only the top-2 important hyperparameters for two datasets of \texttt{lcbench} \cite{zimmer-tpami21a}.}
    \label{fig_fanova_comparison}
\end{figure}

Lastly, considering the experimental setting proposed with fANOVA by \citet{fANOVA}, we compare \tool and fANOVA on the task of selecting a subset of important hyperparameters to tune.
To this end, for a given HPO task, we run fANOVA and \tool using the \gls*{SV} with the \dstunability and \sensitivity game to obtain explanations of order 1. 
Selecting the two most important hyperparameters, we conduct a subsequent HPO run optimizing only these two hyperparameters.
Since \dstunability quantifies performance gains~\cite{tunability}, we expect explanations of this method to be more suitable for this specific task than explanations from the \sensitivity game, which quantifies variances in performances while tuning a certain hyperparameter. Also, the importance quantified over all variances, as in fANOVA, could yield less actionable explanations.

The results in \cref{fig_fanova_comparison} confirm that the anytime performance of the runs informed by \tool is superior to that informed by fANOVA, and that \dstunability outperforms \sensitivity here.
For this specific task, the explanations from the \dstunability game are more suitable.

\subsection{Runtime Analysis}
We found \tool to be efficient across all evaluated settings. An \ablation game took 5s to 2m, while \dstunability ranged from 6m to 15m for 7 and 4 hyperparameters, respectively, showing high dependency on the surrogate's efficiency, and up to 8.5h for 10 hyperparameters on the \jahs benchmark. As all coalition evaluations are independent, the method is highly parallelizable, enabling substantial wall-clock reductions. Overall, this adds modest overhead to typical multi-hour HPO runs.

\section{Conclusion}
In this paper, we proposed \tool, a post-hoc explanation framework for consistently and uniformly explaining hyperparameter optimization using the \gls*{SV} and \glspl*{SI} across three levels: hyperparameter values, sensitivity and tunability of learners, and optimizer capabilities. Unlike previous methods that quantify variance \cite{fANOVA}, \tool attributes performance contributions. We demonstrated that \tool not only enhances understanding of the impact of hyperparameter values or tunability of learners but also provides actionable insights for optimizing hyperparameters on related tasks.

The computational bottleneck is the approximation of the $\max$ over $\conf \in \confs$ via simulated HPO, requiring research on more efficient yet unbiased methods, e.g., via Bayesian algorithm execution \cite{bobax}. 
Furthermore, extensions of \tool to the analysis of optimizing machine learning pipelines are important future work~\citep{HeffetzVKR20,FeurerEFLH22,wever-ida2020a,wever-tpami21a}.
Additionally, we plan to develop HPO methods that use HPI to learn across datasets for improving their efficiency. This may allow warm-starting HPO in an interpretable way, complementing recent work on prior-guided HPO~\citep{HvarfnerHN24,fehring-arxiv25} and human-centered AutoML~\citep{LindauerKKMT0HF24}.

\section*{Acknowledgements}
Marcel Wever and Marius Lindauer gratefully acknowledge funding by the European Union (ERC, ``ixAutoML'', grant no.101041029). Views and opinions expressed are however those of the author(s) only and do not necessarily reflect those of the European Union or the European Research Council Executive Agency. Neither the European Union nor the granting authority can be held responsible for them. Maximilian Muschalik gratefully acknowledges funding by the Deutsche Forschungsgemeinschaft (DFG, German Research Foundation): TRR 318/1 2021 – 438445824.

\bibliography{aaai2026} 

\clearpage

\appendix
\onecolumn
\section*{Organization of the Appendix}

The technical supplement is organized as follows.

\contentsline {section}{\numberline{A}Proofs}{22}{}
\contentsline {subsection}{\numberline{A.1}Proof of Proposition 1}{11}{}
\contentsline {subsection}{\numberline{A.2}Proof for Table 1}{11}{}
\contentsline {subsection}{\numberline{A.3}Proof of Theorem 1}{12}{}
\contentsline {subsection}{\numberline{A.4}Example: Non-Monotone Sensitivity Game}{12}{}
\contentsline {section}{\numberline{B}Experimental Setup}{13}{}
\contentsline {subsection}{\numberline{B.1}Considered Benchmarks}{13}{}
\contentsline {subsection}{\numberline{B.2}Approximation of the argmax}{14}{}
\contentsline {subsection}{\numberline{B.3}Computing Optimizer Bias}{14}{}
\contentsline {subsection}{\numberline{B.4}Hardware Usage and Compute Resources}{14}{}
\contentsline {section}{\numberline{C}Guidance on Interpreting Interaction Visualizations}{15}{}
\contentsline {section}{\numberline{D}Interaction Quantification in Hyperparameter Optimization}{17}{}
\contentsline {subsection}{\numberline{D.1}Measuring the Magnitude of Interactions}{17}{}
\contentsline {subsection}{\numberline{D.2}Analyzing Lower-Order Representations of Games}{17}{}
\contentsline {subsection}{\numberline{D.3}Additional Experimental Details}{18}{}
\contentsline {section}{\numberline{E}Additional Empirical Results}{19}{}
\contentsline {subsection}{\numberline{E.1}Additional Information for the Comparison of Ablation and Tunability}{19}{}
\contentsline {subsection}{\numberline{E.2}Additional Results for Comparison with fANOVA}{19}{}
\contentsline {subsection}{\numberline{E.3}Additional Results for Explaining the SMAC Surrogate During Optimization}{20}{}
\contentsline {subsection}{\numberline{E.4}Additional Interaction Visualizations}{20}{}

\clearpage
\section{Proofs}\label{appx_sec_proofs_notation}

\subsection{Proof of Proposition 1}
\begin{proof}
The \dstunability game is given by the value function
\[
\nu(S) := \nu_{G_T}(S) := \max_{\conf \in \confs}
\,\val(\conf \oplus_S \conf^0, D) \,\, .
\]
We now want to show monotonicity of the value function, i.e., $S \subseteq T$ implies $\nu(S) \leq \nu(T)$. 
Given a coalition $T \subseteq \mathcal N$ with $S \subseteq T$, we immediately see that
\[
    A := \{\conf \oplus_{S} \conf^0: \conf \in \confs \} \subseteq \{ \conf \oplus_{T} \conf^0: \conf \in \confs \} =: B,
\]
since we can set the hyperparameters of $T\setminus S$ to $\conf^0 \in \confs$ on the right-hand side.
Since the \dstunability game takes the $\max$ over these two sets, respectively, we obtain
\begin{align*}
    \nu(S) &=  \max_{\conf \in \confs} \,\val(\conf \oplus_S \conf^0, D)\\
    &= \max_{\conf^\ast \in A} \val(\conf^\ast, D) \overset{A \subseteq B}{\leq} \max_{\conf^\ast \in B} \val(\conf^\ast, D) \\
    &=  \max_{\conf \in \confs}\,\val(\conf \oplus_T \conf^0, D)\\
    &= \nu(T).
\end{align*}

This concludes that the \dstunability game is monotone.
As a consequence, we obtain non-negative \glspl*{SV} due to the \emph{monotonicity} axiom \citep{Fujimoto.2006} of the \gls*{SV}.
We can also give a direct proof of this via the well-known representation of the \gls*{SV} in terms of a weighted average over marginal contributions as
\begin{align*}
    \phi^{\text{SV}}(i) := \sum_{T \subseteq \cN \setminus \{i\}} \frac{1}{n \cdot \binom{n-1}{\vert T \vert }} \big(\nu(T \cup \{i\}) - \nu(T)\big).
\end{align*}
Due to the monotonicity of $\nu_{G_T}$, it follows that $\nu_{G_T}(T) \leq \nu_{G_T}(T \cup i)$, and thus all terms in the above sum are non-negative.
Consequently, the \gls*{SV} is non-negative.

Moreover, the pure individual (main) effects obtained from the functional ANOVA framework are represented by the \gls*{MI} of the individuals \citep{Fumagalli.2024a}.
By the monotonicity of $\nu$, we obtain again
\[
    m(i) := \nu(i) - \nu(\emptyset) \geq 0,
\]
which concludes the proof.
\end{proof}

\subsection{Proof for Table 1}
\begin{proof}
    Given the synthetic \dstunability and \sensitivity game with two dimensions $\mathcal N = \{1,2\}$, our goal is to show that the main and interaction effects are given by \cref{tab:synthetic_example}.

    \paragraph{Tunability Game.}
    We first proceed to compute the game values of the \dstunability game for $S \subseteq \mathcal N$ with the optimal configuration $\conf^\ast = (1,m)$ as
    \begin{align}\label{appx_eq_synth_tunability}
    \nu_{G_T}(S) &= \max_{\conf \in \confs}\val(\conf \oplus_S \conf^0, D)\\ 
    &= \val(\conf^* \oplus_S \conf^0, D)\\
    &= \sum_{i \in \{1,2\}} \mathbf{1}_{(\conf \oplus_S \conf^0)_i = \lambda^\ast_i} \\
    &= \vert S \vert + \sum_{i \in \mathcal N: i \notin S} \mathbf{1}_{\lambda^0_i = \lambda^\ast_i}.
    \end{align}.

For the baseline set to $\conf^0 := (0,0)$, the second sum in \cref{appx_eq_synth_tunability} vanishes and we thus obtain
\[
    \nu_{G_T}(S) = \begin{cases}
        0, &\text{ if } S = \emptyset, \\
        1, &\text{ if } \vert S \vert = 1, \\
        2, &\text{ if } S=\{1,2\}.
    \end{cases}
\]
Hence, the \glspl*{MI} are given by
\[
    m_{G_T}(S) =\sum_{L \subseteq S}(-1)^{\vert S \vert - \vert L \vert} \nu_{G_T}(L) = \begin{cases}
        0, &\text{ if } S = \emptyset, \\
        1, &\text{ if } \vert S \vert = 1, \\
        0, &\text{ if } S=\{1,2\}.
    \end{cases}
\]

Clearly, the interaction $\lambda_1 \times \lambda_2$, i.e., $m(\{1,2\})$, vanishes, and thus the HPI scores of the individuals are given by their main effects in terms of the \glspl*{MI}.
In summary, the HPI main effects using the \gls*{SV} and the \gls*{MI} are both equal to $1$, whereas the interaction is zero, confirming the values shown in \cref{tab:synthetic_example}.

For the baseline set to $\conf^0 := \conf^\ast$, the second sum in \cref{appx_eq_synth_tunability} equals  $\vert \mathcal N \vert - \vert S \vert$ and thus we obtain a constant game
\[
    \nu_{G_T}(S) = \vert S \vert + \vert \mathcal N \vert - \vert S \vert = 2 \text{ for all } S\subseteq \mathcal N.
\]
Consequently, all interactions and main effects are zero due to the dummy axiom \citep{Fujimoto.2006}, confirming \cref{tab:synthetic_example}.

\paragraph{Sensitivity Game.}
We now proceed to compute the game values of the \sensitivity game for $S \subseteq \mathcal N$.
First, for $S=\emptyset$, we obtain $\nu_{G_V}(\emptyset)=0$, since $\conf \oplus_{\emptyset} \conf^0 = \conf^0$, and thus there is no variance with respect to $\conf$.
Due to independence of the hyperparameter distribution, we can decompose the variance as
\begin{align*}\label{appx_eq_synth_sens}
    \nu_{G_V}(S) &=  \mathbb{V}_{\conf^\ast \sim p(\conf^\ast)}[\val(\conf^\ast \oplus_S \conf^0, D)]\\
    &=  \mathbb{V}_{\conf^\ast \sim p(\conf^\ast)}[\sum_{i \in \{1,2\}} \mathbf{1}_{(\conf^\ast \oplus_S \conf^0)_i = \lambda^0_i}] \\
    &= \sum_{i \in S} \mathbb{V}_{\lambda_i^\ast \sim p(\lambda_i^\ast)}[\mathbf{1}_{(\conf^\ast \oplus_S \conf^0)_i = \lambda^0_i}]
\end{align*}

To compute $\mathbb{V}_{\lambda_i^\ast \sim p(\lambda_i^\ast)}[\mathbf{1}_{(\conf^\ast \oplus_S \conf^0)_i = \lambda^0_i}]$, we note that $\mathbf{1}_{(\conf^\ast \oplus_S \conf^0)_i = \lambda^0_i}$ is described by a Bernoulli variable.

Given any baseline, we have $\mathbf{1}_{(\conf^\ast \oplus_S \conf^0)_i = \lambda^0_i} \sim \text{Ber}(q_i)$ with $q_1 = 1/2$ and $q_2 = 1/(m+1)$ due to the uniform distribution, which sets this value to $1$, if the optimal configuration value is chosen.
The variance of this Bernoulli variable is then given by $q(1-q)$, which yields 
\[
    \mathbb{V}_{\lambda_i^\ast \sim p(\lambda_i^\ast)}[\mathbf{1}_{(\conf^\ast \oplus_S \conf^0)_i = \lambda^0_i}] = \begin{cases}
        \frac{1}{4}, &\text{if } i=1, \\
        \frac{1}{m+1}(1-\frac{1}{m+1}) =  \frac{m}{(m+1)^2}, &\text{if } i=2,
    \end{cases}
\]
which yields the game values 
\[
 \nu_{G_V}(S) = 
        \begin{cases}
        0, &\text{ if } S = \emptyset, \\
        \frac{1}{4}, &\text{ if } S = \{1\}, \\
        \frac{m}{(m+1)^2}, &\text{ if } S = \{2\}, \\
         \frac{1}{4} + \frac{m}{(m+1)^2}, &\text{ if } S=\{1,2\}.
        \end{cases}
\]
Hence, the \glspl*{MI} are given by

\[
 m_{G_V}(S) =\sum_{L \subseteq S}(-1)^{\vert S \vert - \vert L \vert} \nu_{G_V}(L) = 
        \begin{cases}
        0, &\text{ if } S = \emptyset, \\
        \frac{1}{4}, &\text{ if } S = \{1\}, \\
        \frac{m}{(m+1)^2}, &\text{ if } S = \{2\}, \\
        0, &\text{ if } S=\{1,2\},
        \end{cases}
\]
which confirms the values given in \cref{tab:synthetic_example} and concludes the proof.
\end{proof}

\subsection{Proof of Theorem 1}\label{apx:proof-error-bound}
\begin{proof}
Consider an upper bound for the approximation $\vert \hat\nu(T) - \nu(T) \vert \leq \epsilon$ for all $T \subseteq N$. We can then naively bound the difference of Shapley values using its representation over marginal contributions as follows: 
\[
\vert \phi_{\hat\nu}(i) - \phi_{\nu}(i) \vert \leq \sum_{T \subseteq N \setminus i} \frac{1}{n\binom{n-1}{t}}\vert\hat\nu(T \cup i)-\nu(T\cup i) - (\hat\nu(T) - \nu(T))\vert \leq \sum_{T \subseteq N \setminus i} \frac{1}{n\binom{n-1}{t}}2 \epsilon = 2\epsilon \, ,
\]
where we have used for the sum that the Shapley value is a probabilistic value. A similar bound can be established for order-$k$ Shapley interactions by bounding their discrete derivatives with $2^k \epsilon$.
\end{proof}

\subsection{Example: Non-Monotone Sensitivity Game}\label{appx_sec_non_monotone_example}
In this section, we give an example of a non-monotone \sensitivity game.
To this end, we consider two hyperparameters $\mathcal N = \{1,2\}$ equipped with independent Bernoulli distributions $\lambda_1,\lambda_2 \overset{\text{iid}}{\sim} \text{Ber}(1/2)$.
We consider a performance measure as
\begin{align*}
    \val(\conf) := \mathbf{1}_{\lambda_1=0}\mathbf{1}_{\lambda_2=0},
\end{align*}
and set the baseline configuration to $\conf^0 := (0,0)$. 
The \sensitivity game values are then computed by observing that $\val(\conf^\ast)$ with $\conf^\ast \sim p^\ast(\conf^\ast)$ is described as a Bernoulli variable Ber($q$).
For $S=\{1,2\}$, the probability of $\val$ being 1 is $q=1/4$, since both hyperparameters have to be set to zero.
In contrast, for $\vert S \vert = 1$, we have $q=1/2$, since the remaining variable is already set at zero due to the baseline configuration.
We thus obtain again the variances with $q(1-q)$ as 
\[
\nu_{G_V}(S) = \mathbb{V}_{\conf^\ast \sim p^\ast(\conf^\ast)}[\mathbf{1}_{\lambda^\ast_1=0}\mathbf{1}_{\lambda^\ast_2=0}] = 
    \begin{cases}
        0, & \text{if } S=\emptyset, \\
        \frac{1}{2}\frac{1}{2} = \frac{1}{4}, &\text{if } \vert S \vert = 1, \\
        \frac{1}{4}\frac{3}{4} = \frac{3}{16} &\text{if } S= \{1,2\}.

    \end{cases}
\]
Hence, we obtain that $\nu_{G_V}(\{1\}) = 1/4 \geq 3/16 = \nu_{G_V}(\{1,2\})$, which shows that $\nu_{G_V}$ is not monotone.

\section{Experimental Setup}\label{app_experiment-setup}

Our implementation builds upon the \texttt{shapiq} package (version 1.1.1) \citep{Muschalik.2024}, which is publicly available on GitHub\footnote{\url{https://github.com/mmschlk/shapiq}} and distributed via pypi, for computing Shapley values and interactions. Furthermore, for the experiments, we use YAHPO-Gym \citep{DBLP:conf/automl/PfistererSMBB22}, a surrogate-based benchmark for multi-fidelity hyperparameter optimization. YAHPO-Gym provides several benchmark suites, i.a., \lcbench \citep{zimmer-tpami21a}, which we focused on in the main paper. However, in the subsequent sections, we also present results from the \rbvranger benchmark set, a random forest benchmark, from YAHPO-Gym, demonstrating the more general applicability of \tool. Furthermore, we run evaluations on the benchmark \pdone and \jahs to showcase \tool's wide applicability. In our repository, we provide pre-computed games to foster reproducibility of our results and allow for faster post-processing of the game values, e.g., for plotting different representations of the played games.

For better readability in terms of the font size, hyperparameter names are abbreviated in the interaction graphs. An overview of abbreviations and the names of their corresponding hyperparameters can be found in the subsequent section.

\subsection{Considered Benchmarks}\label{app_section_benchmarks}

\paragraph{lcbench \citep{DBLP:conf/automl/PfistererSMBB22,zimmer-tpami21a}.} \lcbench is a benchmark considering joint optimization of the neural architecture and hyperparameters that has been proposed by \cite{zimmer-tpami21a} together with the automated deep learning system Auto-PyTorch. The benchmark consists of 35 datasets with 2000 configurations each for which the learning curves have been recorded, allowing for benchmarking multi-fidelity HPO. However, in YAHPO-Gym only 34 of the 35 original datasets are contained which is why our evaluation is also restricted to those 34 datasets.
\begin{center}
    \begin{tabular}{l c c}
    \toprule
    \edit{Hyperparameter Name} & \edit{Abbreviation} & \edit{Type}\\
    \midrule
    \texttt{weight\_decay} & \edit{W-D} & \edit{float} \\
    \texttt{learning\_rate} & \edit{L-R} & \edit{float} \\
    \texttt{num\_layers} & \edit{N-L} & \edit{integer} \\
    \texttt{momentum} & \edit{M} & \edit{float} \\
    \texttt{max\_dropout} & \edit{M-D} & \edit{float} \\
    \texttt{max\_units} & \edit{M-U} & \edit{integer} \\
    \texttt{batch\_size} & \edit{B-S} & \edit{float} \\
    \bottomrule
    \end{tabular}
\end{center}

\paragraph{rbv2\_ranger \citep{DBLP:conf/automl/PfistererSMBB22}.} As already mentioned above, \rbvranger is a benchmark faced with tuning the hyperparameters of a random forest. We consider the hyperparameters of ranger as listed below:
\begin{center}
    \begin{tabular}{l c c}
    \toprule
    Hyperparameter Name & Abbreviation & Type\\
    \midrule
    \texttt{min\_node\_size} & M-N & integer \\
    \texttt{mtry\_power} & M-P & float \\
    \texttt{num\_impute\_selected\_cpo} & N-I & categorical \\
    \texttt{num\_trees} & N-T & integer \\
    \texttt{respect\_unordered\_factors} & R-U & categorical \\
    \texttt{sample\_fraction} & \edit{S-F} & float \\
    \texttt{splitrule} & S & categorical/Boolean \\
    \texttt{num\_random\_splits} & N-R & integer \\
    \bottomrule
    \end{tabular}
\end{center}

\paragraph{PD1 \citep{wang-jmlr24}.}
The \pdone benchmark is a testbed for evaluating hyperparameter optimization methods in the deep learning domain. It consists of tasks derived from realistic hyperparameter tuning problems, including transformer models and image classification networks. Across these different types of models, 4 hyperparameters are subject to tuning:
\begin{center}
\begin{tabular}{l c c}
    \toprule
    \edit{Hyperparameter Name} & \edit{Abbreviation} & \edit{Type}\\
    \midrule
    \edit{\texttt{lr\_decay\_factor}} & \edit{L-D} & \edit{float}\\
    \edit{\texttt{lr\_initial}} & \edit{L-I} & \edit{float}\\
    \edit{\texttt{lr\_power}} & \edit{L-P} & \edit{float}\\
    \edit{\texttt{opt\_momentum}} & \edit{O-M} & \edit{float}\\
     \bottomrule
\end{tabular}
\end{center}

\paragraph{JAHS-Bench-201 \citep{bansal-neurips22a}.}
To democratize research on neural architecture search, various table look-up and surrogate-based benchmarks have been proposed in the literature. Going even beyond plain neural architecture search, in \jahs, the combined task of searching for a suitable neural architecture and optimizing the hyperparameters of the learning algorithm is considered. We include it via the ```mf-prior-bench``` package that serves it with a surrogate model for predicting the validation error of a given architecture and hyperparameter configuration. The considered hyperparameters, including those for the neural architecture, are as follows:
\begin{center}
\begin{tabular}{lcc}
    \toprule
    \edit{Hyperparameter Name} & \edit{Abbreviation} & \edit{Type}\\
    \midrule
    \edit{\texttt{Activation}} & \edit{A} & \edit{categorical}\\
    \edit{\texttt{LearningRate}} & \edit{L} & \edit{float}\\
    \edit{\texttt{Op1}} & \edit{Op1} & \edit{categorical}\\
    \edit{\texttt{Op2}} & \edit{Op2} & \edit{categorical}\\
    \edit{\texttt{Op3}} & \edit{Op3} & \edit{categorical}\\
    \edit{\texttt{Op4}} & \edit{Op4} & \edit{categorical}\\
    \edit{\texttt{Op5}} & \edit{Op5} & \edit{categorical}\\
    \edit{\texttt{Op6}} & \edit{Op6} & \edit{categorical}\\
    \edit{\texttt{TrivialAugment}} & \edit{T} & \edit{Boolean}\\
    \edit{\texttt{WeightDecay}} & \edit{W} & \edit{float}\\
    \bottomrule
\end{tabular}
\end{center}

\subsection{Approximation of the argmax}\label{appx_section_approximation_of_argmax}
As per \cref{def:tunability} to \cref{def:opthabit}, for every coalition $S$, we need to determine the $\arg\max$. However, the true $\arg\max$ is difficult to determine, so we approximate it throughout our experiments. For the sake of implementation simplicity and unbiased sampling, we use random search with a large evaluation budget of $10\,000$ candidate evaluations. As the configurations are independently sampled, for evaluating a configuration, we simply blind an initially sampled batch of 10,000 hyperparameter configurations for the hyperparameters not contained in the coalition $S$ by setting their values to the default value. This procedure is fast to compute and reduces the noise potentially occurring through randomly sampling entirely new configurations for every coalition evaluation. After blinding, the surrogate model provided by YAHPO-Gym is then queried for the set of hyperparameter configurations, and the maximum observed performance is returned.

\begin{figure*}[htb]
    \includegraphics[width=\textwidth]{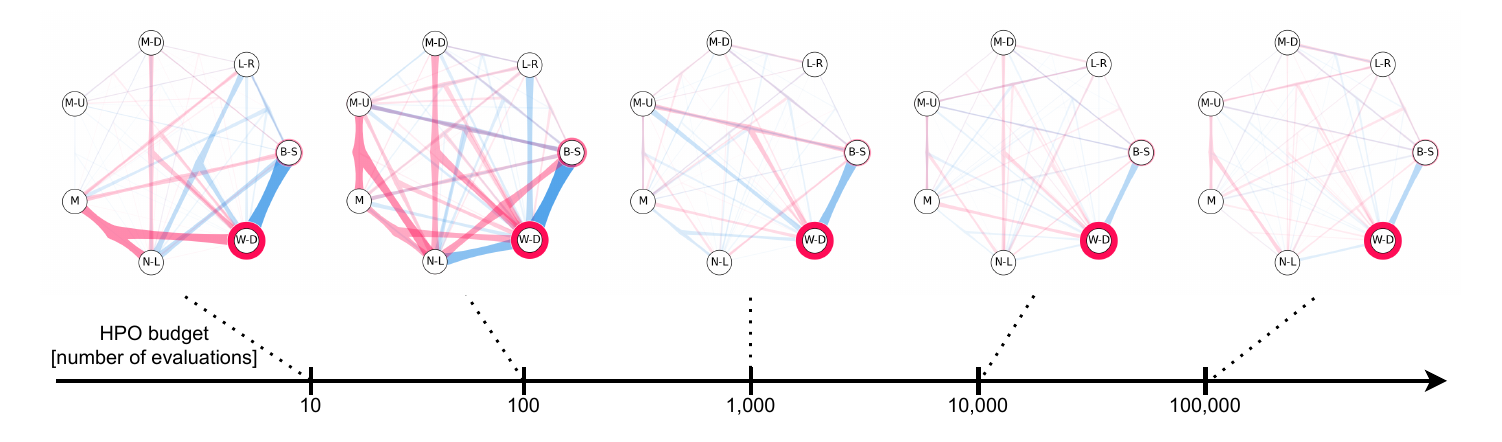}
    \caption{Hyperparameter importance with HyperSHAP, approximating the $\arg\max$ in Definition~\ref{def:tunability} of the value function \edit{via hyperparameter optimization with increasing budgets} for dataset ID 7593 of \lcbench. For tuning, we consider the following hyperparameters of \texttt{lcnet}: learning rate (L-R), batch size (B-S), weight decay (W-D), num layers  (N-L), momentum (M), max units (M-U), and max dropout (M-D).}
    \label{fig:hpo-quality}
\end{figure*}

In Figure~\ref{fig:hpo-quality}, we show how explanations evolve with higher budgets for simulating a hyperparameter optimization run with random search in combination with a surrogate model. To this end, we investigate explanations obtained through a random search with 10, 100, 1,000, 10,000, and 100,000 hyperparameter configurations sampled during optimization. We find that for low budgets of up to 1,000 samples, explanations are not really stable and change with higher budgets. In particular, we observe higher-order interactions that diminish for higher budgets, reflecting a decreasing uncertainty about the actual interactions. For the higher budgets of 10,000 and 100,000 hyperparameter configurations, the interaction graphs do not change as much, so 10,000 hyperparameter configurations appear to be a reasonable tradeoff between computational complexity and faithfulness of the explanations.
Therefore, we chose to conduct our experiments throughout the paper by simulating HPO runs with random search, simulating HPO with a surrogate model and a budget of 10,000 hyperparameter configurations.

\subsection{Computing \opthabit}
For the experiments considering the HPI game of Data-Specific \opthabit, we designed three HPO methods that focus on different structural parts of the hyperparameter configuration space. For the hyperparameter optimization approach, tuning every hyperparameter individually, when considering a hyperparameter for tuning, we sampled 50 random values for every hyperparameter. For the hyperparameter optimizer focusing on a subset of hyperparameters, we allowed for 50,000 hyperparameter configurations. For the VBO, we employed the considered limited hyperparameter optimizer and a random search with a budget of 50,000 evaluations on the full hyperparameter configuration space. We chose larger HPO budgets for these experiments to immediately ensure the built-in deficiencies become apparent and reduce noise effects. Howevér, they might also already be visible with substantially smaller budgets.

\subsection{Hardware Usage and Compute Resources}
Initial computations for lcbench and rbv2\_ranger have been conducted on consumer hardware, i.e., Dell XPS 15 (Intel i7 13700H, 16GB RAM) running Windows 11 and a MacBook Pro (M3 Max - 16C/40G, 128GB RAM) with MacOS 15. Overall computations took around 10 CPUd, highlighting \tool being lightweight when combined with surrogates. For measuring runtimes, we re-computed the games for \ablation and Data-Specific \tunability of lcbench and rbv2\_ranger and added PD1 and JAHS-Bench-201. The latter computations have been conducted on a high-performance computer with nodes equipped with 2$\times$ AMD Milan 7763 ($2\times 64$ cores)  and 256GiB RAM, running Red Hat Enterprise Linux Ootpa and Slurm, of which 1 core and 8GB RAM have been allocated to the computations for a single game. While the latter experiments amounted to 10.71 CPU days, in sum, the computations for this paper accumulate roughly 21 CPU days. Using the implementation at \url{https://github.com/mwever/hypershap}, the average runtimes per benchmark and game are as follows (\cref{tab:benchmark_runtime}):
\begin{table}[h!]
\centering
\caption{\edit{Mean $\pm$ standard deviation of the runtimes on a single CPU per benchmark and game.}}
\resizebox{\textwidth}{!}{
\begin{tabular}{lccccc}
\toprule
\edit{\textbf{Benchmark}} & \edit{\textbf{$|\confs|$}} & \edit{\textbf{$|\mathcal{D}|$}} & \edit{\textbf{Runtime Ablation [s]}} & \edit{\textbf{Runtime Tunability [s]}} & \edit{\textbf{Runtime Multi-Data Tunability [s]}} \\ 
\midrule
\edit{PD1} & \edit{4} & \edit{4} & \edit{64.9$\pm$16.0} & \edit{862.4$\pm$13.7} & \edit{-} \\
\edit{JAHS} & \edit{10} & \edit{3} & \edit{123.7$\pm$4.4} & \edit{30,406.7$\pm$4750.9 (8h26m)} & \edit{-} \\
\edit{LCBench} & \edit{7} & \edit{34} & \edit{4.8$\pm$0.4} & \edit{357.3$\pm$3.1} & \edit{10,713.4 (2h58m)} \\
\edit{rbv2\_ranger} & \edit{8} & \edit{119} & \edit{26.4$\pm$6.8} & \edit{6,717$\pm$767.3} & \edit{-} \\
\bottomrule
\end{tabular}}
\label{tab:benchmark_runtime}
\end{table}

After vectorization, implementing parallelization, and leveraging approximation methods for 15 hyperparmeters (HPs) and more, HyperSHAP's runtimes for \dstunability become much faster (\url{https://github.com/automl/hypershap}):

\begin{table}[h]
    \centering
    \caption{Mean runtimes in seconds across 10 runs of HyperSHAP's package implementation for \dstunability. Values that have not been computed are indicated by "-".}
    \label{tab:runtimes-package}
    \begin{tabular}{c|c|c|c|c|c|c|c|c|c|c|c|c|c|c}
        \toprule
        \#Cores / \#HPs & 2 & 3 & 4 & 5 & 6 & 7 & 8 & 9 & 10 & 11 & 12 & 13 & 14 & 15 \\
        \midrule
         1 &  0.2 & 0.3 & 0.5 & 1.0 & 2.0 & 3.9 & 7.8 & 15.8 & 31.7 & 62.7 & 125.0 & 251.0 & 501.4 & - \\
         8 & - & - & - & 0.3 & 0.5 & 0.9 & 1.8 & 3.5 & 6.9 & 13.9 & 27.4 & 55.2 & 111.4 & 97.2\\
         \bottomrule
    \end{tabular}
\end{table}

\section{Guidance on Interpreting Interaction Visualizations}\label{appx_guidance}

\begin{figure}[htb]
    \centering
    \includegraphics[height=20em]{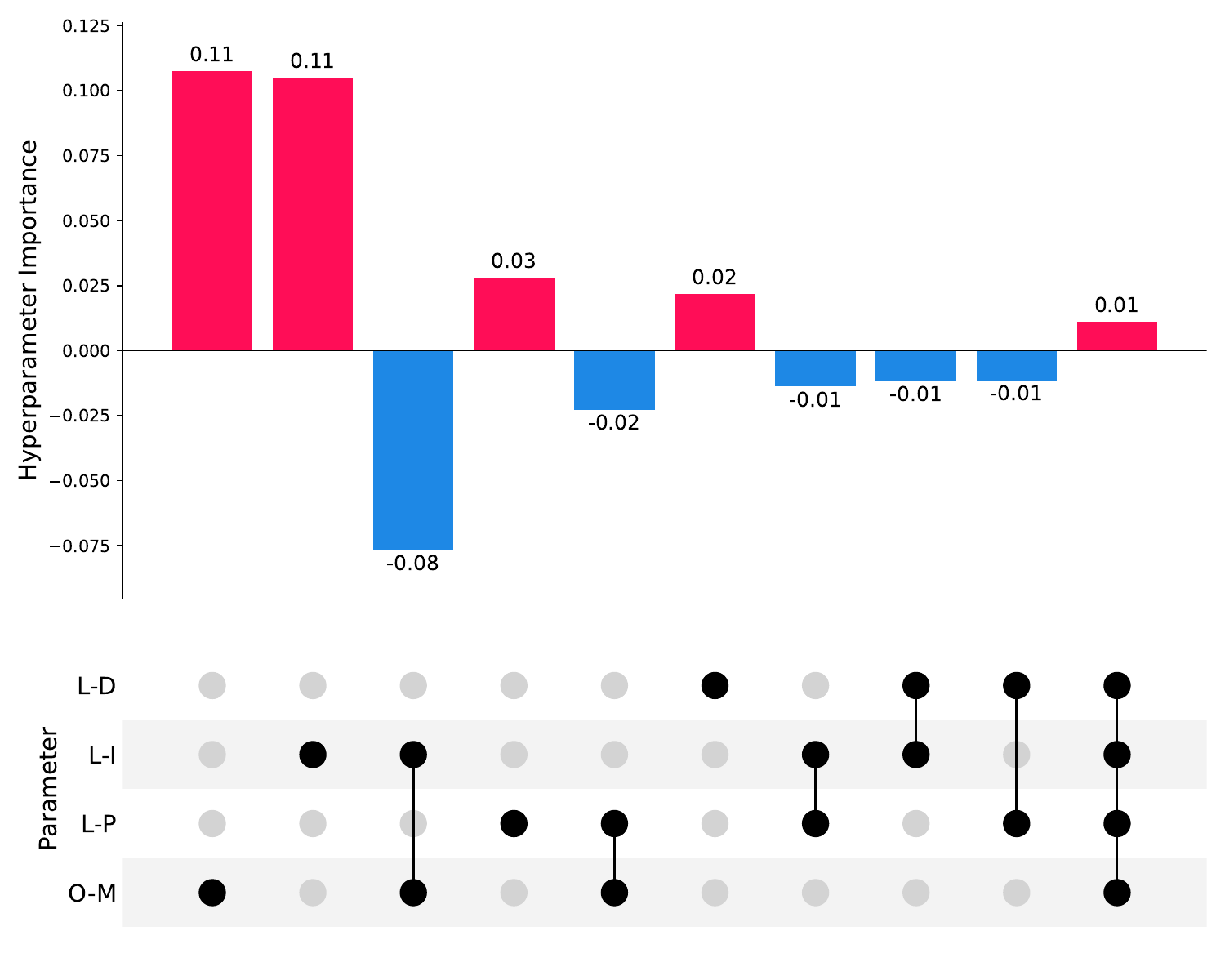}
    \includegraphics[height=20em]{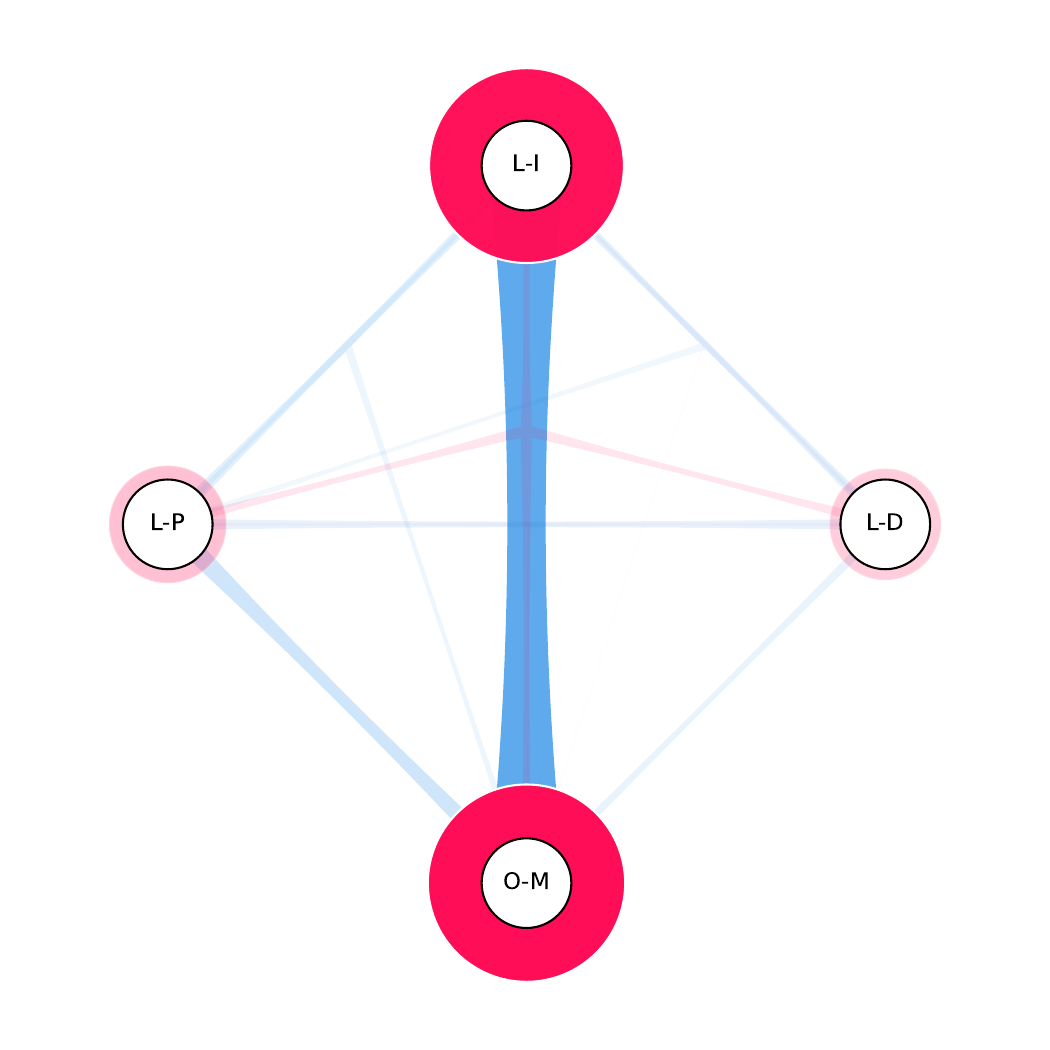}
    \caption{An UpSet plot (left) and a \gls*{SI} graph plot (right) for the \dstunability game from \cref{sec_experiments_ablation_tunability}.}
    \label{fig_si_graph_example}
\end{figure}

To visualize and interpret lower-, and higher-order interactions such as \gls*{SI} or \gls*{MI}, we employ the \emph{SI graph visualization} and the \emph{UpSet plot} from \texttt{shapiq} \citep{Muschalik.2024}.
The SI graph visualization is an extension of the network plot for Shapley interactions \cite{DBLP:conf/aaai/MuschalikFHH24} and can be used to visualize higher order interactions.
The UpSet plot \cite{Lex2014UpSet} is a well-established method for visualizing set-based scores, which can also be used for representing higher-order interactions.
\cref{fig_si_graph_example} shows an exemplary SI graph and UpSet plot.

For better readability in terms of the font size, hyperparameter names are abbreviated in the interaction graphs.

\paragraph{Interpretation of the UpSet Plot.}
An UpSet plot for \glspl*{SI} or \glspl*{MI} shows a selection of high-impact interactions and their scores. The plot is divided into two parts. The upper part shows the interaction values as bars and the lower part shows the considered interactions as a matrix. The first two bars in \cref{fig_si_graph_example} show the main effects of the O-M and L-I hyperparameters. The third bar shows the negative interaction of both of these features (denoted as the connection between the interactions).
A red color denotes a positive score, and a blue color denotes a negative score.
The bars and interactions are plotted in descending order according to the absolute value of an interaction (i.e., higher-impact interactions first).

\paragraph{Interpretation of the SI Graph.}
An SI graph plot in \cref{fig_si_graph_example} can be interpreted as follows.
Each individual player (e.g. hyperparameter) is represented as a node with connecting hyperedges representing the strength and direction of interactions.
Akin to the well-established force plots \citep{Lundberg.2017}, positive interactions are colored in red and negative interactions in blue, respectively.
The strength of an interaction is represented by the size and opacity of the hyperedge.
To reduce visual clutter, small interactions below a predefined absolute threshold may be omitted from the graph.
Notably, first-order interactions (i.e., individual player contributions, or main effects) are represented by the size of the nodes.

\clearpage
\section{Interaction Quantification in Hyperparameter Optimization}\label{app:interaction_quantification}

\subsection{Measuring the Magnitude of Interactions}
In this section, we provide further details for measuring the presence of interactions discussed in \cref{sex_exp_interaction_quantification}.
The \glspl*{MI} describe the pure additive effect of a coalition to the payout of the game.
They thus serve as an important tool to analyze the interactions present in a game $\nu$.
For instance, low-complexity games, where \glspl*{MI} are non-zero only up to coalitions of size $k$, are typically referred as $k$-additive games \citep{Grabisch.2016}.
In this case, \glspl*{SI} with explanation order $k$ perfectly recover all game values \citep{Bord.2023}.
In this case, the \glspl*{SI} correspond to the \glspl*{MI}.
We thus analyze the absolute values of \glspl*{MI} for varying size of coalitions, i.e., displaying the strata $q(k) := \{ \vert m(S) \vert : S\subseteq \mathcal N, \vert S \vert = k\}$ for varying interaction order $k=1,\dots,n$.
Analyzing $q(k)$ indicates, if the game $\nu$ has lower- order higher-order interactions present by investigating the magnitudes and distributions in the strata $q(k)$.

\subsection{Analyzing Lower-Order Representations of Games}\label{appx_sec_fsii}
In this section, we provide additional details for the lower-order representations and $R^2$ scores discussed in \cref{sec:exp-higher-order-interactions}.
The \gls*{SV} that capture the fair contribution in a game $\nu$ of an individual to the joint payout $\nu(\mathcal N)$.
However, the \gls*{SV} $\phi^{\text{SV}}(i)$ is also the solution to a constrained weighted least squares problem \citep{Charnes.1988,DBLP:conf/icml/FumagalliMKHH24}
\begin{align*}
    \phi^{\text{SV}} = \argmin_{\phi} \sum_{T\subseteq \mathcal N} \frac{1}{\binom{n-2}{\vert T \vert - 1}} \left(\nu(T)-\nu(\emptyset)-\sum_{i \in T} \phi(i) \right)^2 \text{ s.t. } \nu(\mathcal N) = \nu(\emptyset) + \sum_{i \in \mathcal N} \phi(i).
\end{align*}

In other words, the \gls*{SV} is the best additive approximation of the game $\nu$ in terms of this weighted loss constrained on the efficiency axiom.
Based on this result, the \gls*{FSII} \citep{Tsai.2022} was introduced as 

\begin{align*}
    \Phi_k^{\text{FSII}} := \argmin_{\Phi_k} \sum_{T \subseteq N} \mu(\vert T \vert) \left(\nu(T) - \sum_{S \subseteq T, \vert S \vert \leq k} \Phi_k(S)\right)^2
    \text{ with } \mu(t) := \begin{cases}
        \mu_\infty &\text{ if } t \in \{0,n\} 
        \\
        \frac{1}{\binom{n-2}{t-1}} &\text{ else}
    \end{cases},
\end{align*}
where the infinite weights capture the constraints $\nu(\emptyset)=\Phi_k(\emptyset)$ and $\nu(\mathcal N) = \sum_{S \subseteq \mathcal N} \Phi_k(S)$.
Note that \cite{Tsai.2022} introduces \gls*{FSII} with a scaled variant of $\mu$ that does not affect the solution.
The \gls*{FSII} can thus be viewed as the best possible approximation of the game $\nu$ using additive components up to order $k$ constrained on the efficiency axiom.
It is therefore natural to introduce the \emph{Shapley-weighted faithfulness} as
\begin{align*}
    \mathcal F(\nu,\Phi_k) := \sum_{T \subseteq N} \mu(\vert T \vert) \left(\nu(T) - \sum_{S \subseteq T, \vert S \vert \leq k} \Phi_k(S)\right)^2.
\end{align*}

Based on this faithfulness measure, the Shapley-weighted $R^2$ can be computed.
More formally, we compute the weighted average and the total sum of squares as
\begin{align*}
    \bar y := \frac{\sum_{T \subseteq \mathcal N} \mu(\vert T \vert) \nu(T)}{\sum_{T \subseteq \mathcal N} \mu(\vert T \vert)} \text{ and } \mathcal F_{\text{tot}} := \sum_{T \subseteq \mathcal N} \mu(\vert T \vert) \left(\nu(T) - \bar y\right)^2,
\end{align*}
which yields the Shapley-weighted $R^2$ as
\begin{align*}
    R^2(k) := R^2(\nu,\Phi_k) := 1 - \frac{\mathcal F(\nu,\Phi_k)}{\mathcal F_{\text{tot}}}.
\end{align*}

In our experiments, we rely on \gls*{FSII}, since this interaction index optimizes the faithfulness measure $\mathcal F$ by definition.
However, \gls*{k-SII} satisfies a similar faithfulness property \citep{DBLP:conf/icml/FumagalliMKHH24}.
Since the \gls*{FSII} is equal to the \glspl*{MI} for $k=n$, we have that $\mathcal F(\nu,\Phi_n) = 0$  due to the additive recovery property of the \glspl*{MI}.
Hence, $R^2(n)=R^2(\nu,\Phi_n)=0$ in this case.
Clearly, the $R^2(k)$ scores are monotonic increasing in $k$ by definition of \gls*{FSII}.
An $R^2(k) \approx 1$ indicates an almost perfect recovery of all game values.
In our experiments, we have shown that higher-order interactions are present, but lower-order representations (low $k$) are mostly sufficient to achieve very high $R^2$ scores.
This indicates that higher-order interactions are present but do not dominate the interaction landscape in our applications.
For instance, a single isolated higher-order interaction would yield much lower $R^2$ scores \citep{Muschalik.2024}.

\subsection{Additional Experimental Details}

\begin{figure}[htb]
    \centering
    \includegraphics[width=0.6\linewidth]{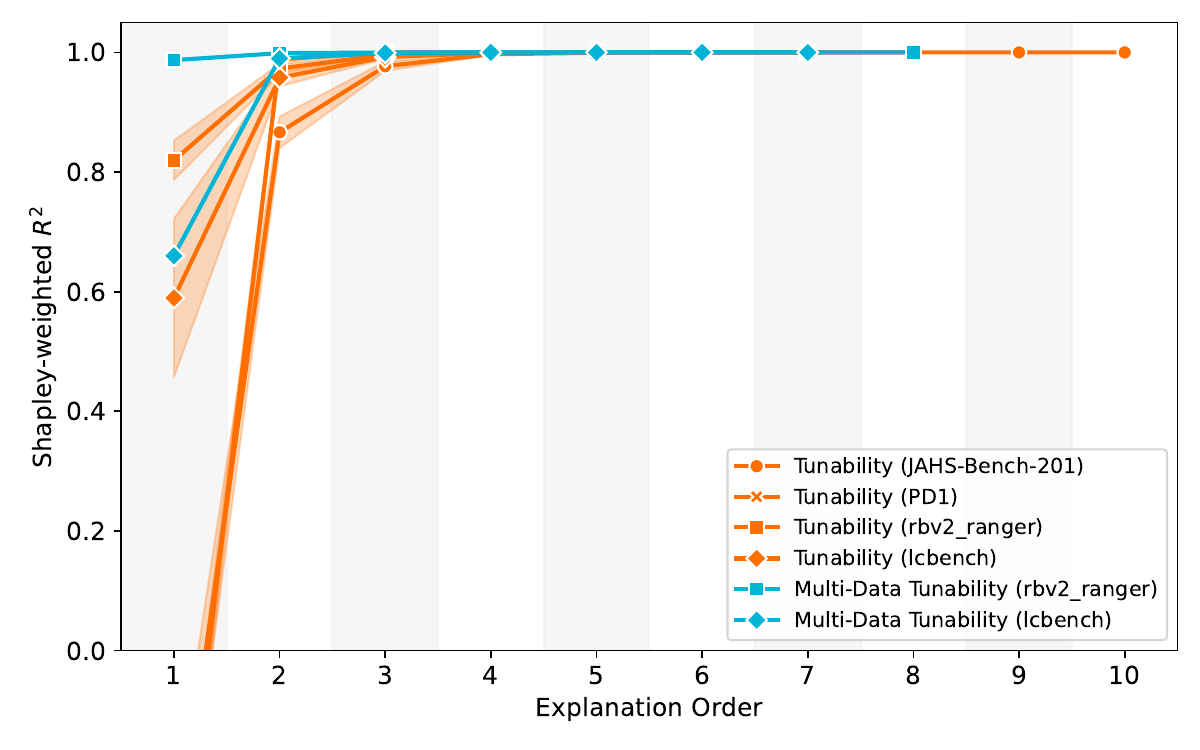}
    \caption{Detailed Reprint of \cref{fig_interaction_quantification} (right). Curves for \tunability contain only one game each. The \dstunability games for \lcbench and \rbvranger are averaged over $20$ randomly selected datasets. The \dstunability curves for \pdone and \jahs are averaged over all datasets contained in the benchmarks ($4$ and $3$, respectively). The shaded bands correspond to the standard error of the mean (SEM).}
    \label{fig_appx_faithfulness}
\end{figure}

In \cref{sex_exp_interaction_quantification}, we investigate how \emph{faithful} \tool explanations capture the interaction structures of the HPO problem. For this we compute \dstunability explanations for all four benchmarks, \lcbench, \rbvranger \pdone, and \jahs.
Further, we compute \tunability explanations for \lcbench and \rbvranger over all instances in the benchmarks.
We then compute the \glspl*{MI} for all of these explanations. We compute \tool \gls*{FSII} explanations up to the highest order. Then we compute the Shapley-weighted $R^2$ loss between the explanations and the original game as a measure of \emph{faithfulness}.
\cref{fig_appx_faithfulness} summarizes the results. 
The high $R^2$ score (almost $1.0$) for both the \dstunability and the \tunability games suggests that most of the explanatory power is captured by interactions up to the third order, \textbf{confirming prior research that suggests hyperparameter interactions are typically of lower order} \cite{DBLP:conf/gecco/PushakH20a}. 

\clearpage
\section{Additional Empirical Results}\label{appx_sec_additional_results}

This section contains additional experimental results, including more detailed plots and visualizations for the experiments conducted in \cref{sec_experiments}.

\subsection{Additional Information for the Comparison of Ablation and Tunability}\label{appx_sec_tunability_ablation_comparison}

\begin{figure}[htb]
    \centering
    \begin{minipage}[c]{0.05\textwidth}
        \rotatebox{90}{\textbf{\ablation}}
    \end{minipage}
    \begin{minipage}[c]{0.45\textwidth}
        \includegraphics[width=\textwidth]{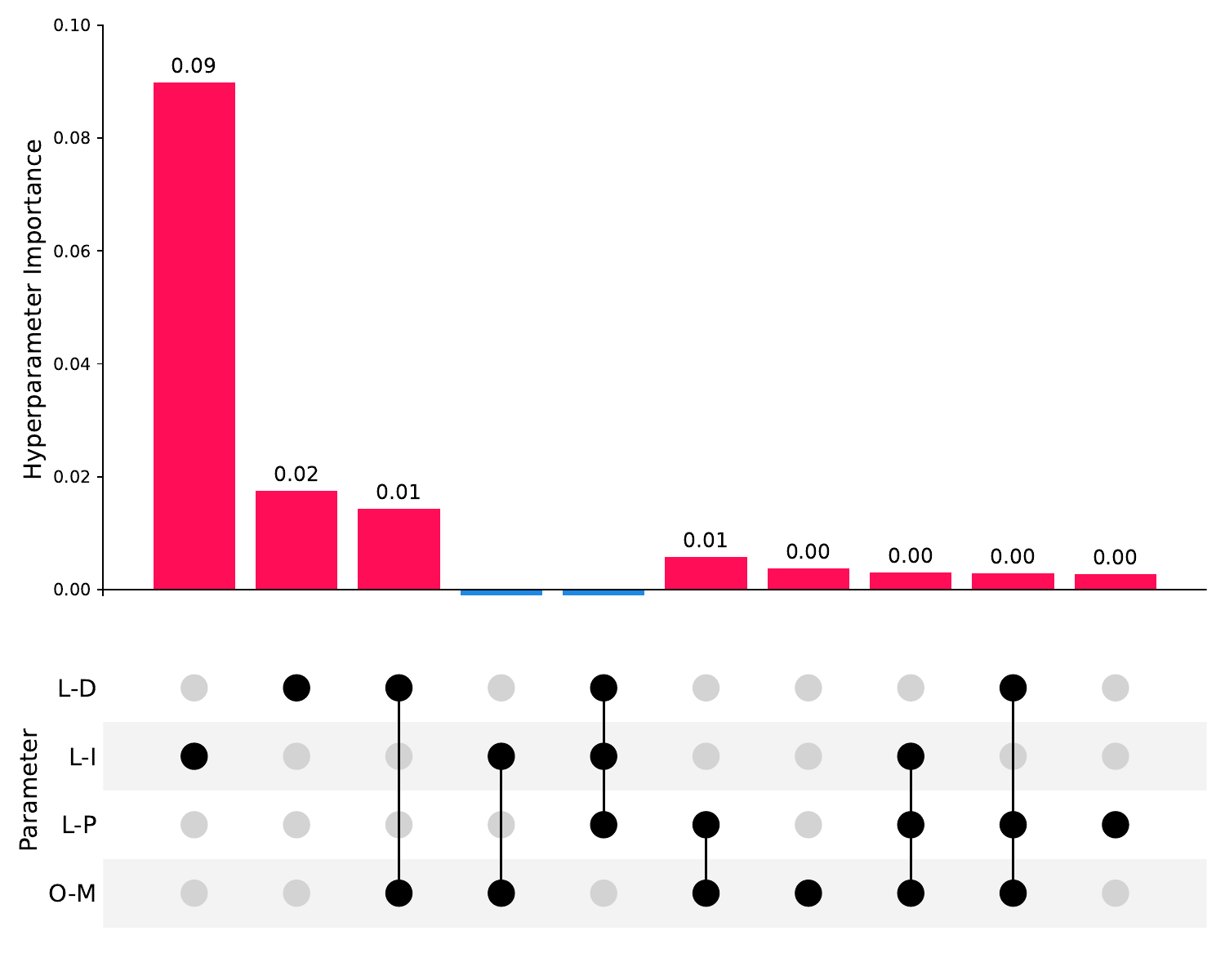}
    \end{minipage}
    \begin{minipage}[c]{0.4\textwidth}
        \includegraphics[width=\textwidth]{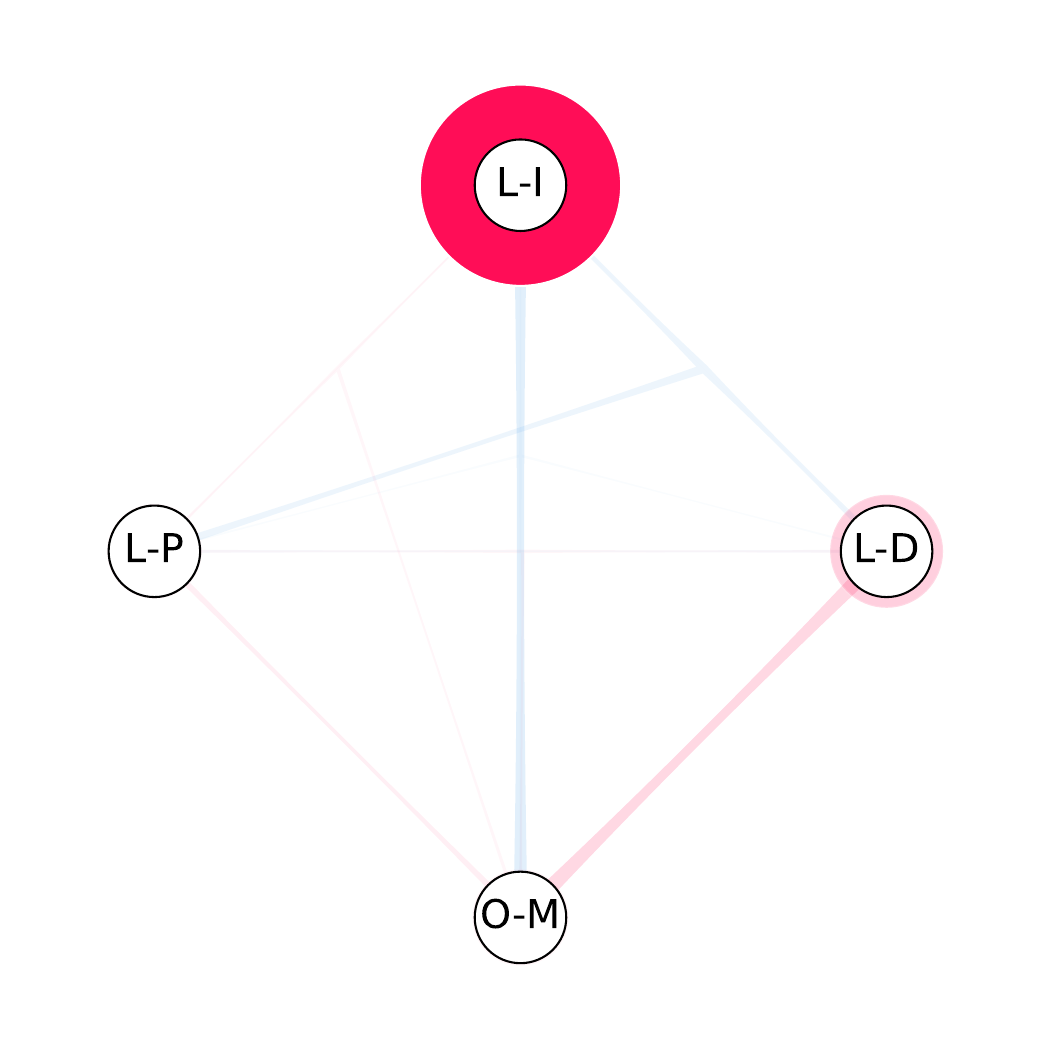}
    \end{minipage}
    \\
    \begin{minipage}[c]{0.05\textwidth}
        \rotatebox{90}{\textbf{\dstunability}}
    \end{minipage}
    \begin{minipage}[c]{0.45\textwidth}
        \includegraphics[width=\textwidth]{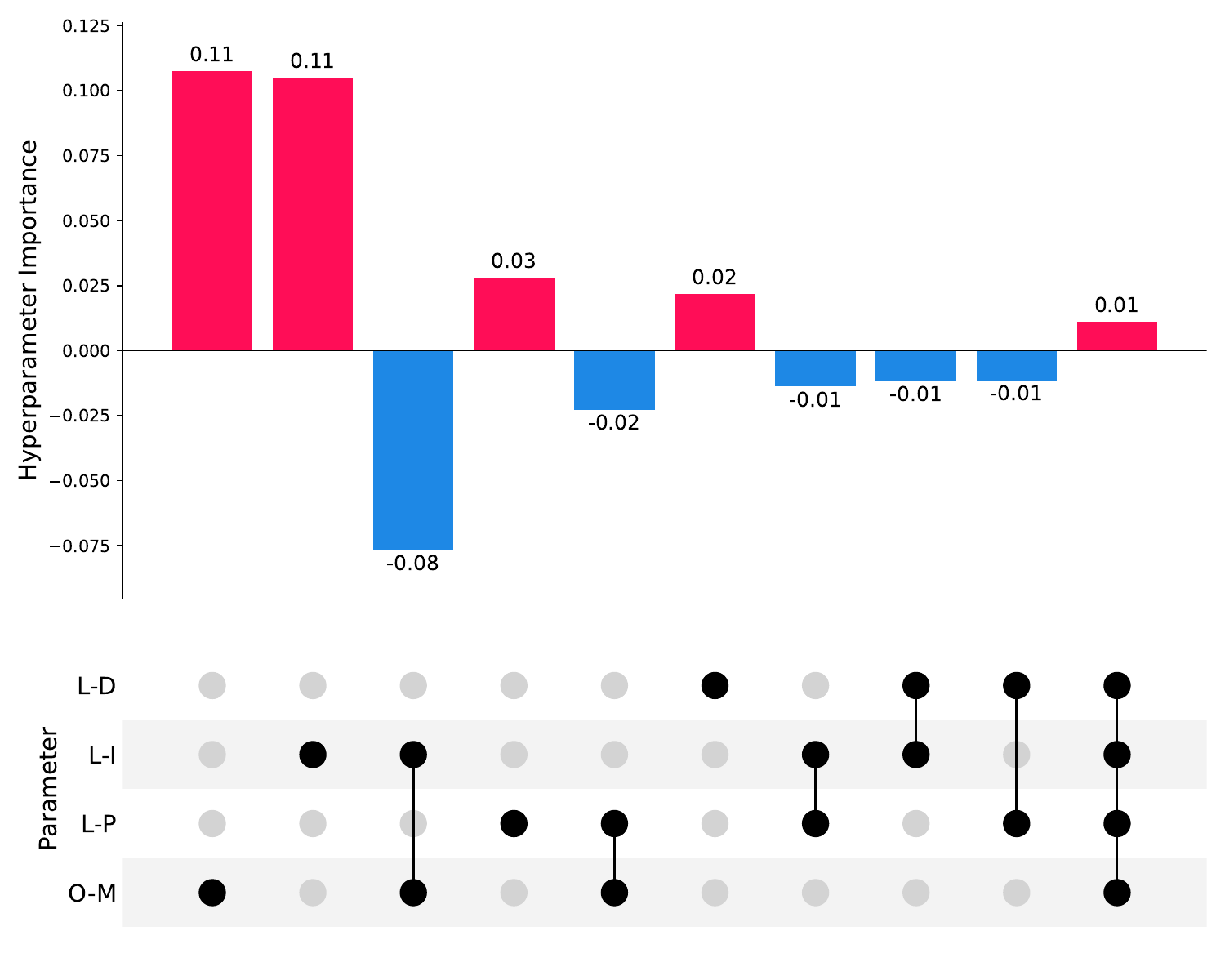}
    \end{minipage}
    \begin{minipage}[c]{0.4\textwidth}
        \includegraphics[width=\textwidth]{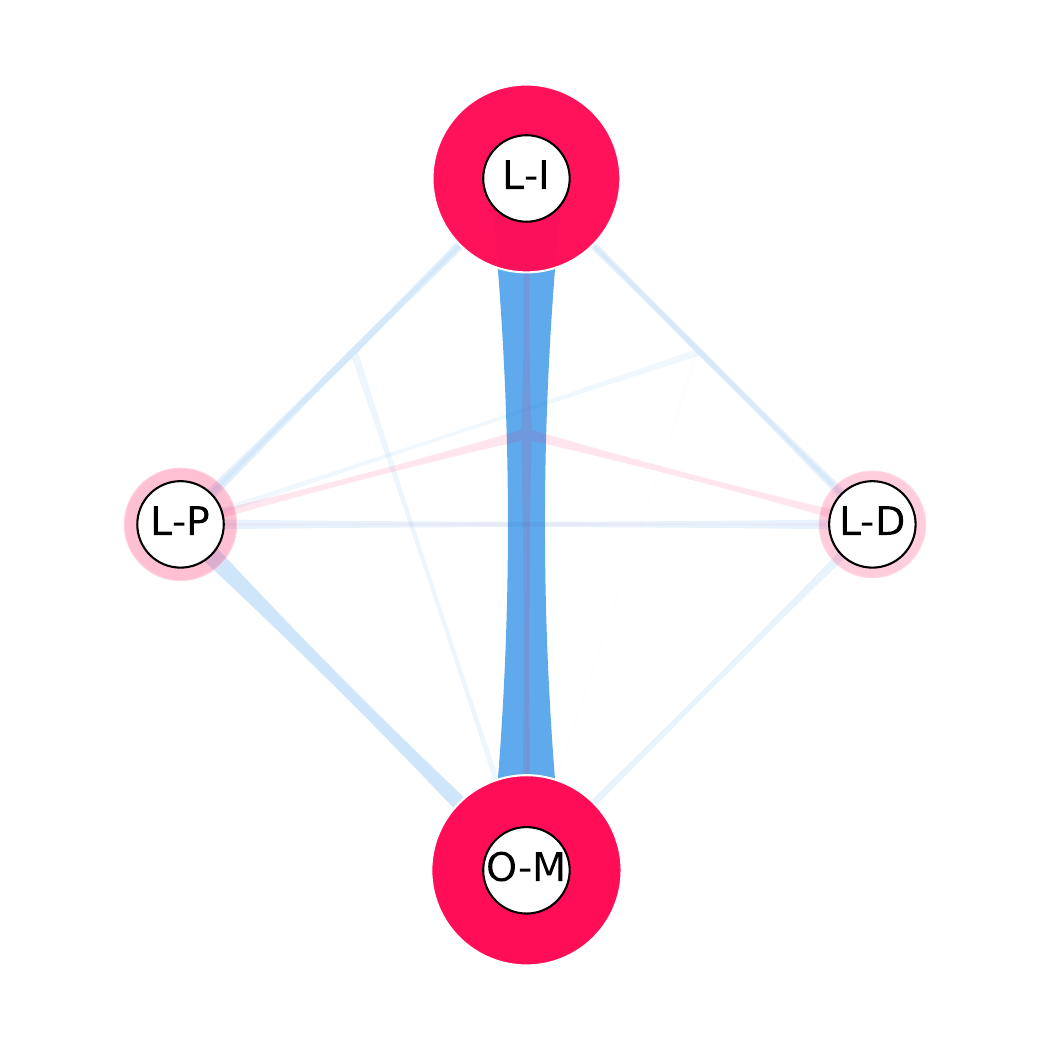}
    \end{minipage}
    \caption{UpSet (left) and SI graph (right) plots for the \ablation (top) and \dstunability (bottom) settings described in \cref{sec_experiments_ablation_tunability}. The SI graph plots show all interactions and the UpSet plot the ten most impactful interactions. }
    \label{fig_appx_ablation_tunability}
\end{figure}

In \cref{sec_experiments_ablation_tunability}, we compare the \ablation and the \dstunability settings and see that we can derive different interpretations from both explanations into the Hyperparameter optimization. 
Interpreting the \ablation explanation suggests that only the \texttt{lr\_initial} (L-I) hyperparameter is important for achieving high performance.
However, the \dstunability explanation reveals that actually both, the \texttt{opt\_momentum} (O-M) and \texttt{initial\_learning\_rate} L-I, hyperparameters are useful for tuning.
The optimizer needs to decide which hyperparameter to focus on. \cref{fig_appx_ablation_tunability} contains shows the same result as in \cref{fig_exp_ablation_tunability} with more detail.

\subsection{Additional Results for Comparison with fANOVA}\label{appx_sec_fanova_comparison}

\begin{figure}[htb]
    \centering
    \begin{minipage}[c]{0.24\textwidth}
        \includegraphics[width=\textwidth]{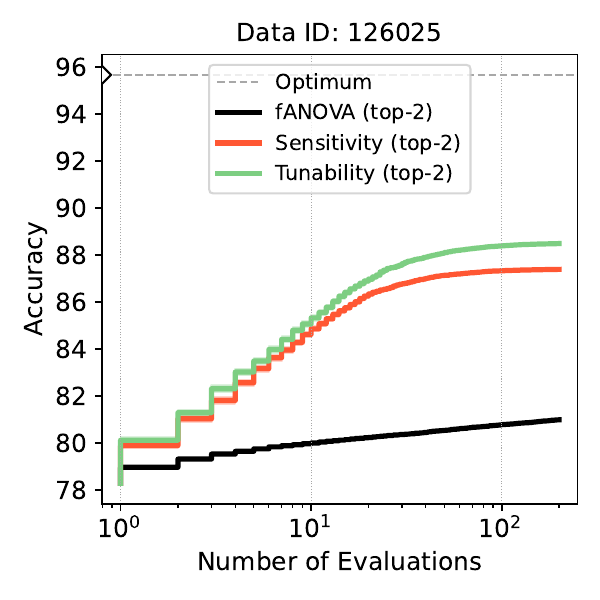}
    \end{minipage}
    \begin{minipage}[c]{0.24\textwidth}
        \includegraphics[width=\textwidth]{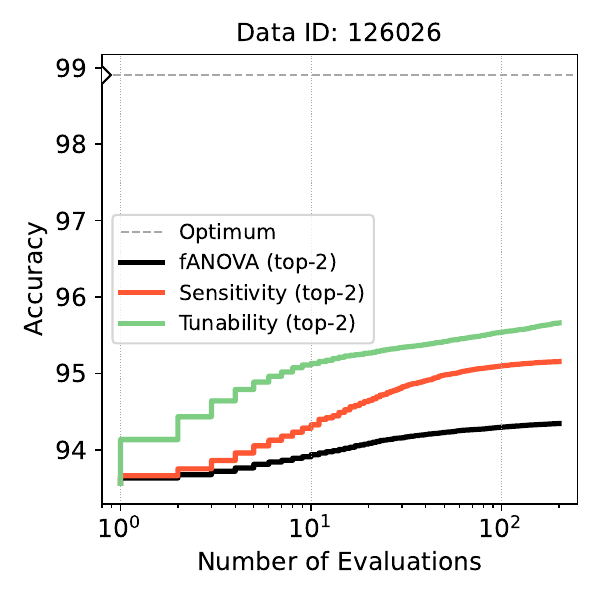}
    \end{minipage}
    \begin{minipage}[c]{0.24\textwidth}
        \includegraphics[width=\textwidth]{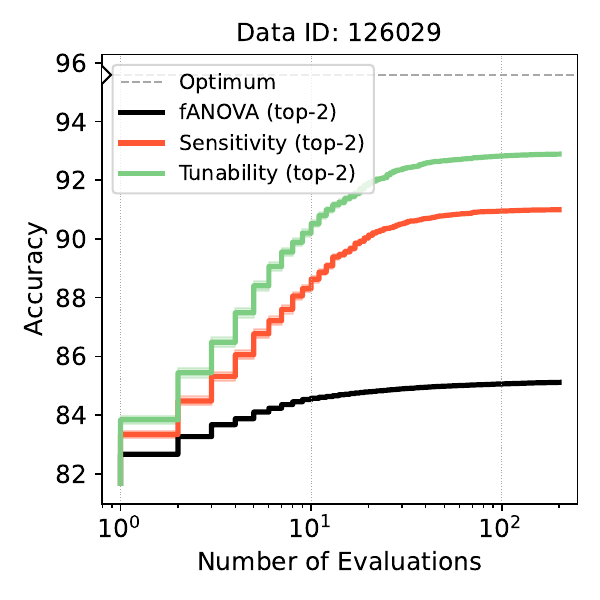}
    \end{minipage}
    \\
    \begin{minipage}[c]{0.24\textwidth}
        \includegraphics[width=\textwidth]{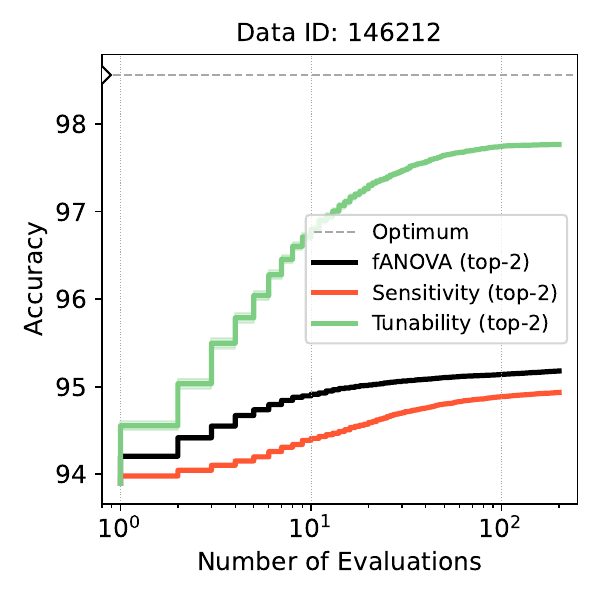}
    \end{minipage}
    \begin{minipage}[c]{0.24\textwidth}
        \includegraphics[width=\textwidth]{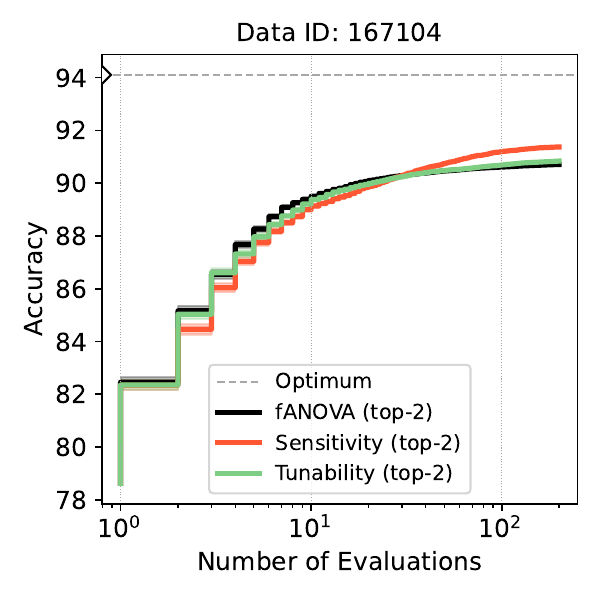}
    \end{minipage}
    \begin{minipage}[c]{0.24\textwidth}
        \includegraphics[width=\textwidth]{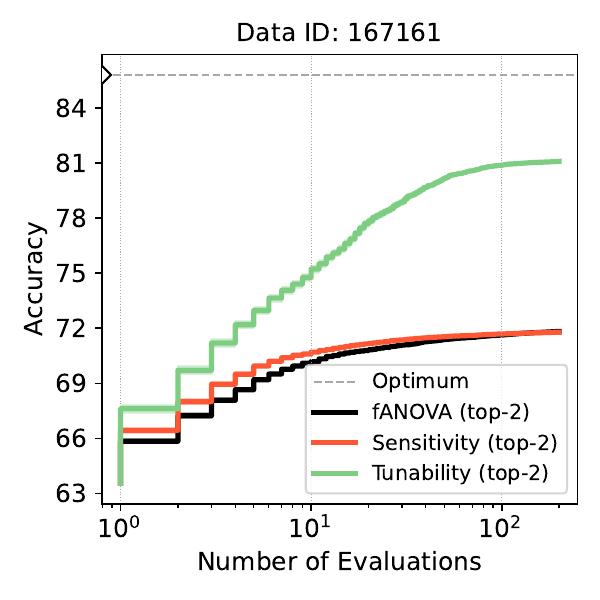}
    \end{minipage}
    \\
    \begin{minipage}[c]{0.24\textwidth}
        \includegraphics[width=\textwidth]{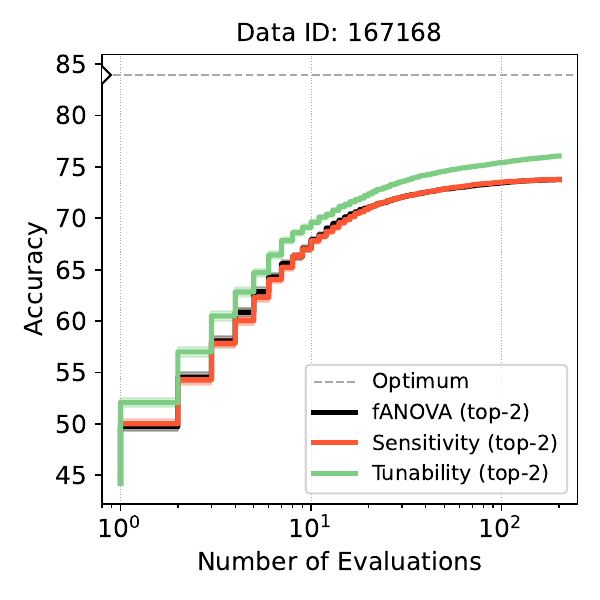}
    \end{minipage}
    \begin{minipage}[c]{0.24\textwidth}
        \includegraphics[width=\textwidth]{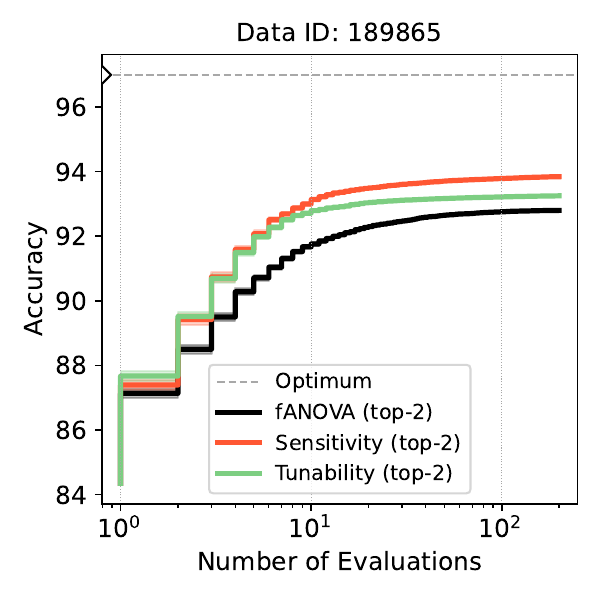}
    \end{minipage}
    \begin{minipage}[c]{0.24\textwidth}
        \includegraphics[width=\textwidth]{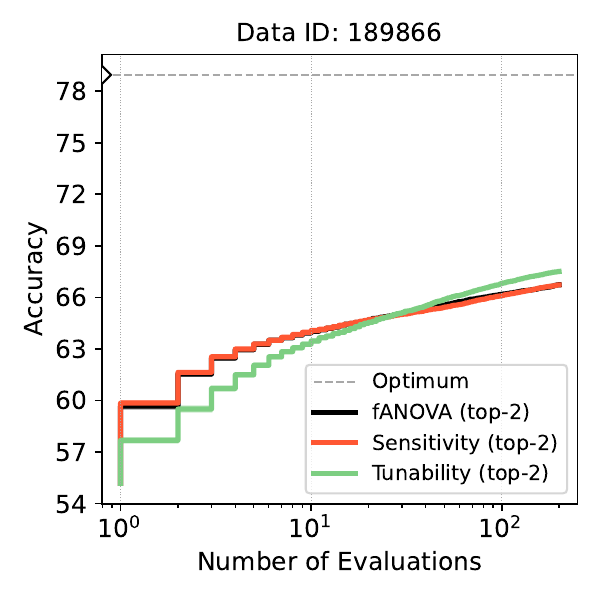}
    \end{minipage}
    \caption{Anytime performance plots showing mean and standard error of the incumbent's performance, comparing hyperparameter optimization runs restricted to the top-2 important hyperparameters as suggested by fANOVA, the sensitivity game, and the \dstunability game of \tool.}
    \label{fig_appx_fanova_comparison}
\end{figure}

This section contains additional results for the evaluation of hyperparameter optimization runs restricted to the top-2 important hyperparameters according to fANOVA \cite{fANOVA}, \sensitivity, and \dstunability of \tool. \cref{fig_appx_fanova_comparison} shows that selecting and tuning hyperparameters with \tool leads to better anytime performance than with fANOVA or \sensitivity. The suggested top-2 hyperparameters for every method are listed in \cref{tab:selected_parameters}. We can observe that overall, although not always perfect, \tool suggests a top-2 that yields higher anytime performance, meaning that the hyperparameter optimizer achieves a higher accuracy quicker. However, hyperparameters are suggested with respect to their overall hyperparameter importance, which does not necessarily guarantee better anytime performance as these hyperparameters can be more difficult to tune than others with lower impact. Still, in this case, the lower impact hyperparameters could result in better anytime performance for smaller budgets. We consider an in-depth study of which hyperparameters to suggest for which subsequent HPO task to be an interesting avenue of future work.

\begin{table}[htb]
\caption{Top-2 Hyperparamters as identified by fANOVA, Sensitivity, and \tool}\label{tab:selected_parameters}
\centering
\resizebox{\textwidth}{!}{
\begin{tabular}{l  ll  ll  ll} \toprule
\textbf{Dataset} & \multicolumn{2}{c}{\textbf{fANOVA}} & \multicolumn{2}{c}{\textbf{Sensitivity}} & \multicolumn{2}{c}{\tool}\\ \midrule
\textbf{126025}& weight\_decay & batch\_size & num\_layers & learning\_rate & num\_layers & weight\_decay \\
\textbf{126026}& momentum & learning\_rate & learning\_rate & num\_layers & weight\_decay & batch\_size \\
\textbf{126029}& batch\_size & momentum & learning\_rate & num\_layers & num\_layers & batch\_size \\
\textbf{146212}& max\_dropout & momentum & learning\_rate & max\_dropout & num\_layers & weight\_decay \\
\textbf{167104}& learning\_rate & batch\_size & learning\_rate & num\_layers & learning\_rate & max\_units \\
\textbf{167161}& learning\_rate & max\_dropout & learning\_rate & batch\_size & num\_layers & learning\_rate \\
\textbf{167168}& num\_layers & learning\_rate & learning\_rate & num\_layers & learning\_rate & max\_units \\
\textbf{189865}& num\_layers & learning\_rate & learning\_rate & batch\_size & learning\_rate & momentum \\
\textbf{189866}& num\_layers & weight\_decay & num\_layers & weight\_decay & weight\_decay & max\_units \\

\bottomrule
\end{tabular}
}
\end{table}

\subsection{Additional Results for Explaining the SMAC Surrogate During Optimization}\label{appx_sec_smac_analysis}

\begin{figure}[htb]
    \centering
    \begin{minipage}[c]{0.05\textwidth}
        \textbf{\gls*{FSII}:}
    \end{minipage}
    \begin{minipage}[c]{0.23\textwidth}
        \includegraphics[width=\textwidth]{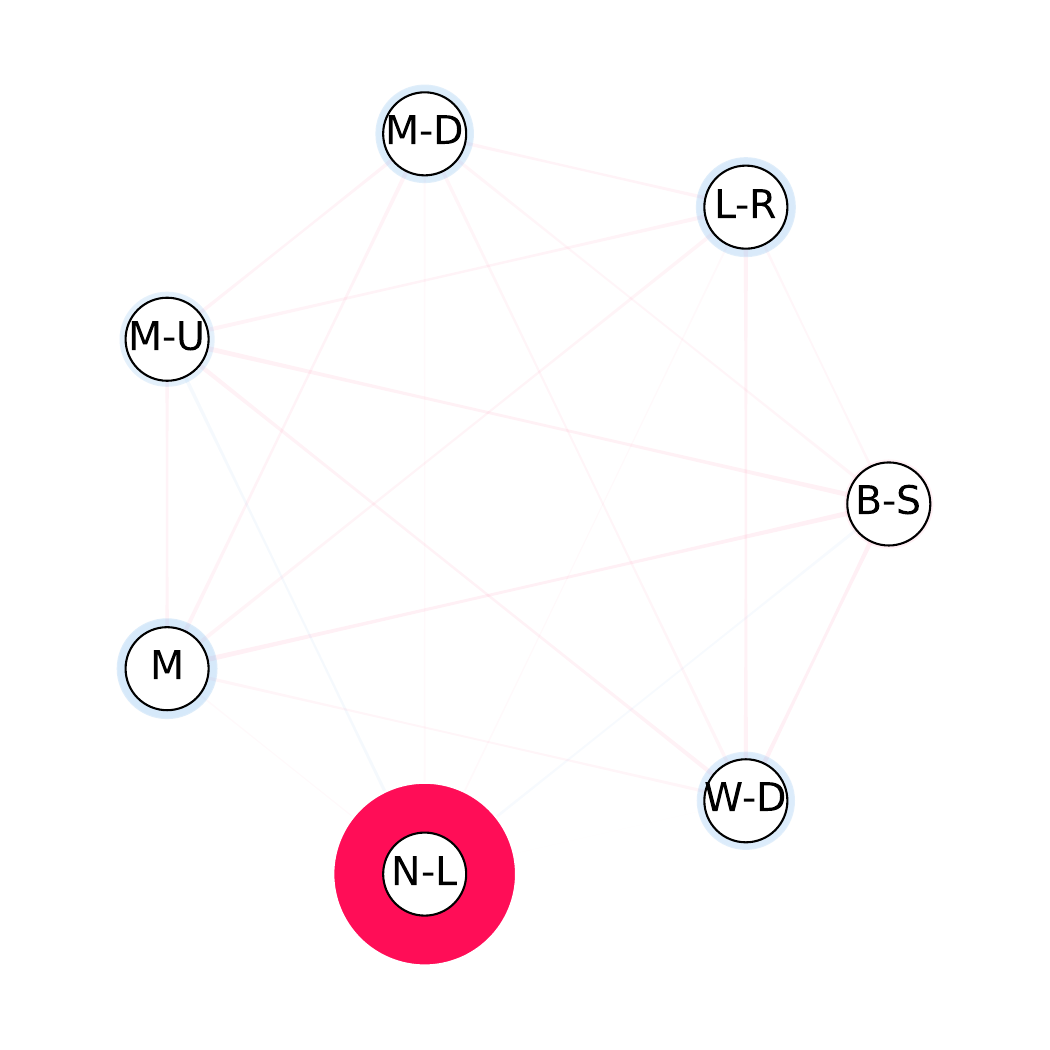}
    \end{minipage}
    \begin{minipage}[c]{0.23\textwidth}
        \includegraphics[width=\textwidth]{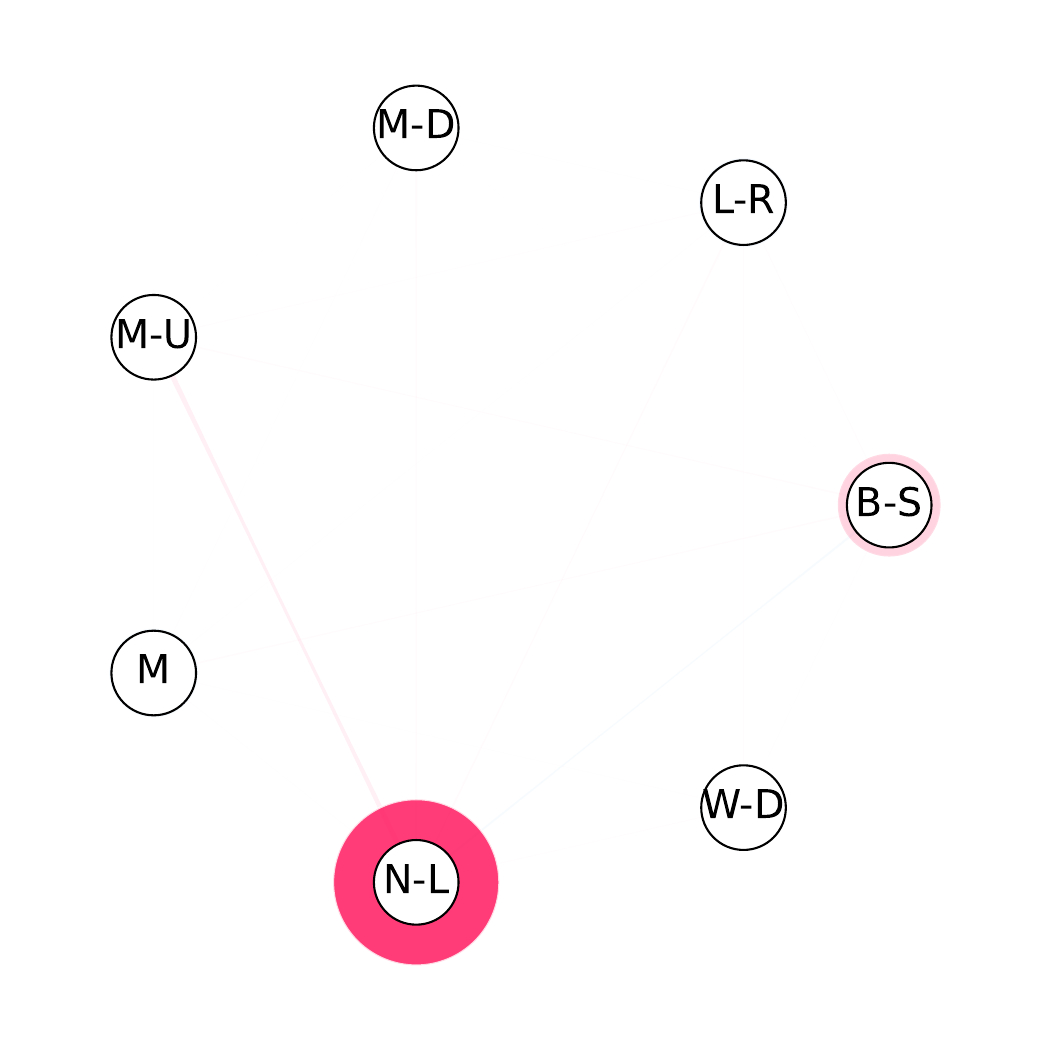}
    \end{minipage}
    \begin{minipage}[c]{0.23\textwidth}
        \includegraphics[width=\textwidth]{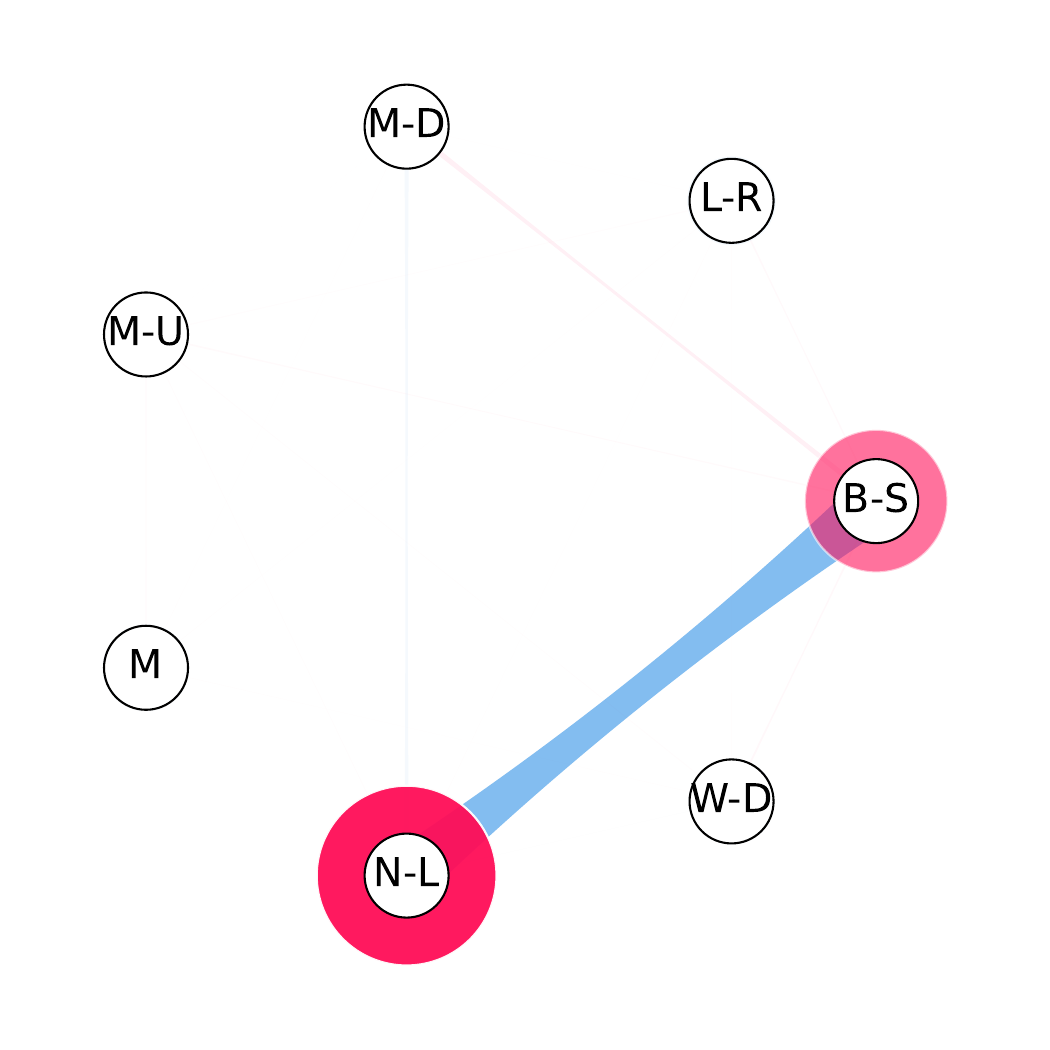}
    \end{minipage}
    \begin{minipage}[c]{0.23\textwidth}
        \includegraphics[width=\textwidth]{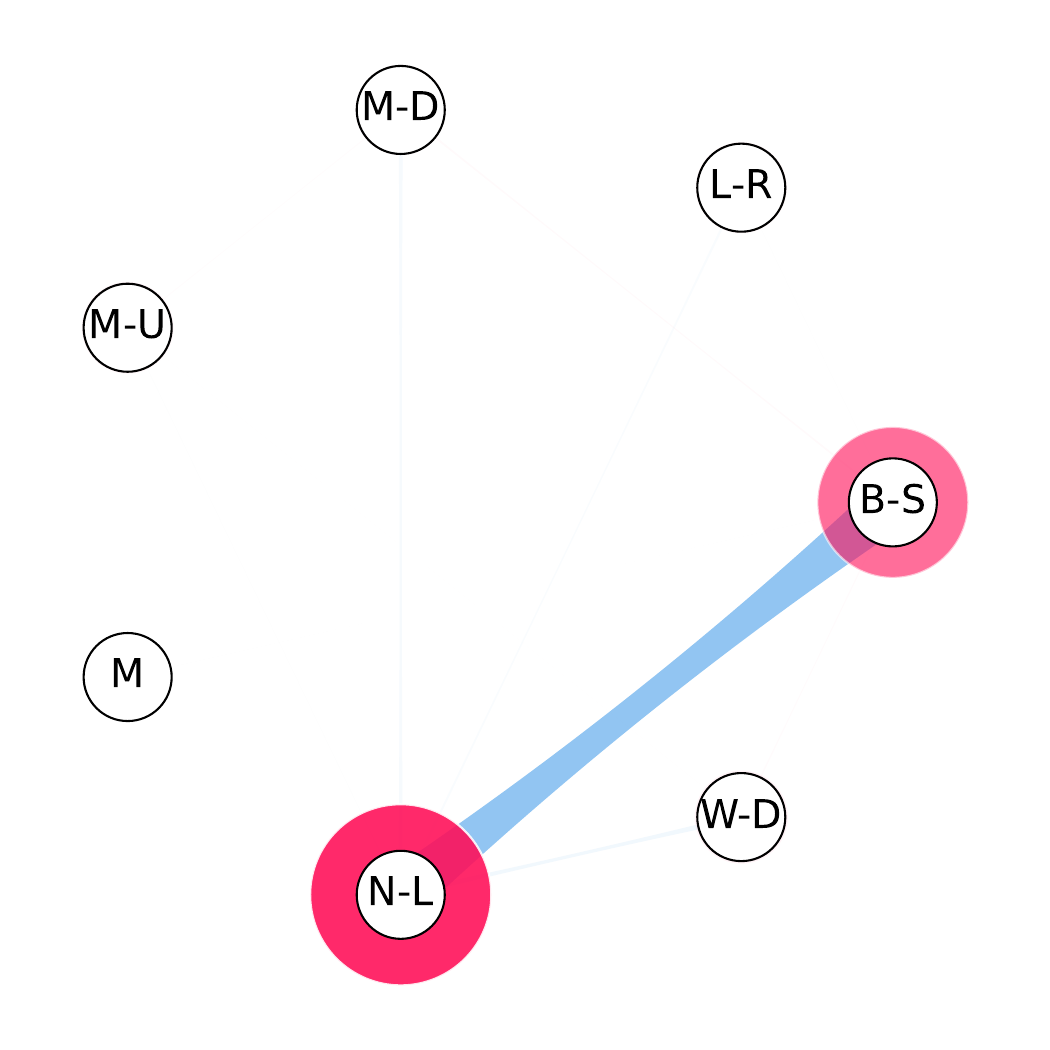}
    \end{minipage}
    \\
    \begin{minipage}[c]{0.05\textwidth}
        \textbf{\gls*{MI}:}
    \end{minipage}
    \begin{minipage}[c]{0.23\textwidth}
        \includegraphics[width=\textwidth]{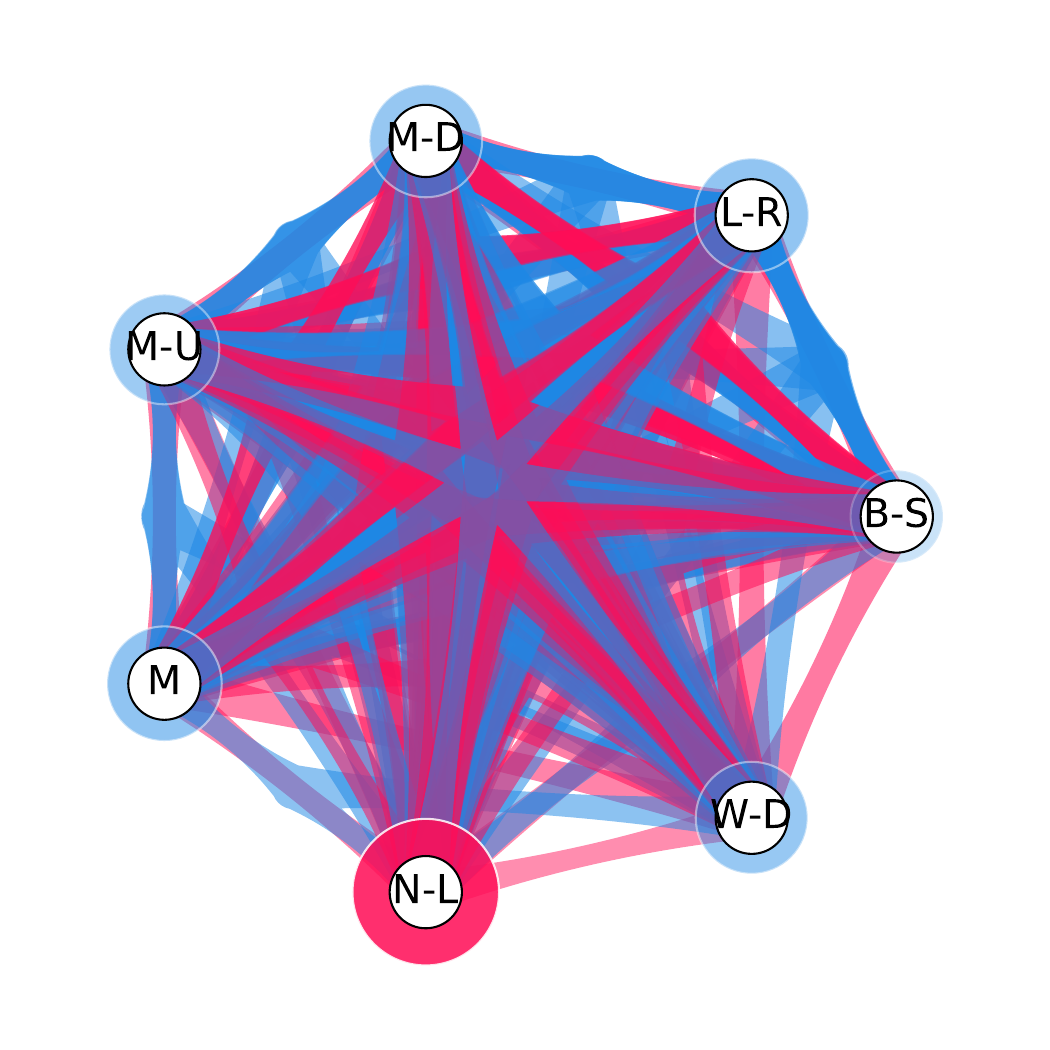}
    \end{minipage}
    \begin{minipage}[c]{0.23\textwidth}
        \includegraphics[width=\textwidth]{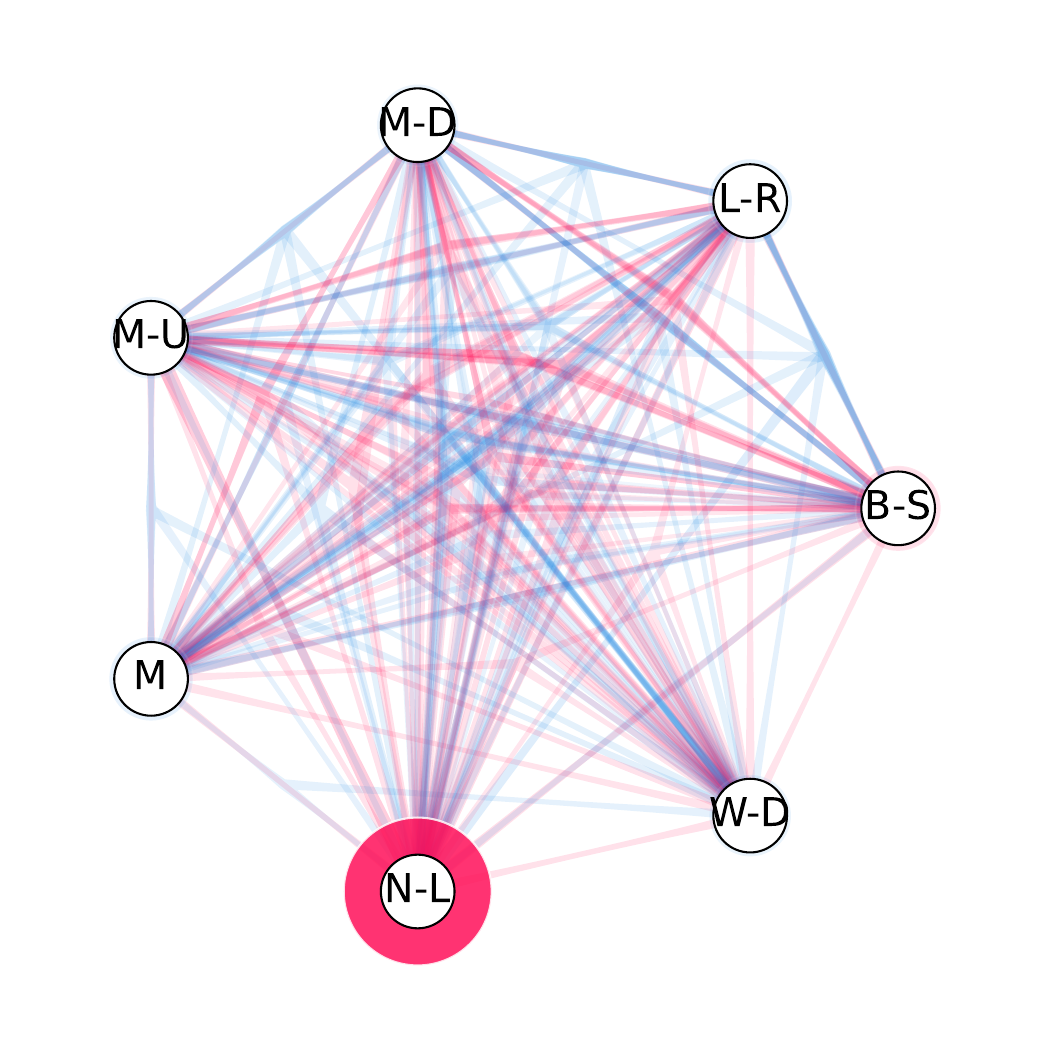}
    \end{minipage}
    \begin{minipage}[c]{0.23\textwidth}
        \includegraphics[width=\textwidth]{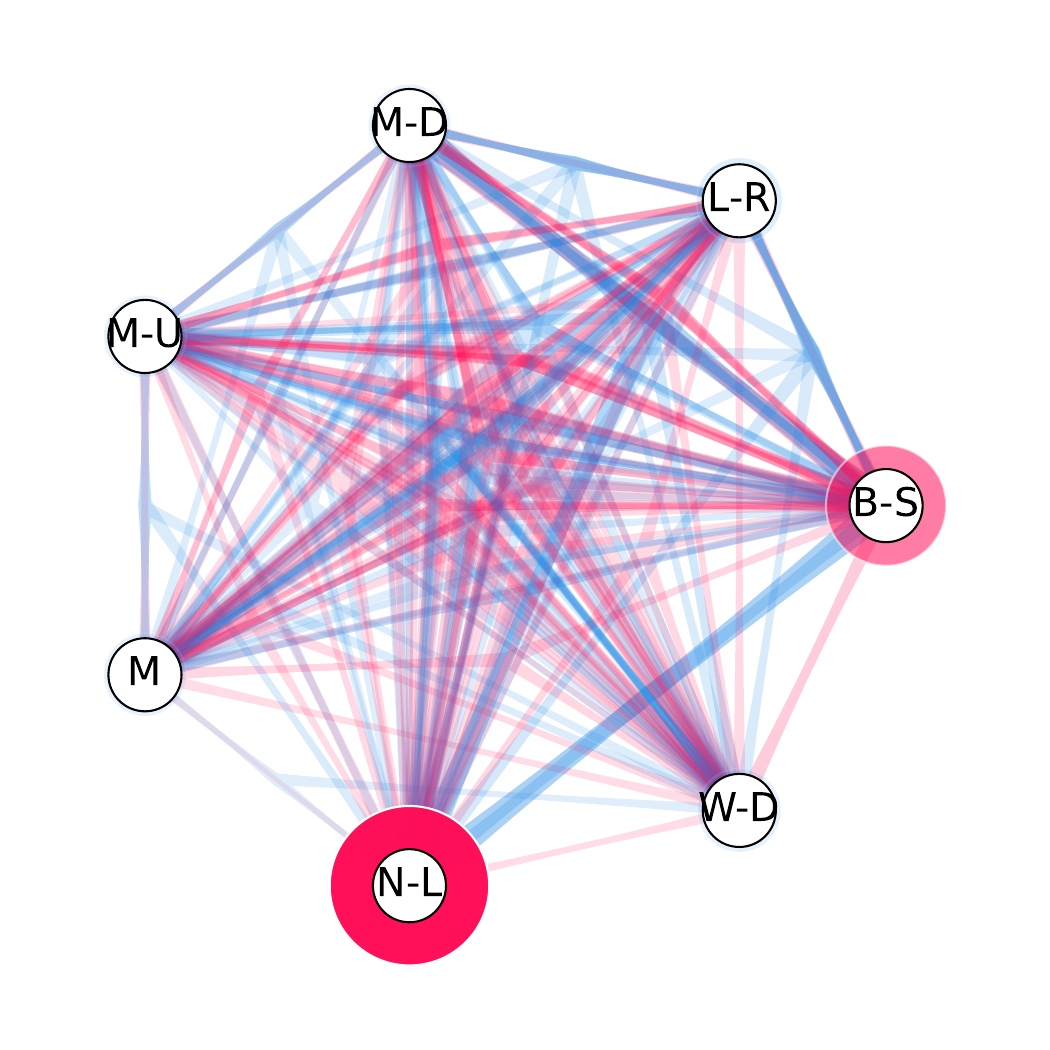}
    \end{minipage}
    \begin{minipage}[c]{0.23\textwidth}
        \includegraphics[width=\textwidth]{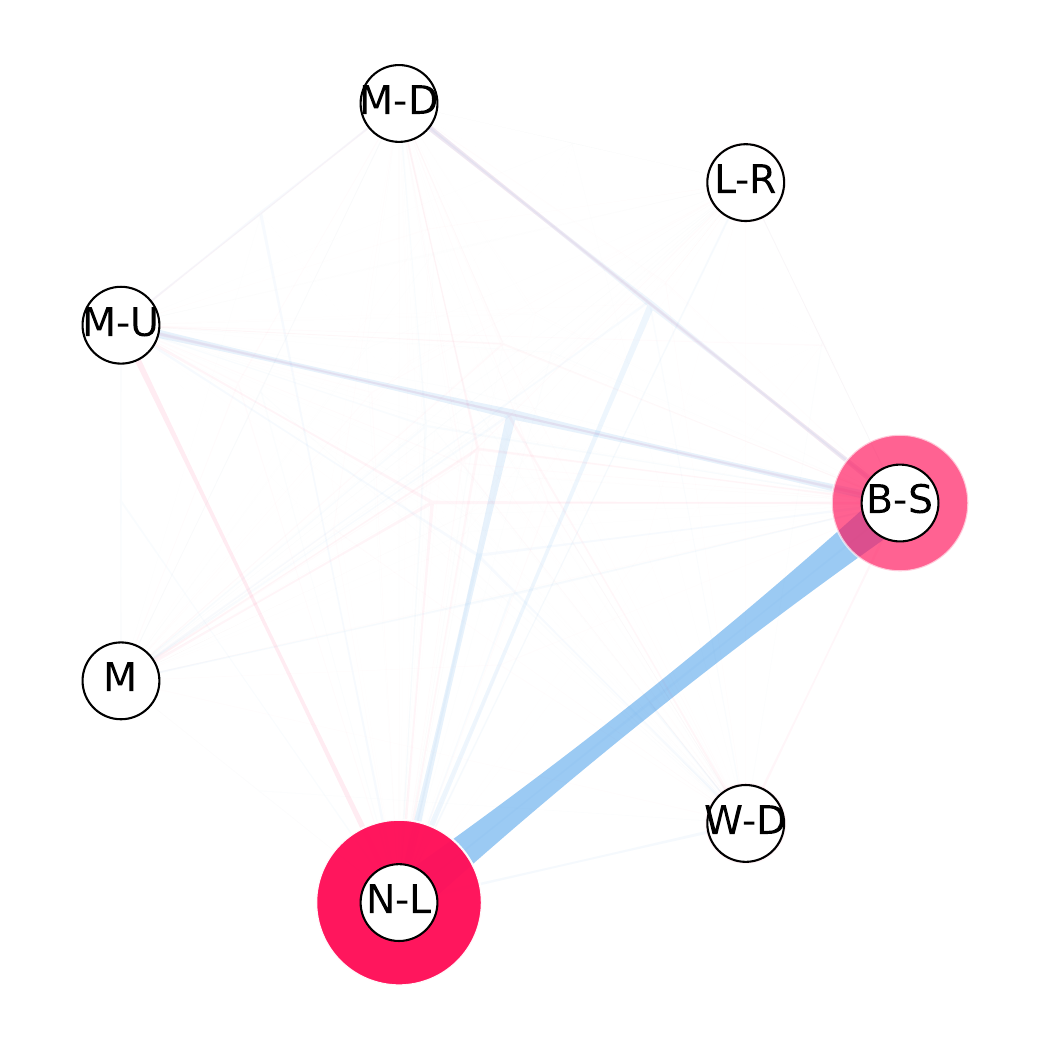}
    \end{minipage}
    \\
    \begin{minipage}[c]{0.05\textwidth}
        \textbf{Budget}: 
    \end{minipage}
    \begin{minipage}[c]{0.23\textwidth}
        \centering\textbf{60}\\(1\%)      
    \end{minipage}
    \begin{minipage}[c]{0.23\textwidth}
        \centering\textbf{300}\\(5\%)
    \end{minipage}
    \begin{minipage}[c]{0.23\textwidth}
        \centering\textbf{1500}\\(25\%)
    \end{minipage}
    \begin{minipage}[c]{0.23\textwidth}
        \centering\textbf{6000}\\(100\%)
    \end{minipage}
    \caption{\tool \dstunability explanations for the surrogate model used in SMAC at different time intervals (1\%, 5\%, 25\%, 100\%) of the optimization procedure for dataset 3945 of \texttt{lcbench} \cite{zimmer-tpami21a}. Over time the model becomes less uncertain about which hyperparameters are important to achieve a high predictive performance. Bottom: Interaction graphs for Moebius Interactions (MI) show all pure main effects and interactions. Top: Higher-order interactions are summarized to main effects and second-order interactions, summarizing the game properly already at early stages when the MI still shows a comparably large number of higher-order interactions.}
    \label{fig_appx_smac_analysis_surrogate}
\end{figure}

In \cref{fig_appx_smac_analysis_surrogate}, in addition to the MI interaction graphs, we summarize explanations with the help of second order \gls*{FSII}, which fairly distributes higher-order interactions to the lower orders, here order one and two. We find that with FSII we can distill the relevant parts of the \glspl*{MI} quite clearly.

\subsection{Additional Interaction Visualizations}

\begin{figure}
    \begin{minipage}[c]{0.1\textwidth}
        \rotatebox{90}{\texttt{imagenet\_resnet}}
    \end{minipage}
    \begin{minipage}[c]{0.40\textwidth}
        \centering
        \textbf{\ablation}
        \includegraphics[width=.9\linewidth]{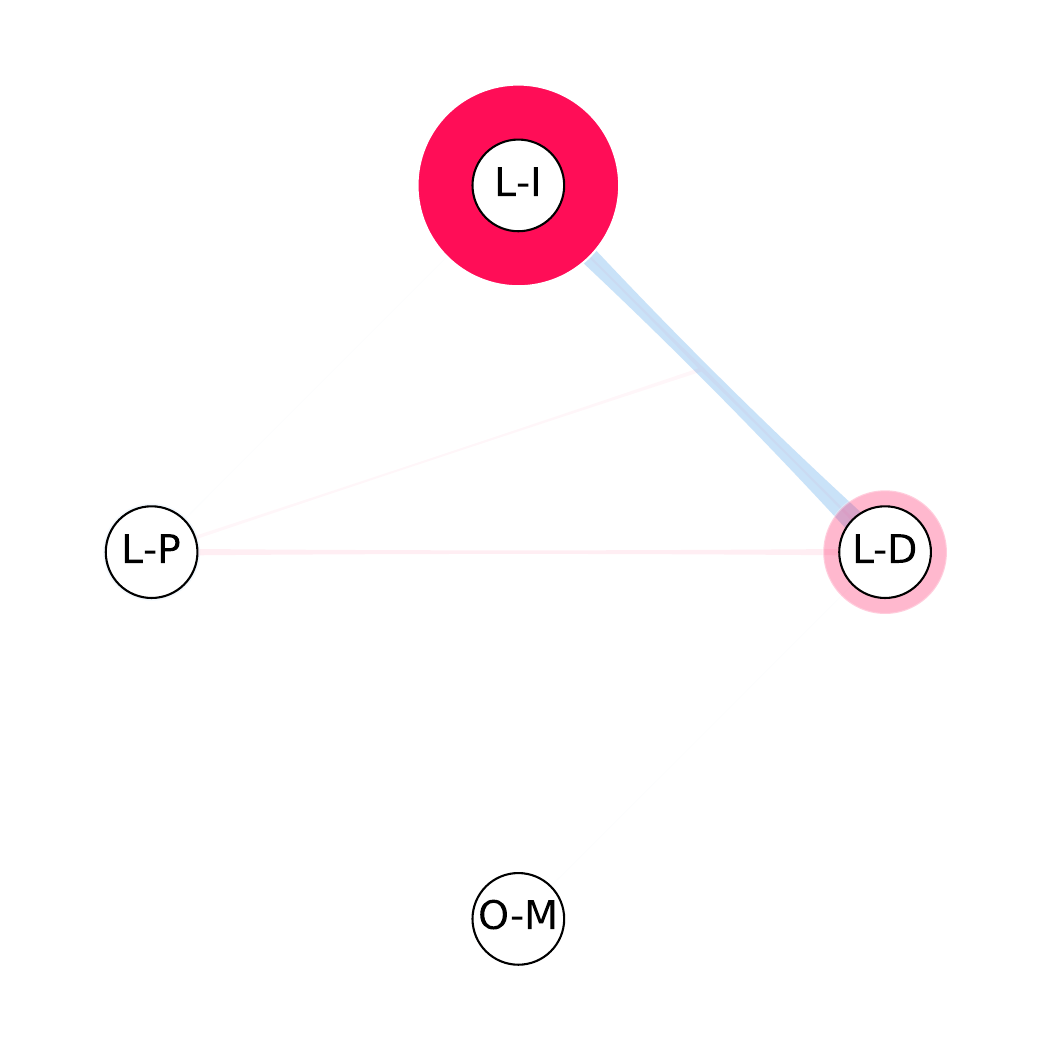}
    \end{minipage}
    \hfill
    \begin{minipage}[c]{0.40\textwidth}
        \centering
        \textbf{\dstunability}
        \includegraphics[width=.9\linewidth]{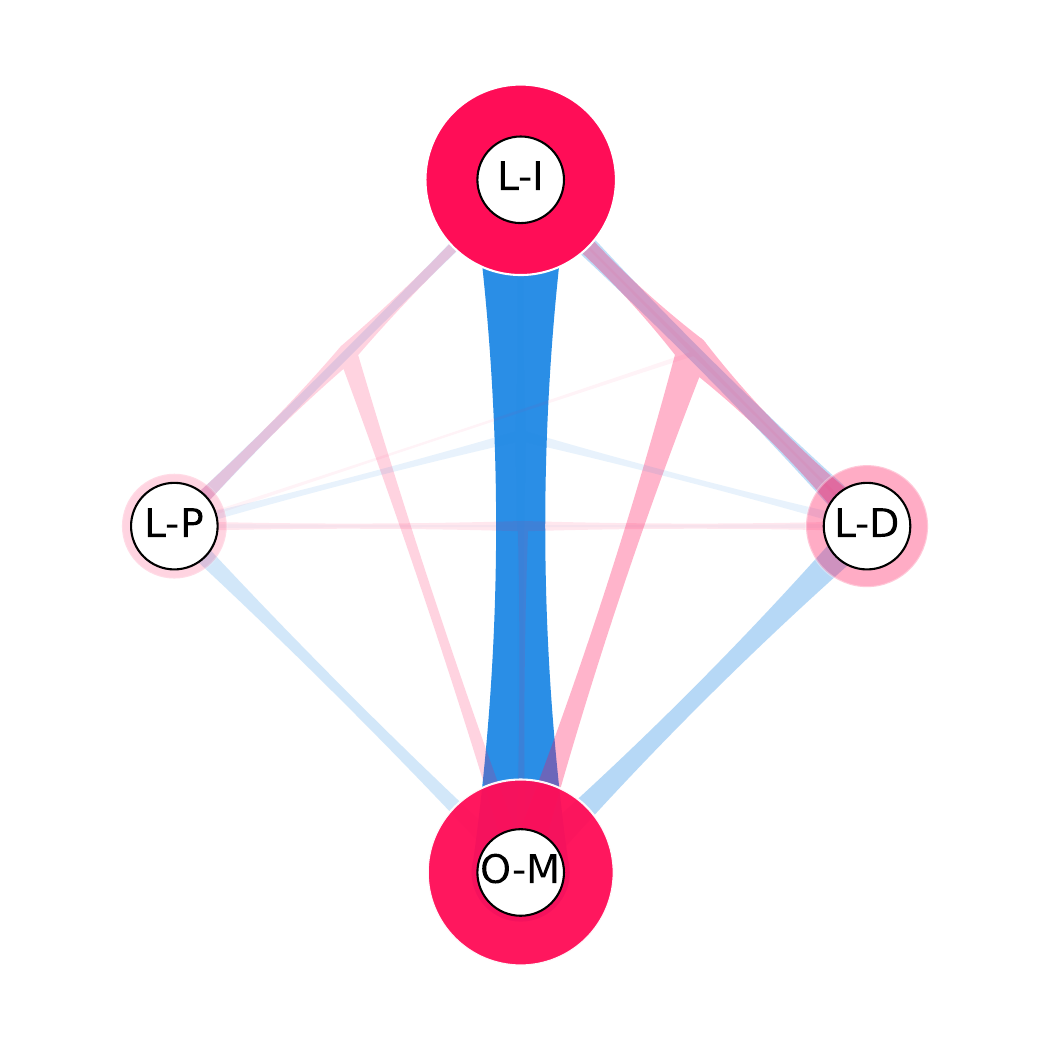}
    \end{minipage}
    \\
    \begin{minipage}[c]{0.1\textwidth}
        \rotatebox{90}{\texttt{lm1b\_transformer}}
    \end{minipage}
    \begin{minipage}[c]{0.40\textwidth}
        \centering
        \includegraphics[width=.9\linewidth]{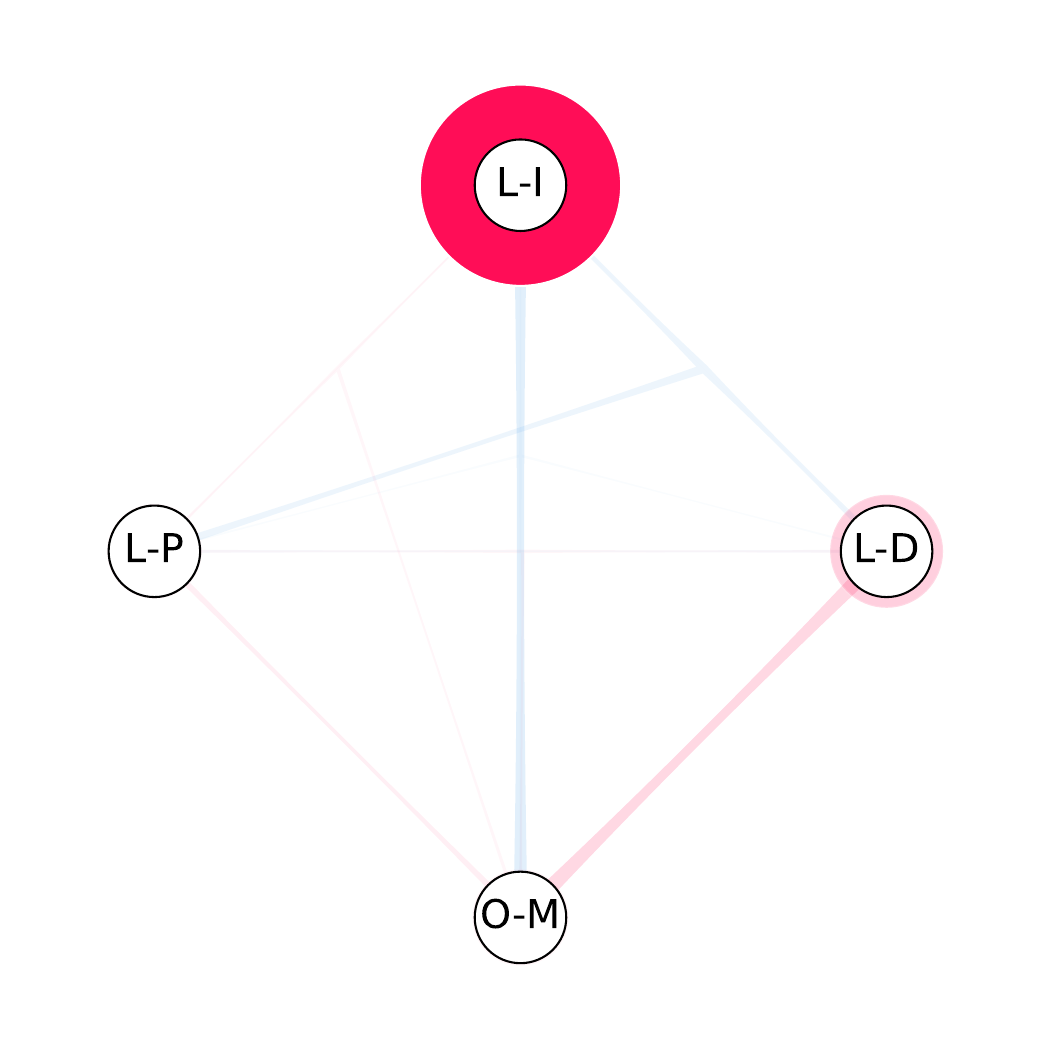}
    \end{minipage}
    \hfill
    \begin{minipage}[c]{0.40\textwidth}
        \centering
        \includegraphics[width=.9\linewidth]{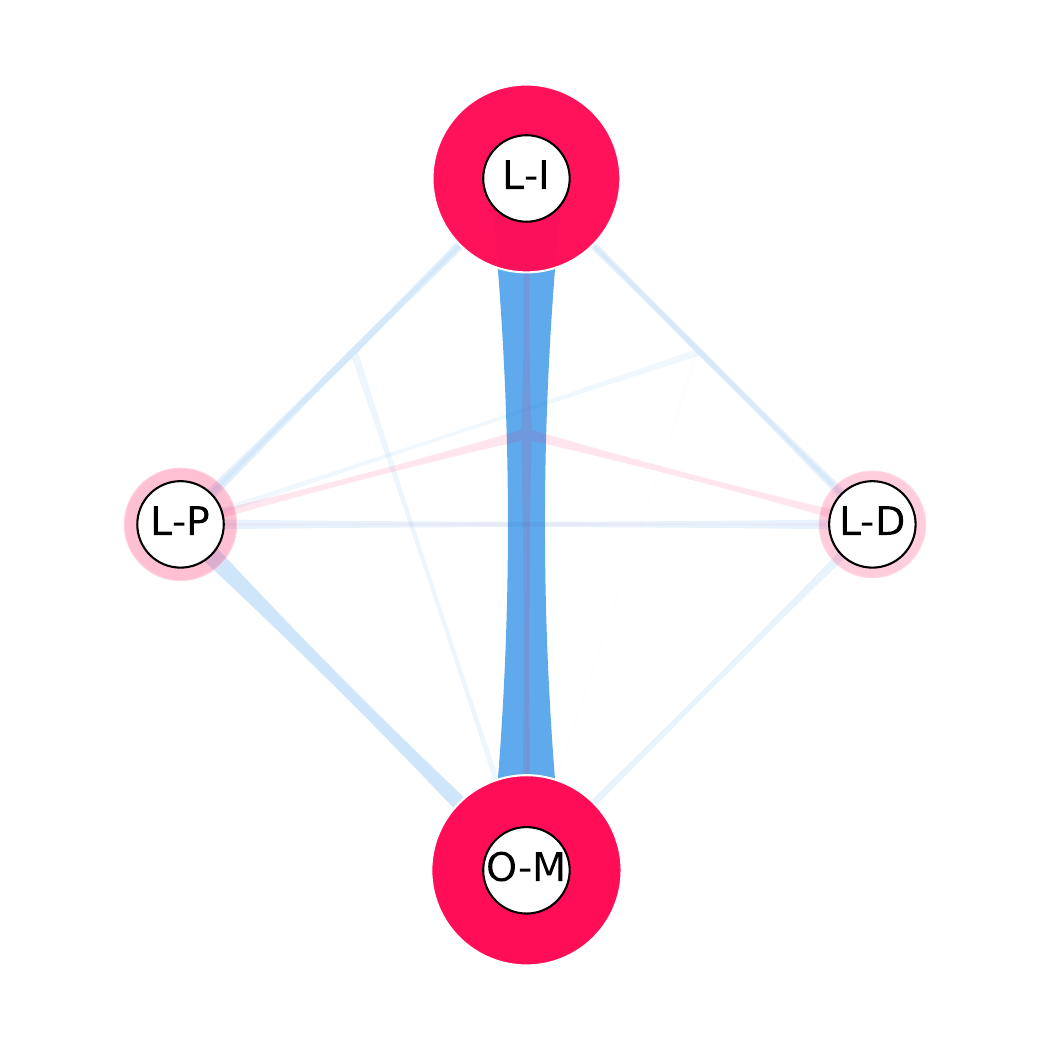}
    \end{minipage}
    \\
    \begin{minipage}[c]{0.1\textwidth}
        \rotatebox{90}{\texttt{translatewmt\_xformer}}
    \end{minipage}
    \begin{minipage}[c]{0.40\textwidth}
        \centering
        \includegraphics[width=.9\linewidth]{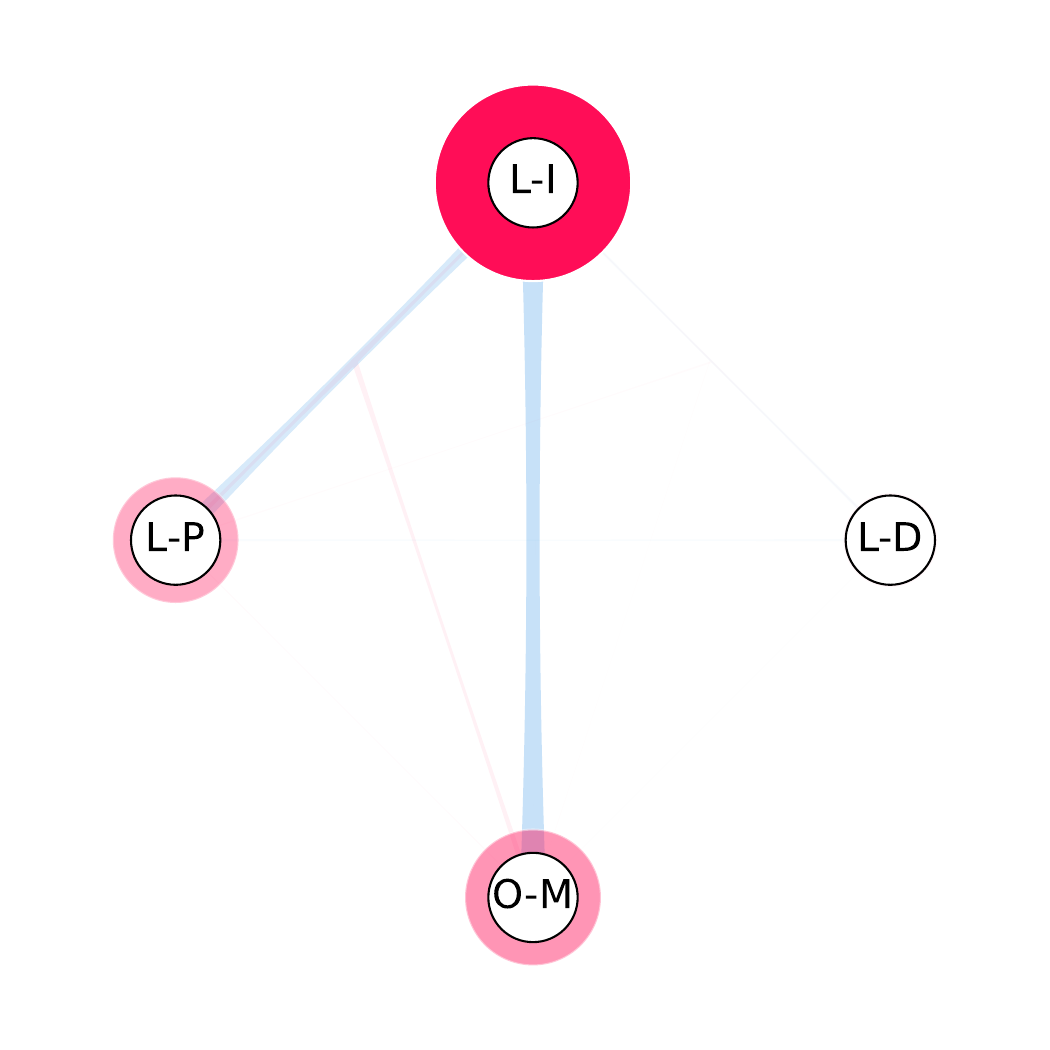}
    \end{minipage}
    \hfill
    \begin{minipage}[c]{0.40\textwidth}
        \centering
        \includegraphics[width=.9\linewidth]{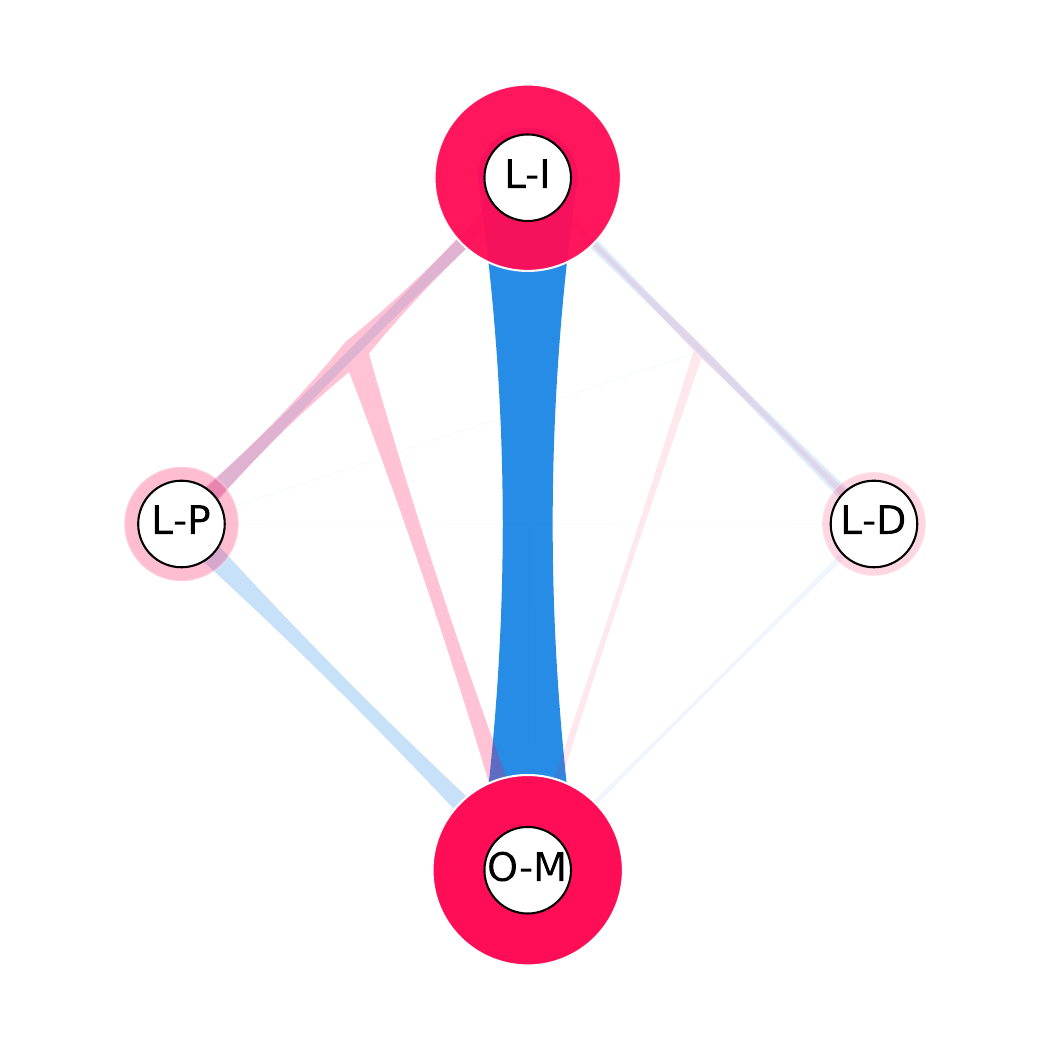}
    \end{minipage}
    \caption{\edit{MIs as computed via \tool for three different scenarios of \pdone, considering hyperparameter optimization for image classifiers and transformers.}}
    \label{fig_appx_pd1}
\end{figure}

\begin{figure}
    \begin{minipage}[c]{0.1\textwidth}
        \rotatebox{90}{\texttt{CIFAR10}}
    \end{minipage}
    \begin{minipage}[c]{0.40\textwidth}
        \centering
        \textbf{\ablation}
        \includegraphics[width=.9\linewidth]{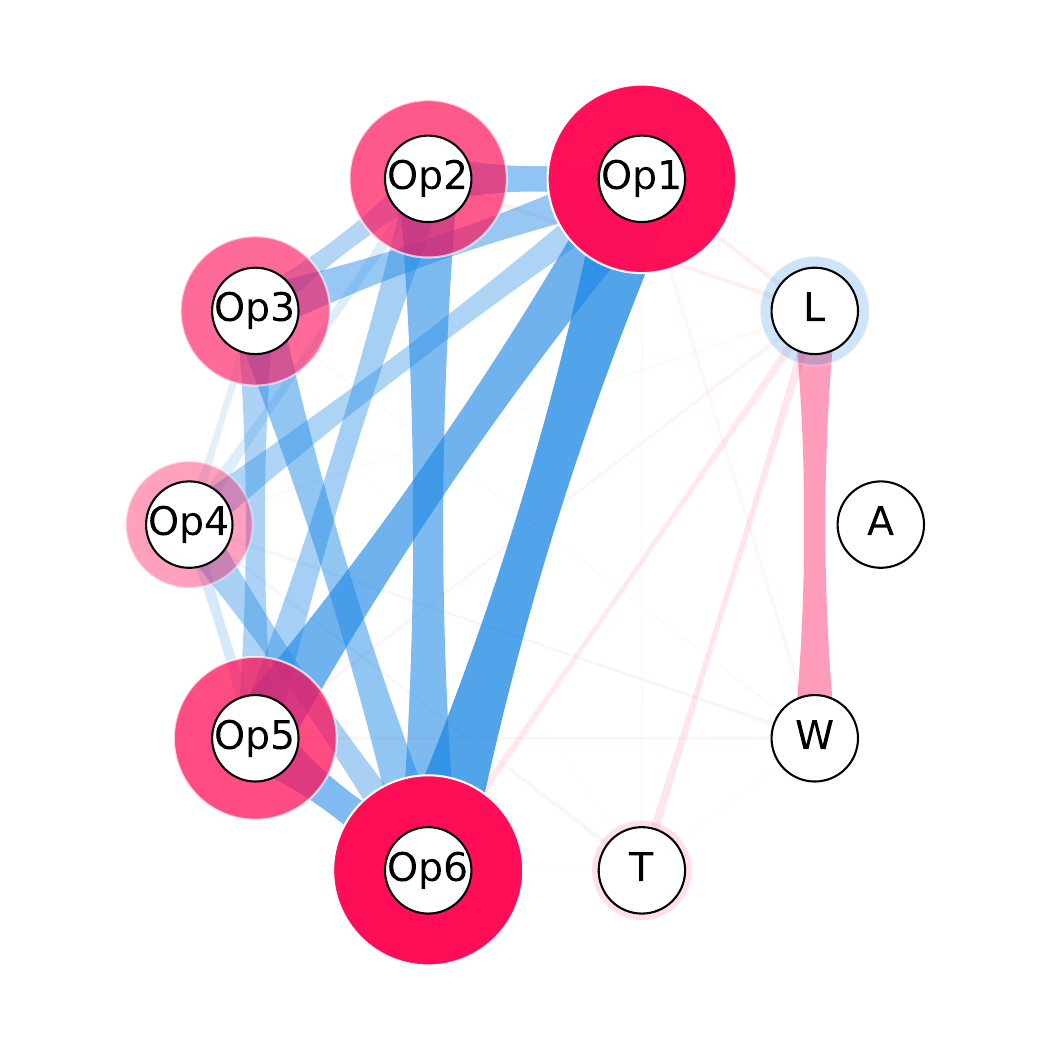}
    \end{minipage}
    \hfill
    \begin{minipage}[c]{0.40\textwidth}
        \centering
        \textbf{\dstunability}
        \includegraphics[width=.9\linewidth]{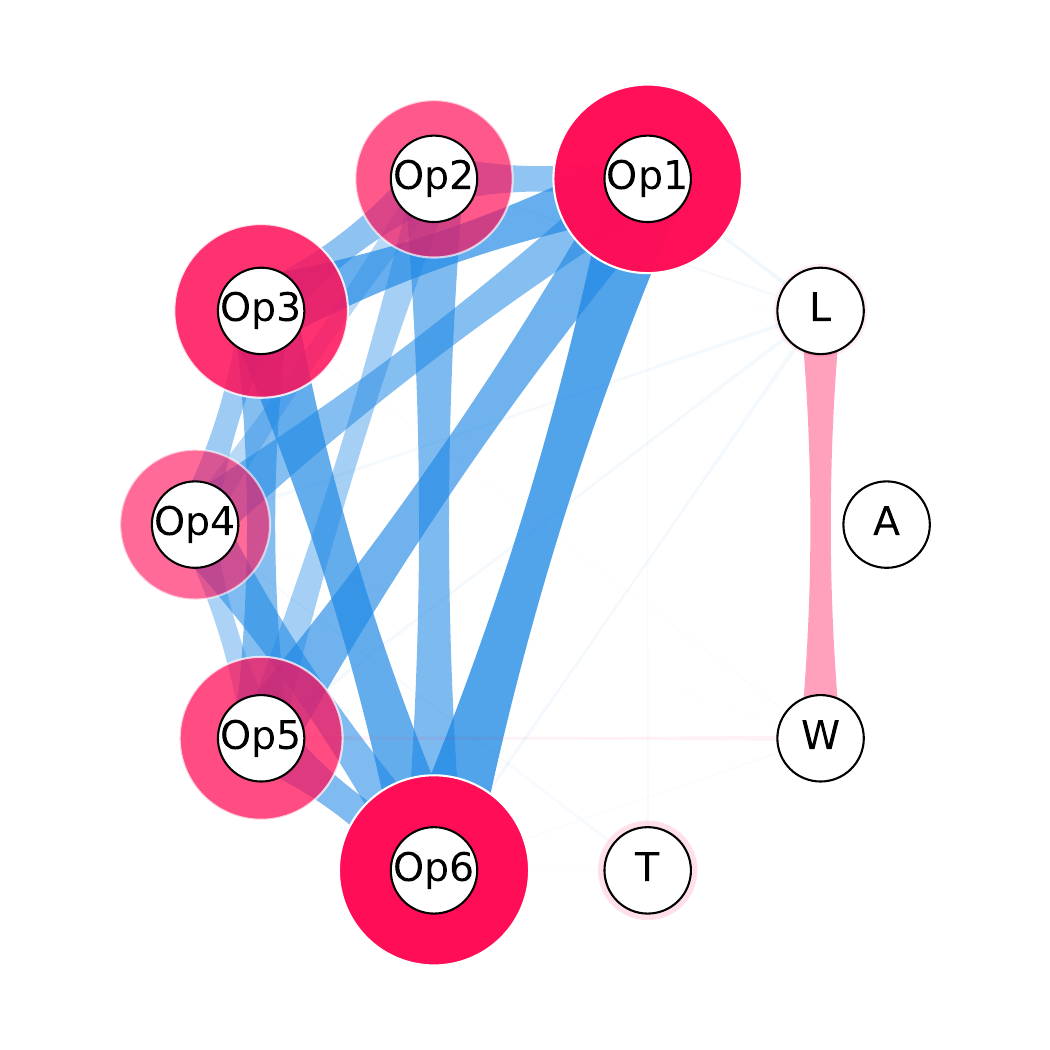}
    \end{minipage}
    \\
    \begin{minipage}[c]{0.1\textwidth}
        \rotatebox{90}{\texttt{FashionMNIST}}
    \end{minipage}
    \begin{minipage}[c]{0.40\textwidth}
        \centering
        \includegraphics[width=.9\linewidth]{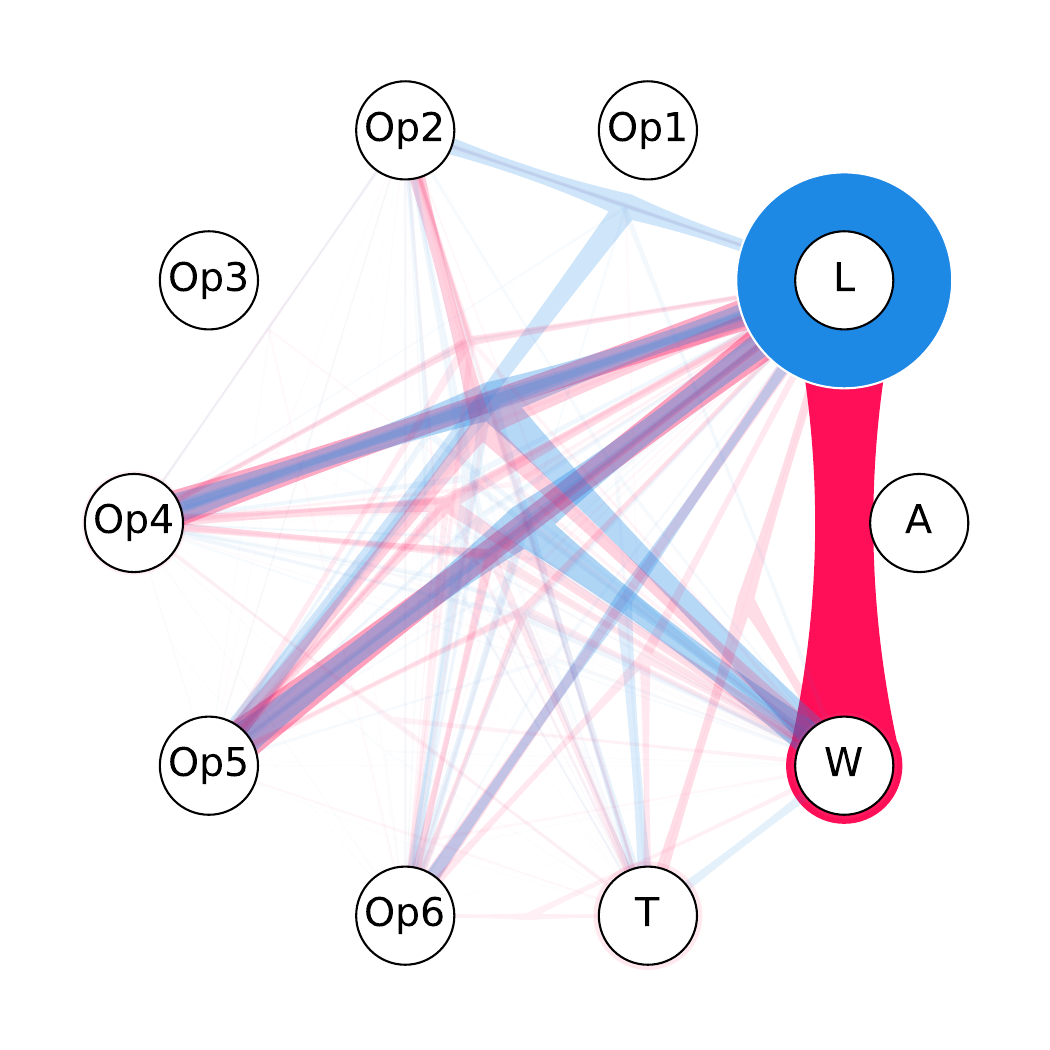}
    \end{minipage}
    \hfill
    \begin{minipage}[c]{0.40\textwidth}
        \centering
        \includegraphics[width=.9\linewidth]{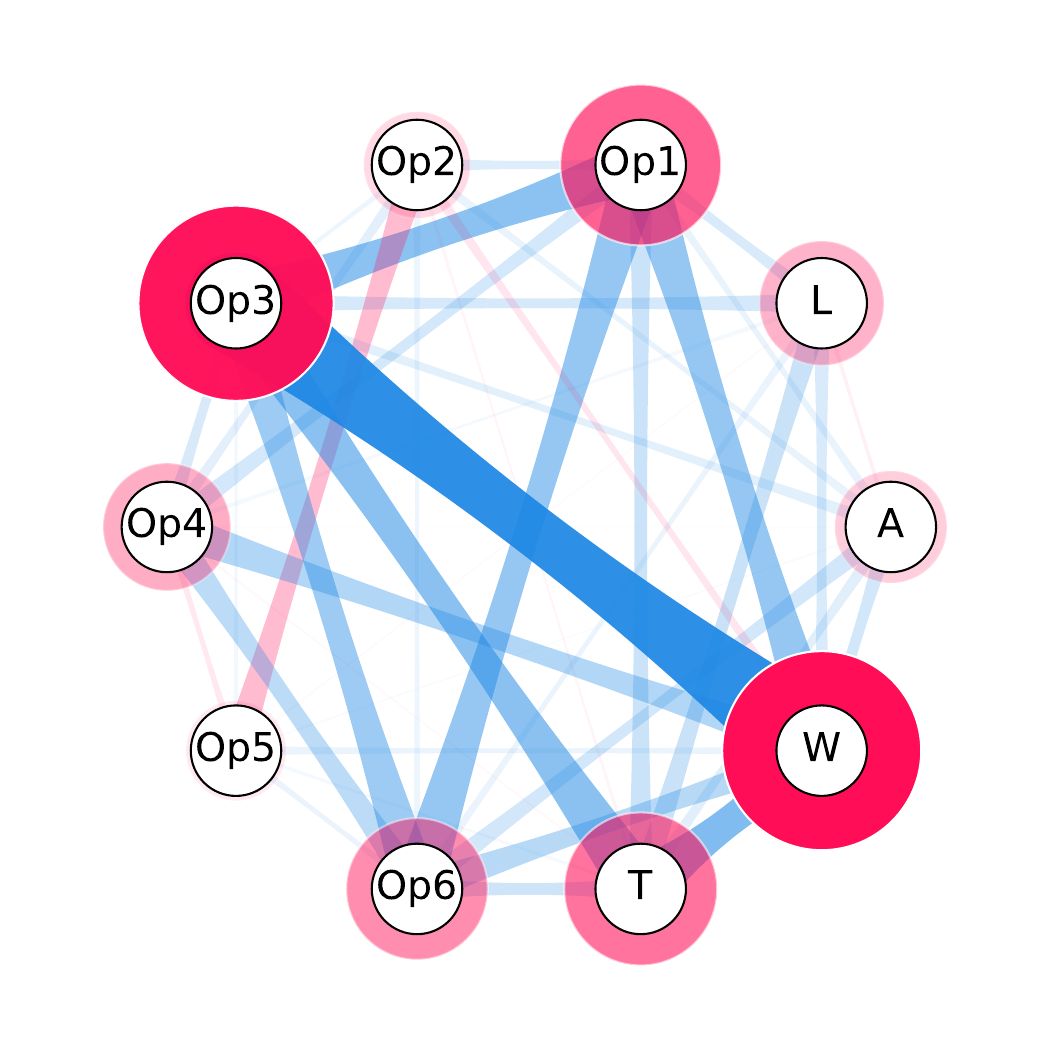}
    \end{minipage}
    \\
    \begin{minipage}[c]{0.1\textwidth}
        \rotatebox{90}{\texttt{ColorectalHistology}}
    \end{minipage}
    \begin{minipage}[c]{0.40\textwidth}
        \centering
        \includegraphics[width=.9\linewidth]{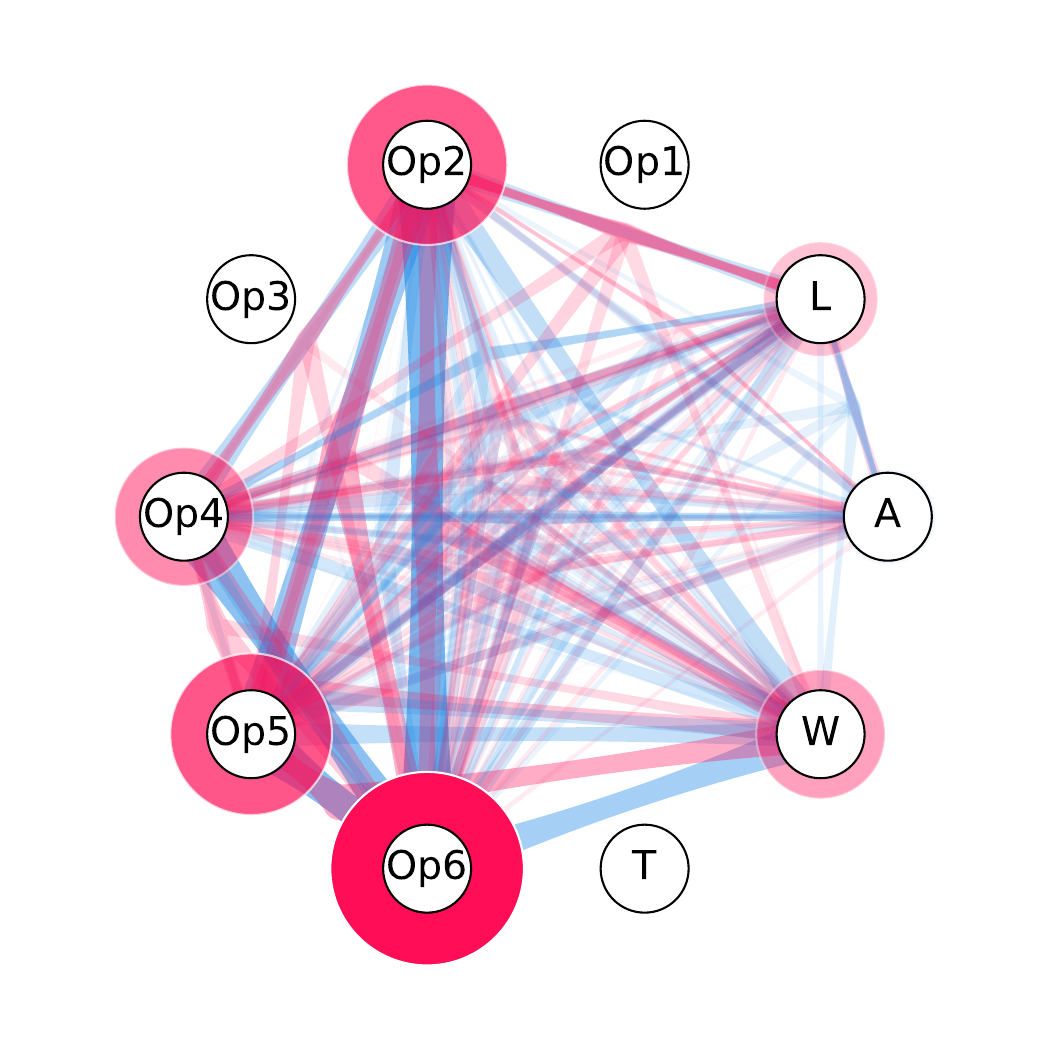}
    \end{minipage}
    \hfill
    \begin{minipage}[c]{0.40\textwidth}
        \centering
        \includegraphics[width=.9\linewidth]{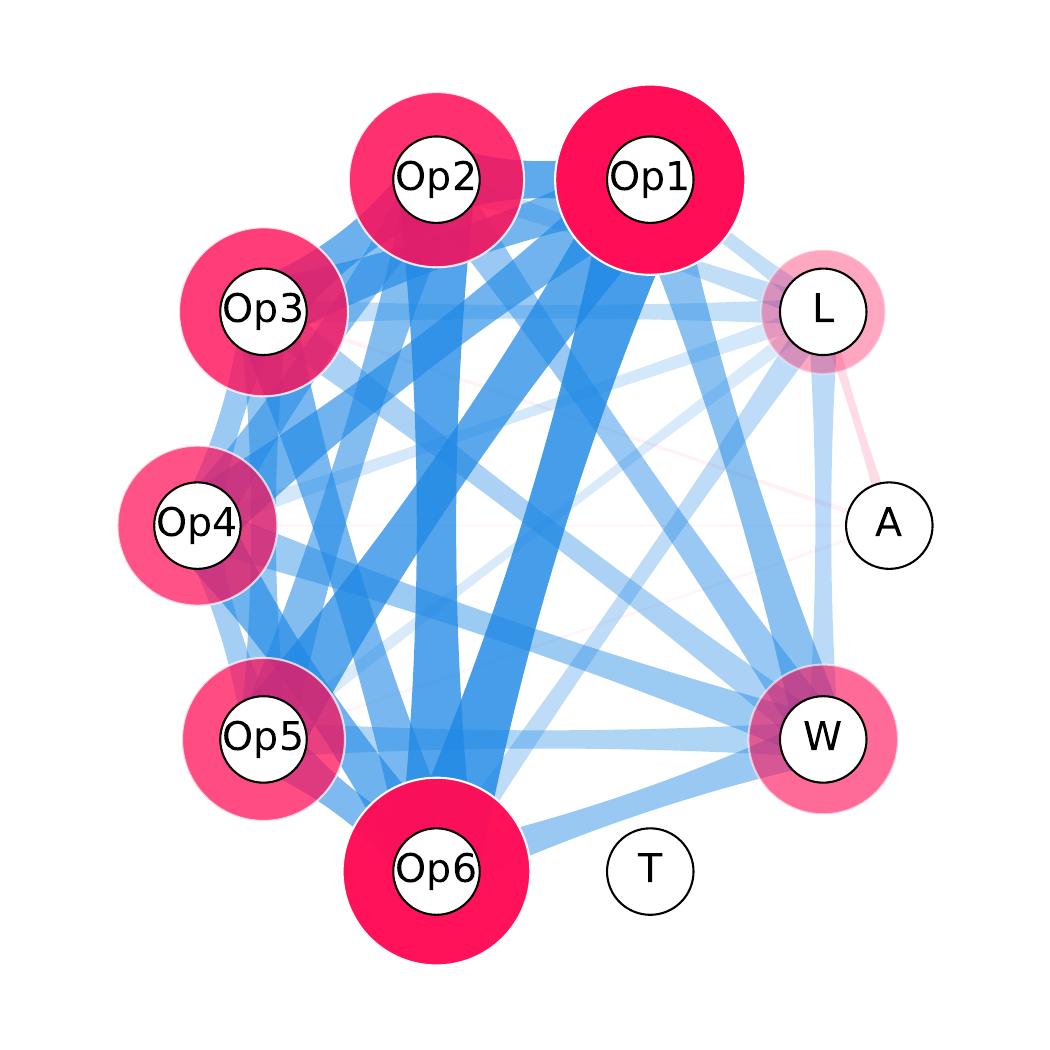}
    \end{minipage}
    \caption{\edit{MIs as computed via \tool for \texttt{CIFAR10} (top), \texttt{FashionMNIST} (middle), and \texttt{ColorectalHistology} (bottom) of \jahs.}}
    \label{fig_appx_jahs}
\end{figure}

\begin{figure}
    \begin{minipage}[c]{0.1\textwidth}
        \textbf{\gls*{SV}:}
    \end{minipage}
    \begin{minipage}[c]{0.40\textwidth}
        \centering
        \includegraphics[width=.9\linewidth]{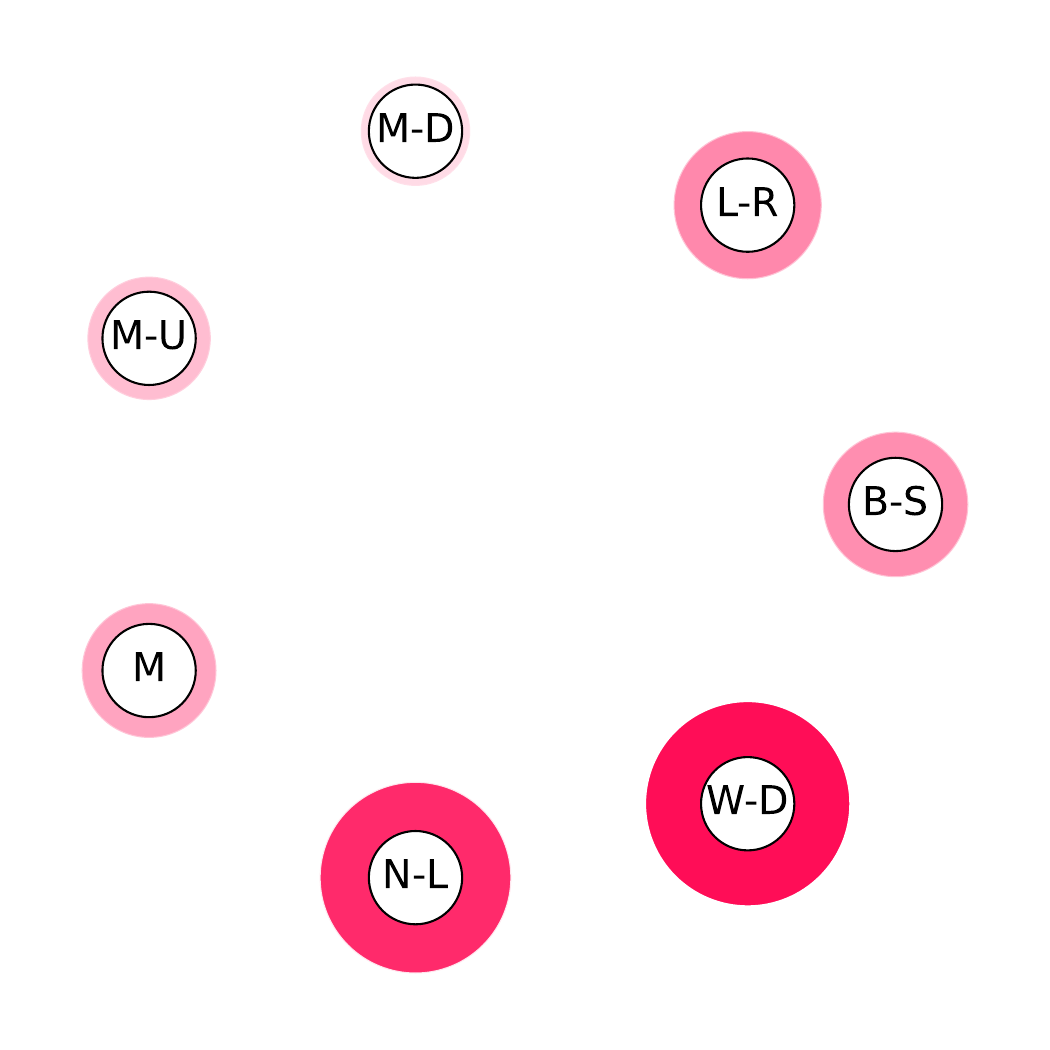}
    \end{minipage}
    \hfill
    \begin{minipage}[c]{0.40\textwidth}
        \centering
        \includegraphics[width=\linewidth]{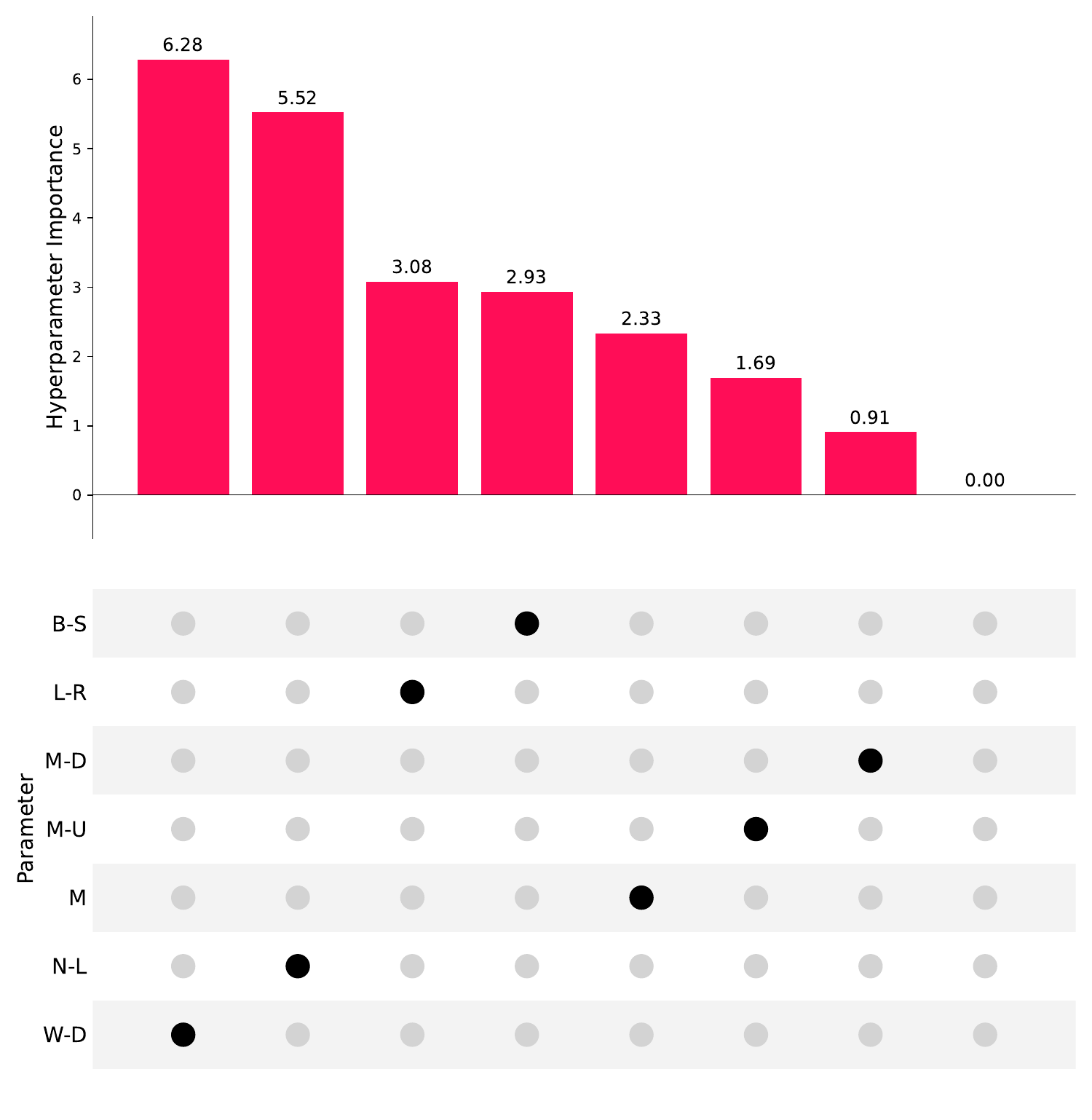}
    \end{minipage}
    \\
    \begin{minipage}[c]{0.1\textwidth}
        \textbf{\gls*{SI}:}
    \end{minipage}
    \begin{minipage}[c]{0.40\textwidth}
        \centering
        \includegraphics[width=.9\linewidth]{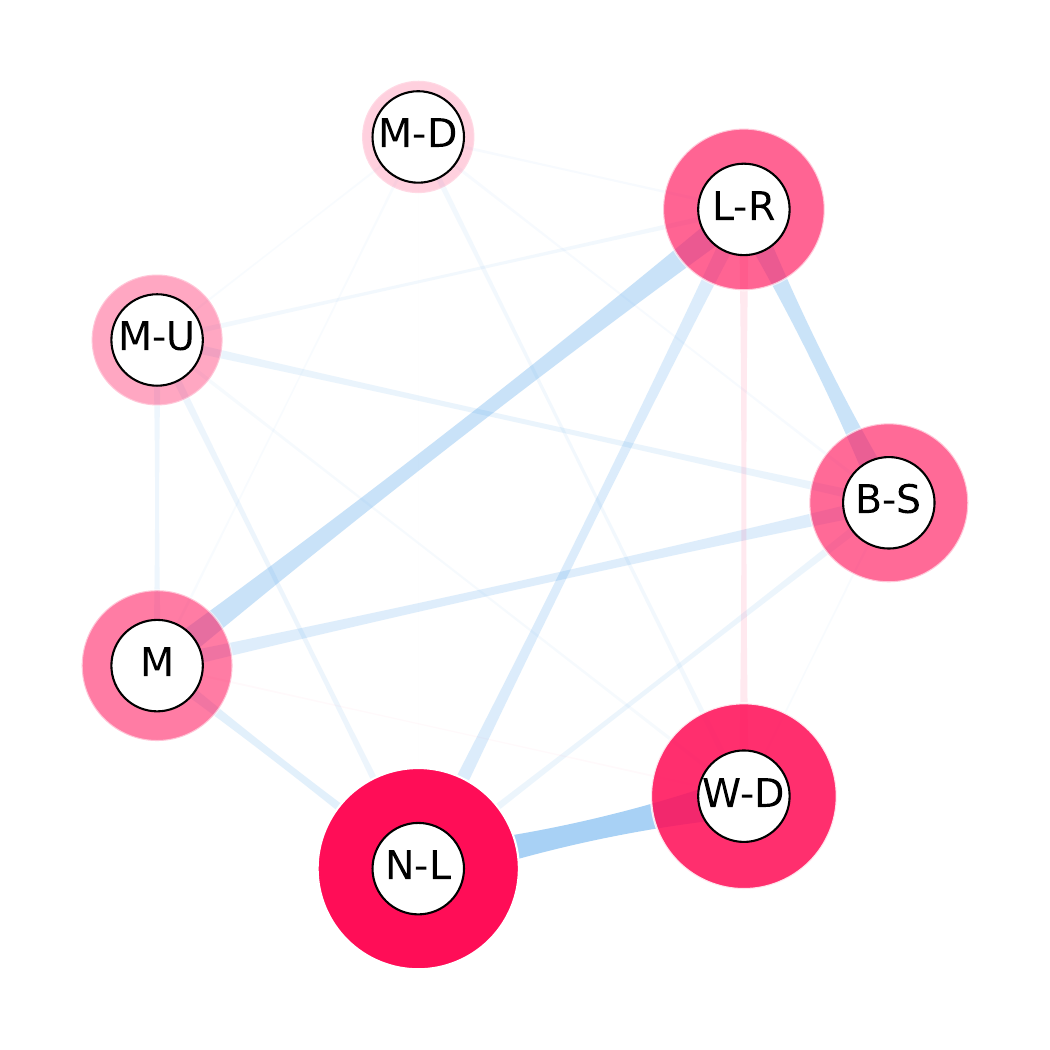}
    \end{minipage}
    \hfill
    \begin{minipage}[c]{0.40\textwidth}
        \centering
        \includegraphics[width=\linewidth]{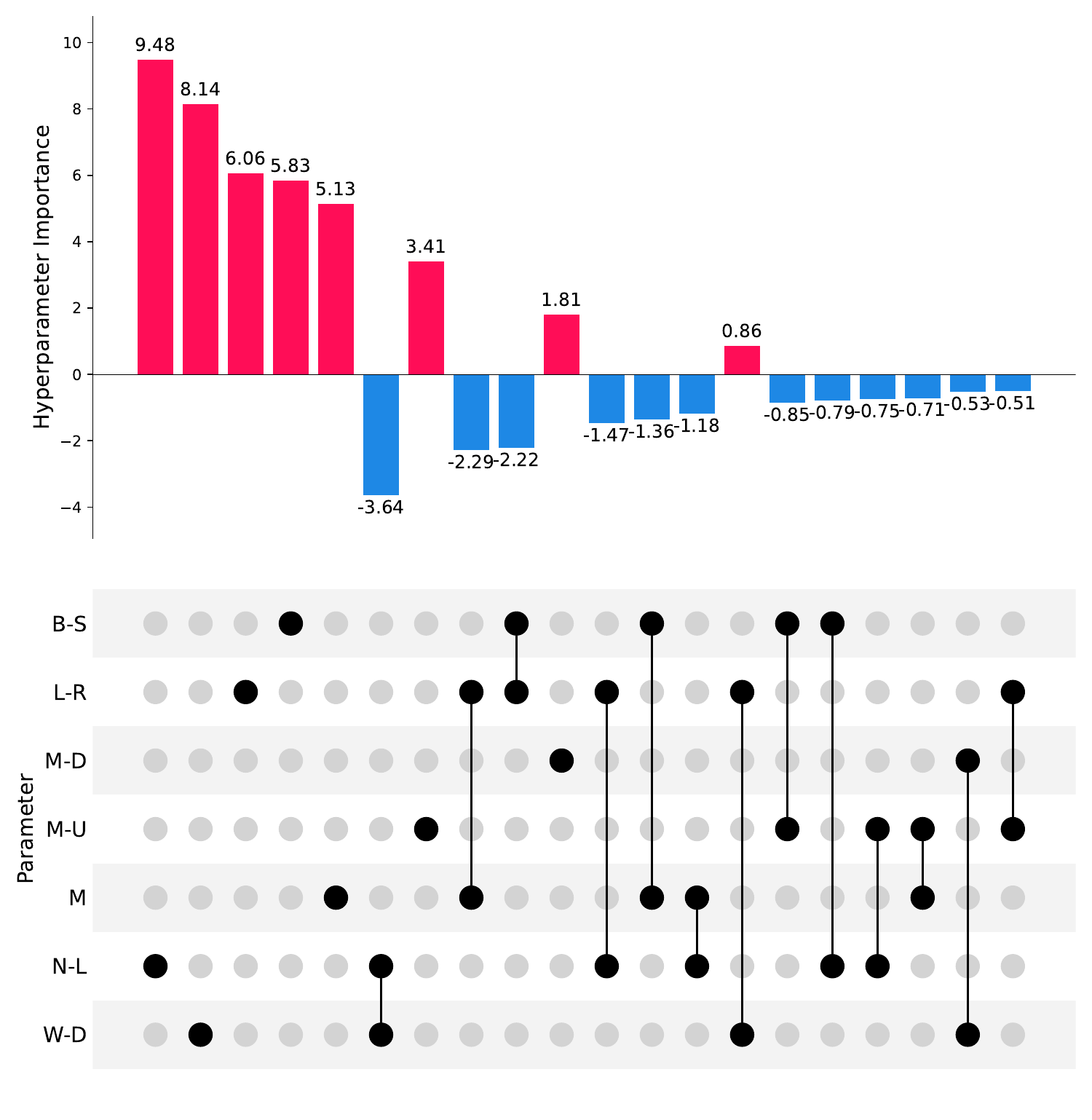}
    \end{minipage}
    \\
    \begin{minipage}[c]{0.1\textwidth}
        \textbf{\gls*{MI}:}
    \end{minipage}
    \begin{minipage}[c]{0.40\textwidth}
        \centering
        \includegraphics[width=.9\linewidth]{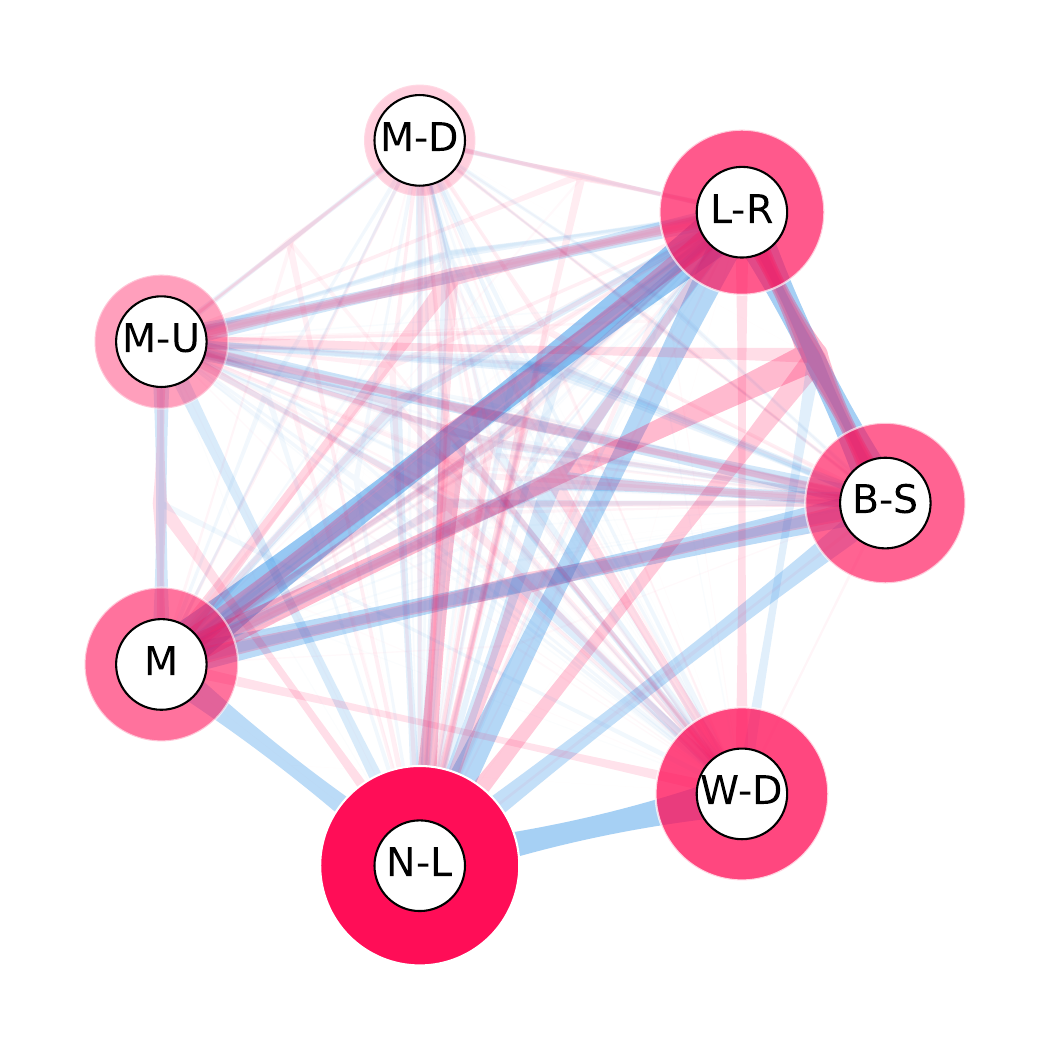}
    \end{minipage}
    \hfill
    \begin{minipage}[c]{0.40\textwidth}
        \centering
        \includegraphics[width=\linewidth]{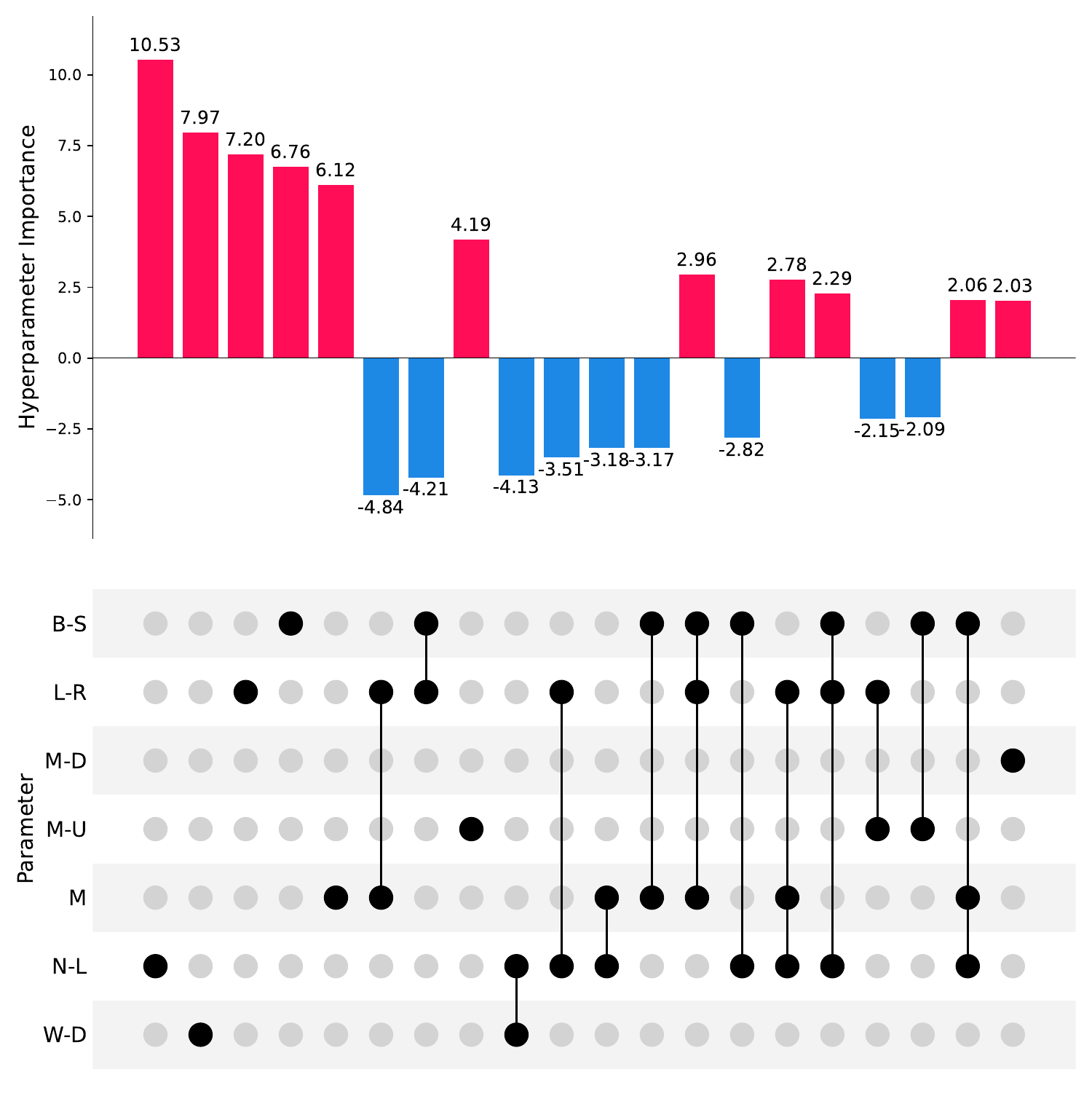}
    \end{minipage}
    \caption{\glspl*{SV}, \glspl*{SI}, and \glspl*{MI} for the \tunability setting on \lcbench. The interactivity in the full decomposition of the \glspl*{MI} is summarized into less complicated explanations by the \glspl*{SV} and \glspl*{SI}. Notably, all \glspl*{SV} are positive.}
    \label{fig_appx_lcbenc_tunability_different_indices}
\end{figure}

\begin{figure}
    \begin{minipage}[c]{0.1\textwidth}
        \textbf{\gls*{SV}:}
    \end{minipage}
    \begin{minipage}[c]{0.40\textwidth}
        \centering
        \includegraphics[width=.9\linewidth]{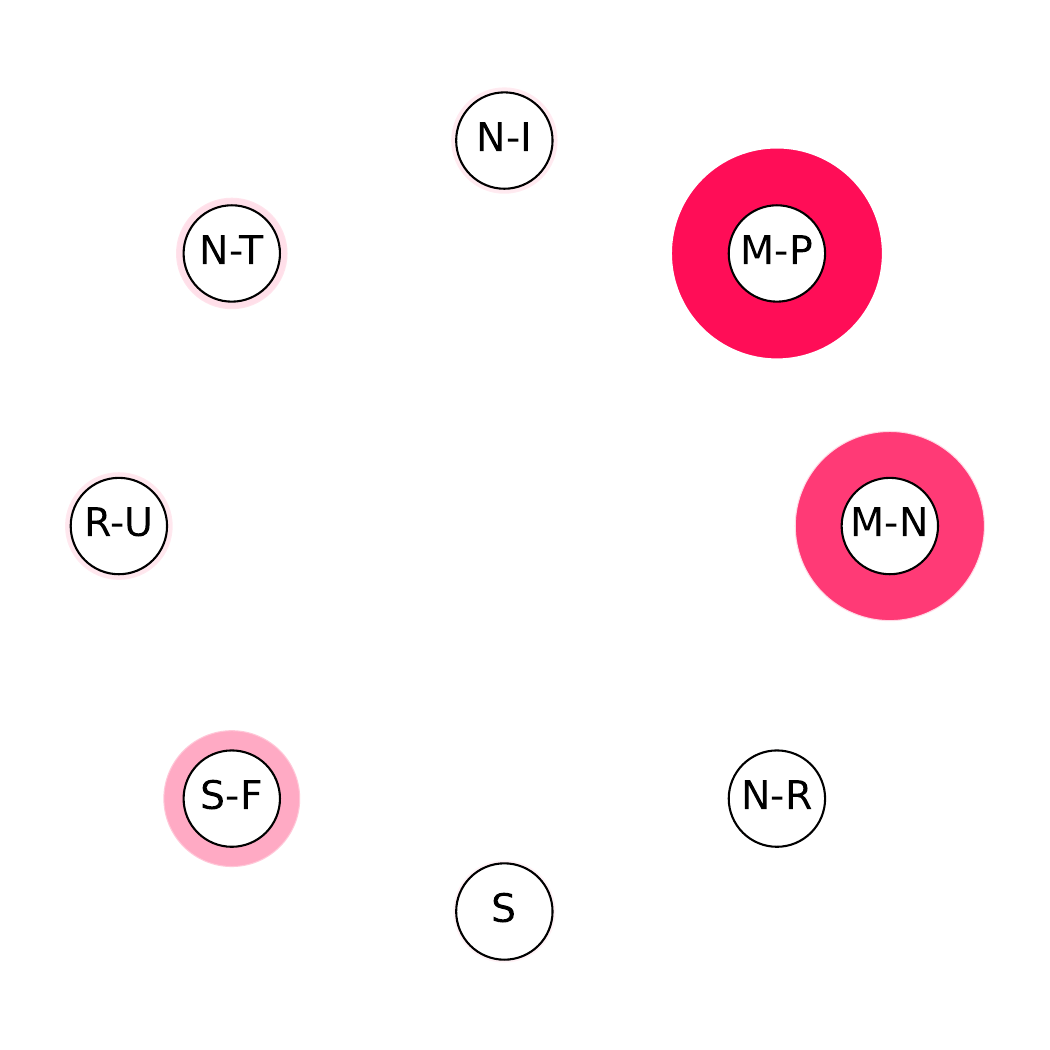}
    \end{minipage}
    \hfill
    \begin{minipage}[c]{0.40\textwidth}
        \centering
        \includegraphics[width=.9\linewidth]{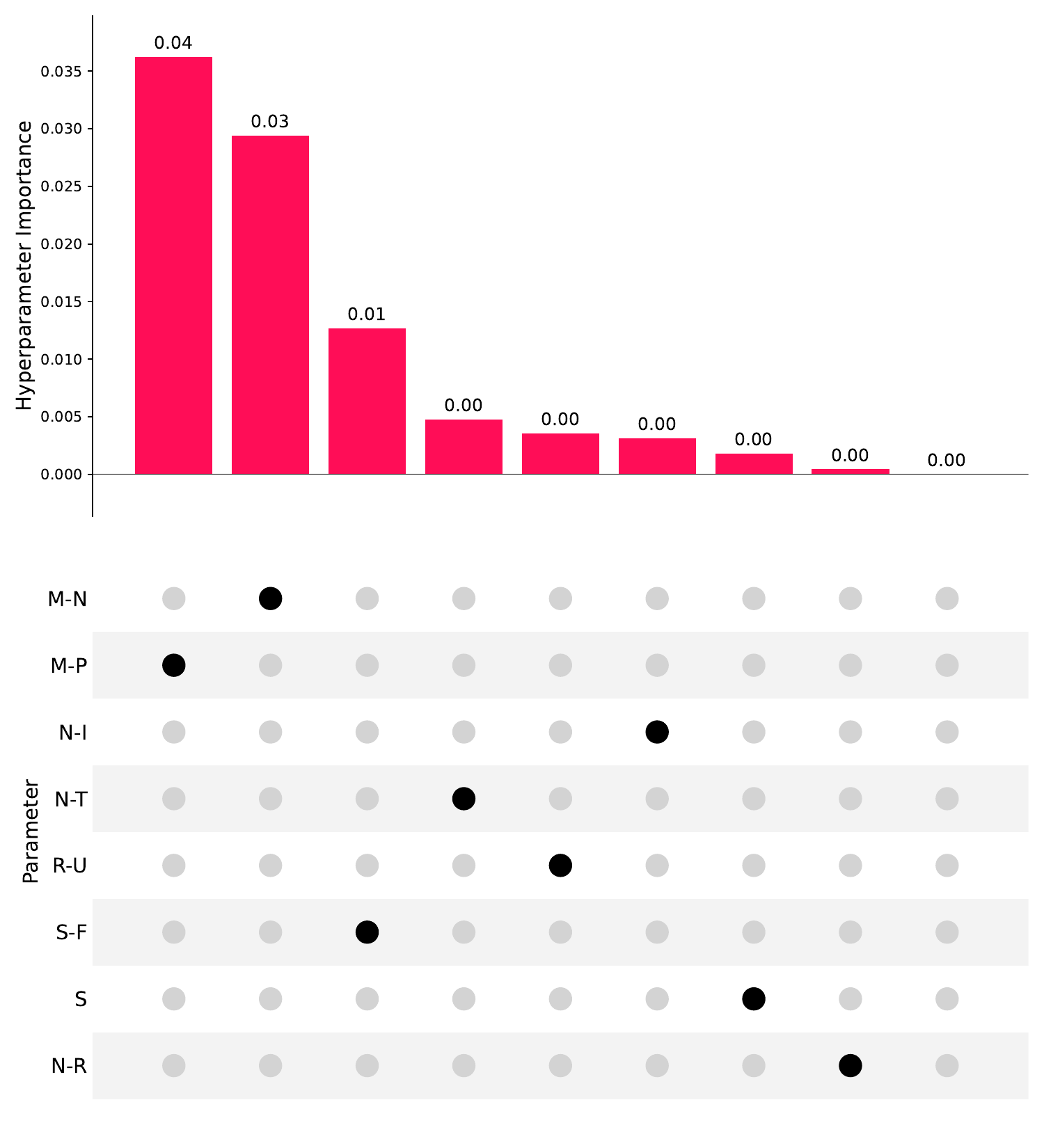}
    \end{minipage}
    \\
    \begin{minipage}[c]{0.1\textwidth}
        \textbf{\gls*{SI}:}
    \end{minipage}
    \begin{minipage}[c]{0.40\textwidth}
        \centering
        \includegraphics[width=.9\linewidth]{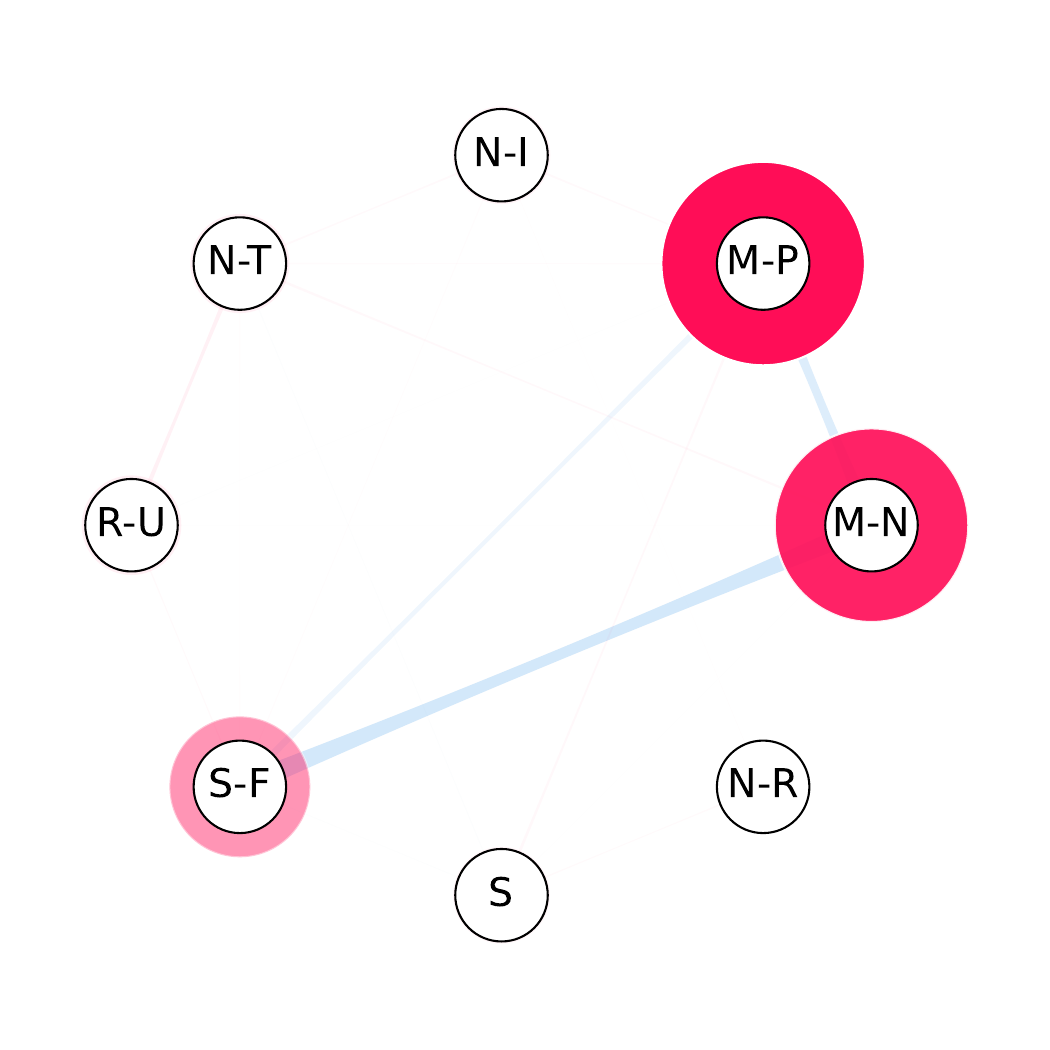}
    \end{minipage}
    \hfill
    \begin{minipage}[c]{0.40\textwidth}
        \centering
        \includegraphics[width=.9\linewidth]{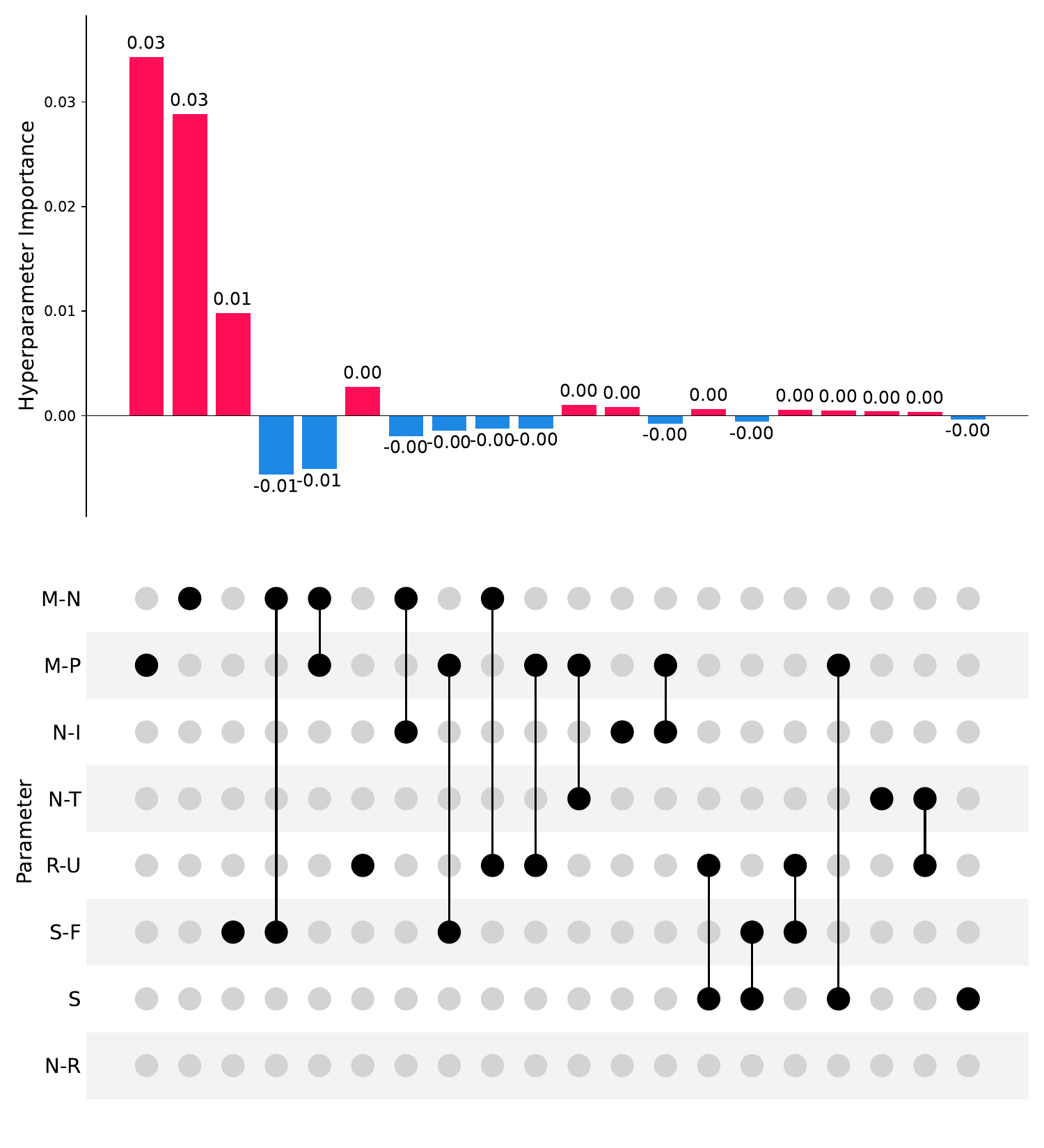}
    \end{minipage}
    \\
    \begin{minipage}[c]{0.1\textwidth}
        \textbf{\gls*{MI}:}
    \end{minipage}
    \begin{minipage}[c]{0.40\textwidth}
        \centering
        \includegraphics[width=.9\linewidth]{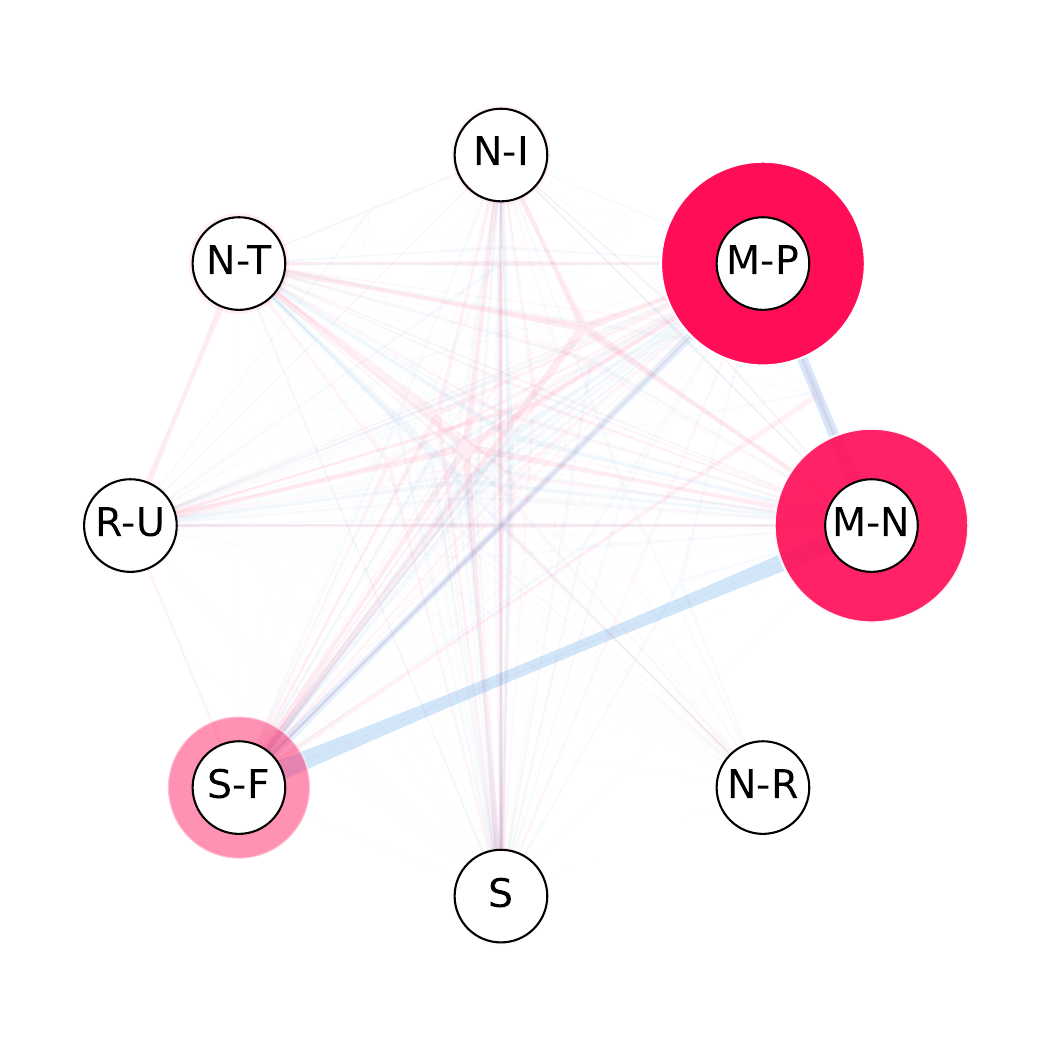}
    \end{minipage}
    \hfill
    \begin{minipage}[c]{0.40\textwidth}
        \centering
        \includegraphics[width=.9\linewidth]{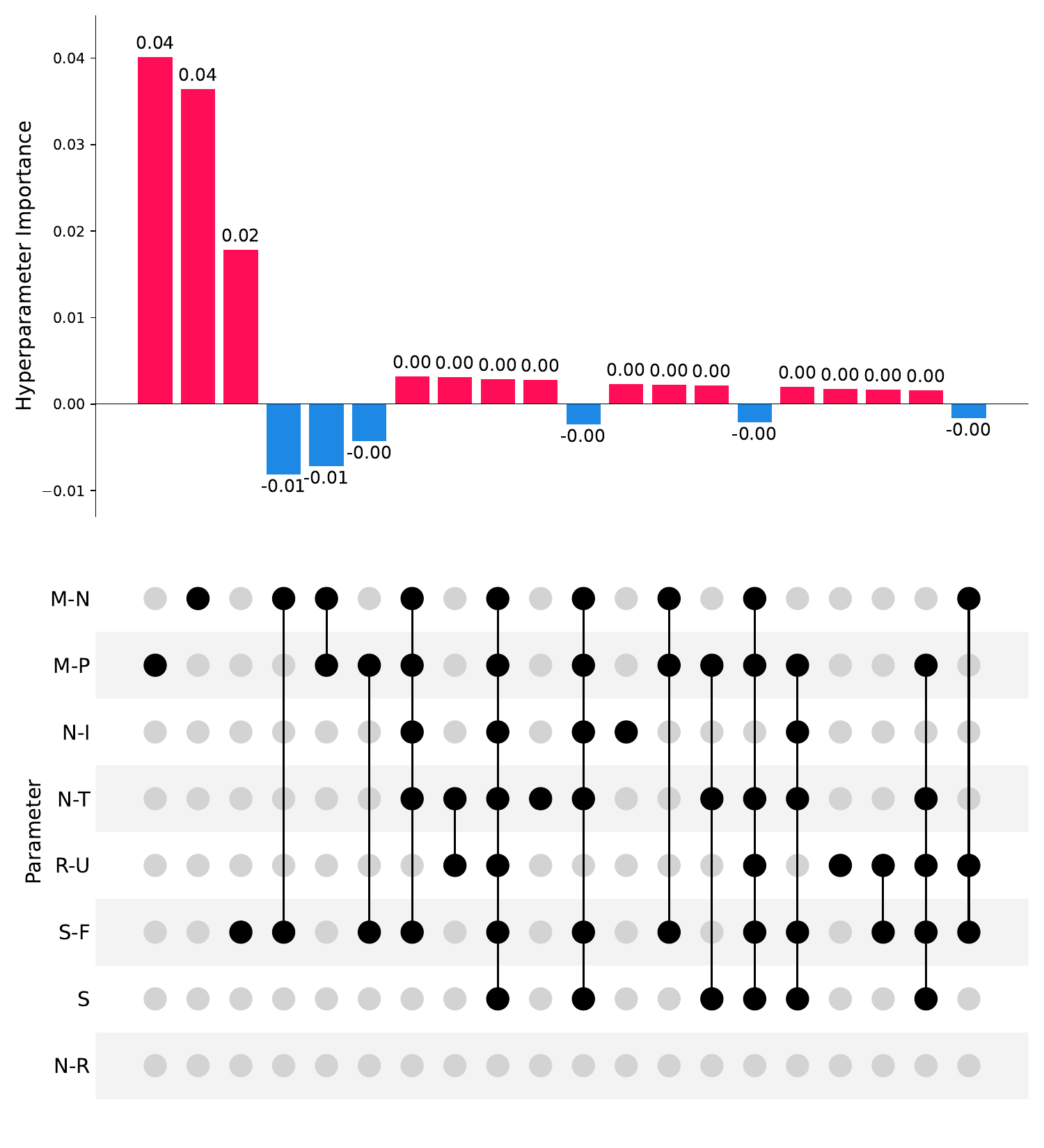}
    \end{minipage}
    \caption{\glspl*{SV}, \glspl*{SI}, and \glspl*{MI} for the \tunability setting on \rbvranger. The interactivity in the full decomposition of the \glspl*{MI} is summarized into less complicated explanations by the \glspl*{SV} and \glspl*{SI}. Notably, all \glspl*{SV} are positive.}
    \label{fig_appx_ranger_tunability_different_indices}
\end{figure}

In \cref{fig_appx_pd1,fig_appx_jahs,fig_appx_lcbenc_tunability_different_indices,fig_appx_ranger_tunability_different_indices}, we show more interaction graphs for the different benchmarks we evaluated \tool on. This includes \pdone (cf. \cref{fig_appx_pd1}, \texttt{JAHS-Bench-201} (cf. \cref{fig_appx_jahs}), \texttt{lcbench} (cf. \cref{fig_appx_lcbenc_tunability_different_indices}), and \texttt{rbvs\_ranger} (cf. \cref{fig_appx_ranger_tunability_different_indices}). We find that with \tool we can elicit interesting interaction structures for the tuning of transformers and neural architectures in more general. Surprisingly, there can be comparably low interaction between hyperparameters steering the learning behavior and hyperparameters controlling the neural architecture, as seen for CIFAR10. However, for the other two datasets, the higher degree of interaction between the learner's hyperparameters and those of the architecture better meets intuition and expectation.

\begin{figure}
    \begin{minipage}[c]{0.1\textwidth}\phantom{\textbf{SV:}}\end{minipage}
    \begin{minipage}[c]{0.40\textwidth}\centering\textbf{\dstunability}\end{minipage}
    \hfill
    \begin{minipage}[c]{0.40\textwidth}\centering\textbf{Sensitivity}\end{minipage}
    \\
    \begin{minipage}[c]{0.1\textwidth}
        \textbf{\gls*{MI}:}
    \end{minipage}
    \begin{minipage}[c]{0.40\textwidth}
        \centering
        \includegraphics[width=.9\linewidth]{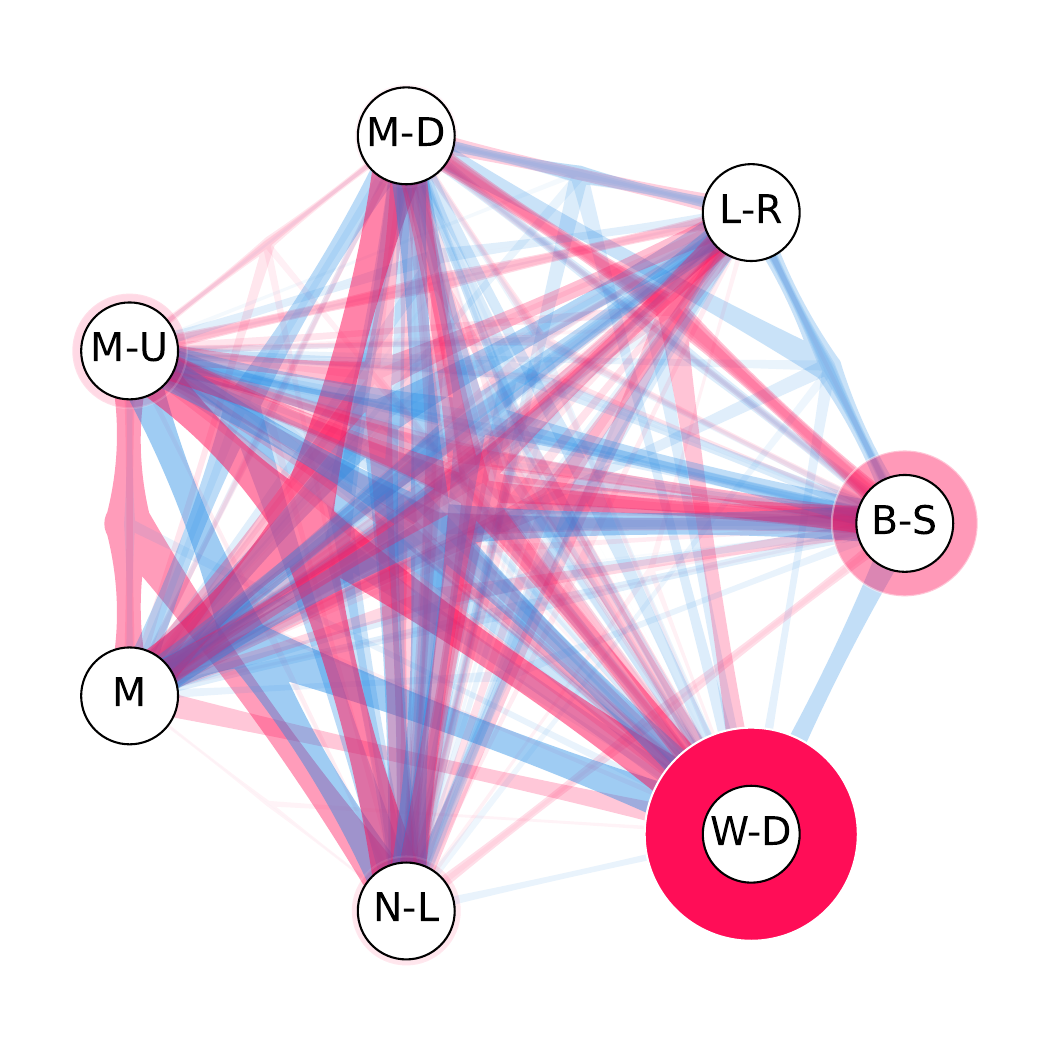}
    \end{minipage}
    \hfill
    \begin{minipage}[c]{0.40\textwidth}
        \centering
        \includegraphics[width=.9\linewidth]{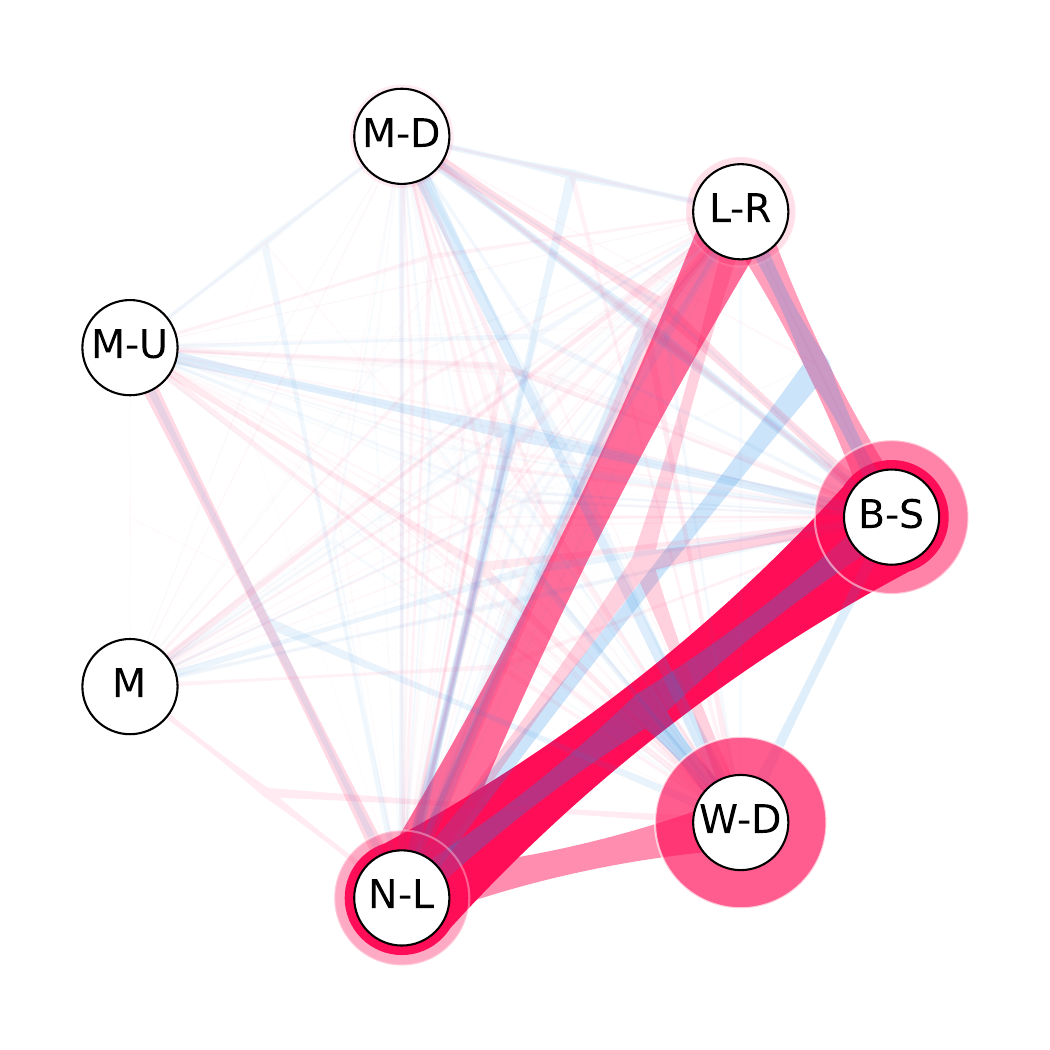}
    \end{minipage}
    \\
    \begin{minipage}[c]{0.1\textwidth}
        \textbf{\gls*{SI}:}
    \end{minipage}
    \begin{minipage}[c]{0.40\textwidth}
        \centering
        \includegraphics[width=.9\linewidth]{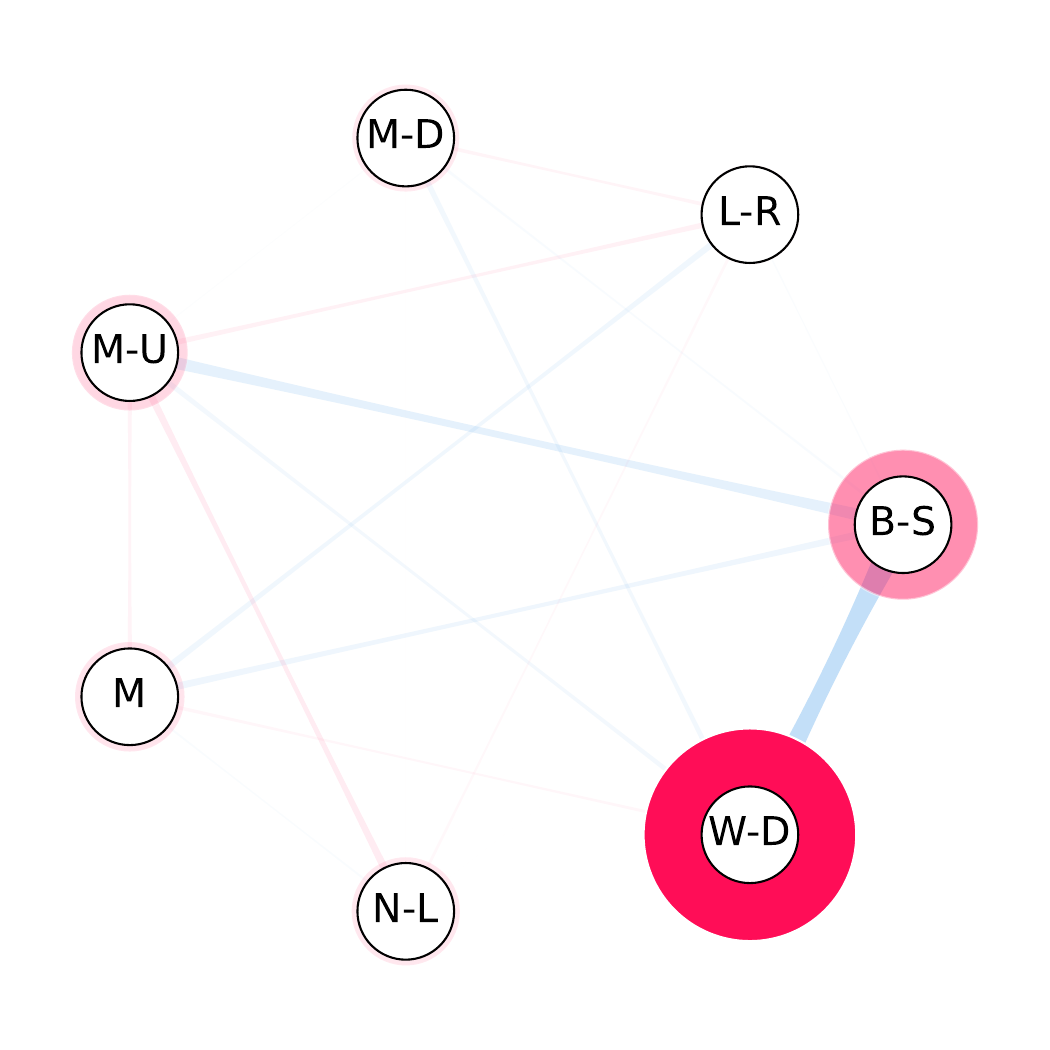}
    \end{minipage}
    \hfill
    \begin{minipage}[c]{0.40\textwidth}
        \centering
        \includegraphics[width=.9\linewidth]{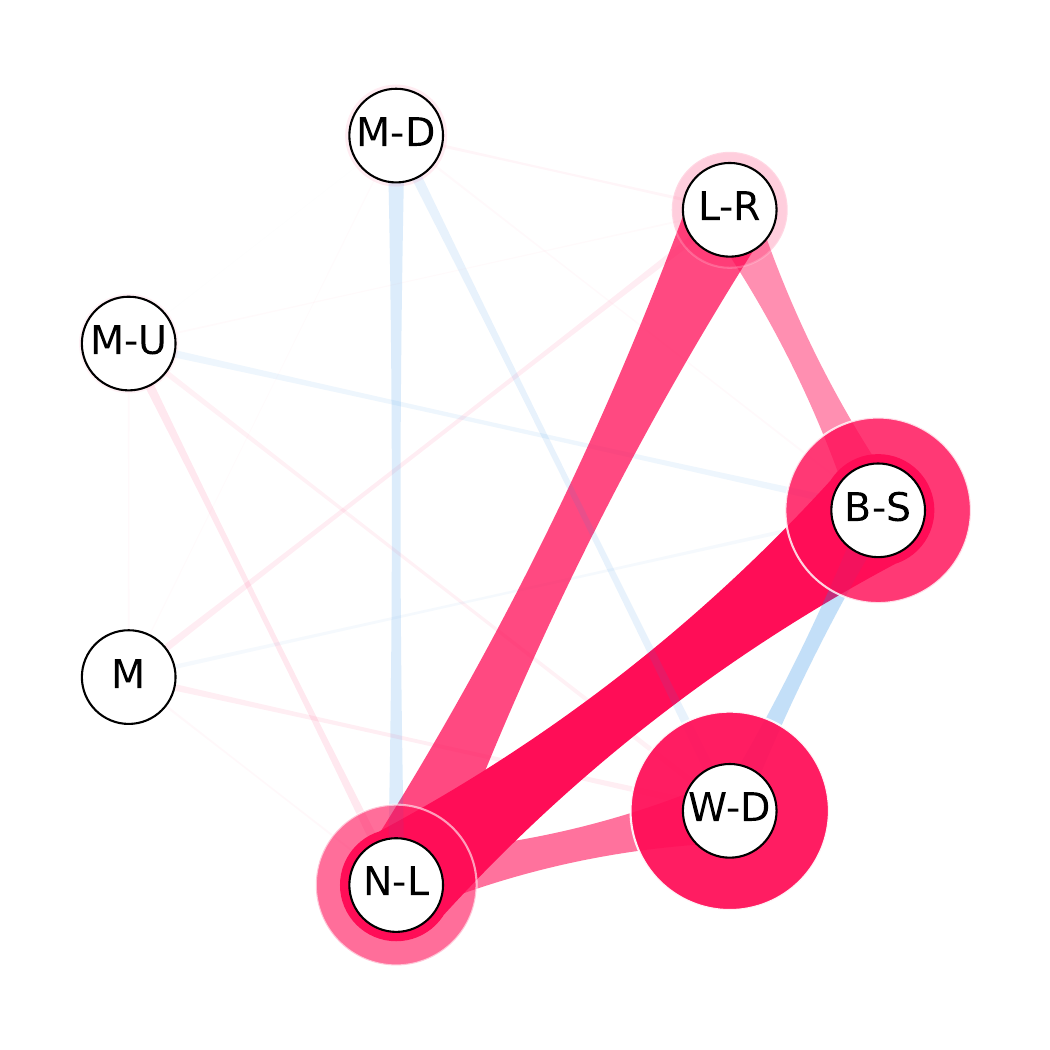}
    \end{minipage}
    \\
    \begin{minipage}[c]{0.1\textwidth}
        \textbf{\gls*{SV}:}
    \end{minipage}
    \begin{minipage}[c]{0.40\textwidth}
        \centering
        \includegraphics[width=.9\linewidth]{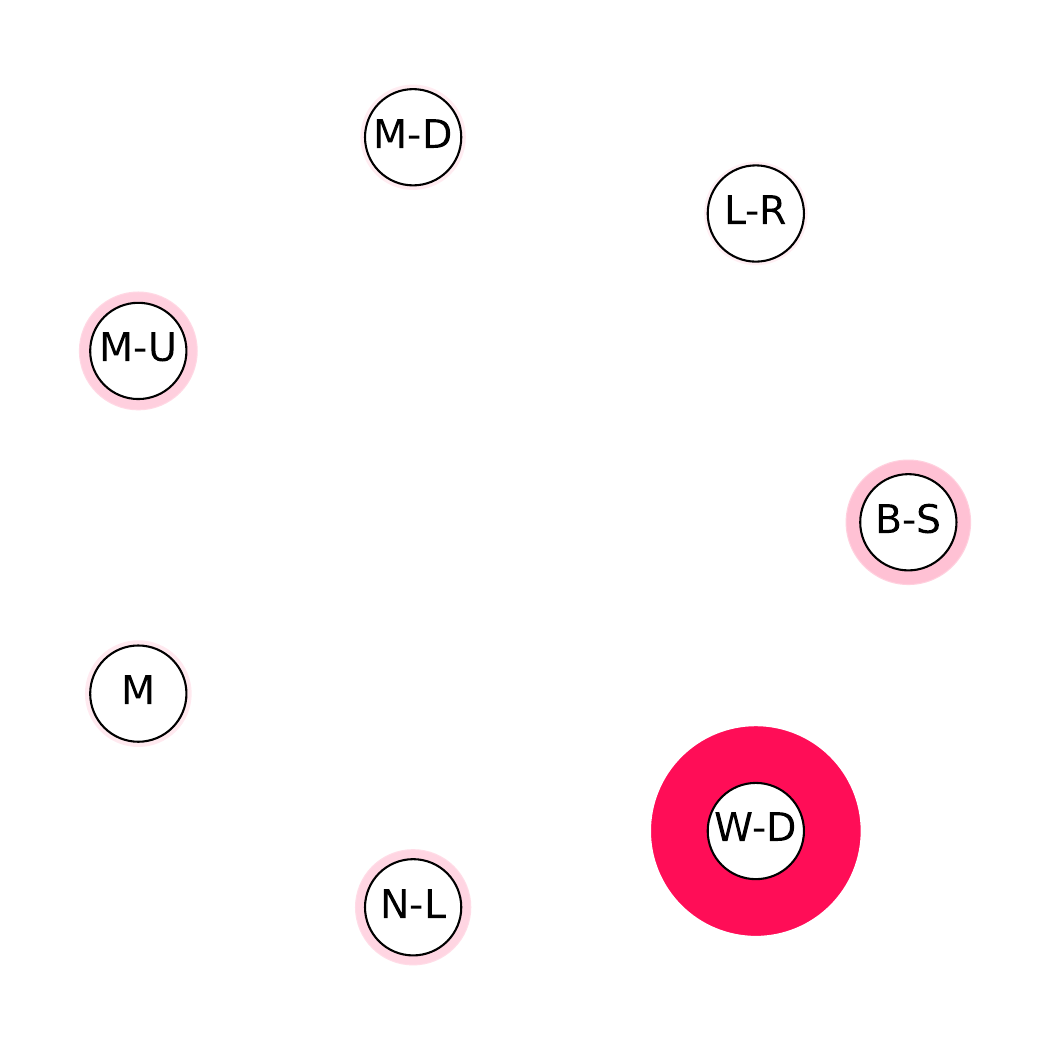}
    \end{minipage}
    \hfill
    \begin{minipage}[c]{0.40\textwidth}
        \centering
        \includegraphics[width=.9\linewidth]{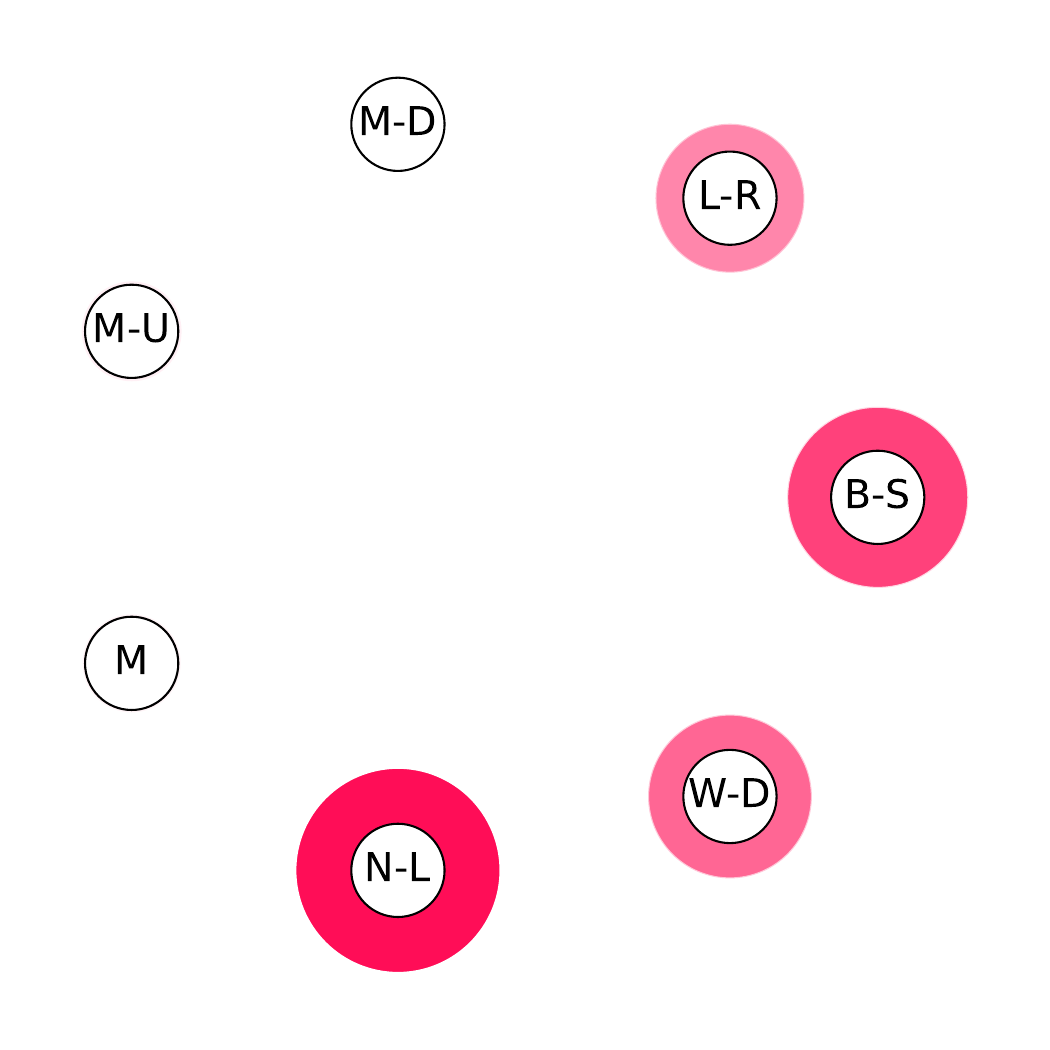}
    \end{minipage}
    \caption{Comparison of \dstunability (left) and Sensitivity (right) games as provided via \tool. Both variants of measuring HPI provide notably different explanations. Note that also for Sensitivity, \tool can be used to compute lower-order explanations summarizing higher-order interactions accordingly.}
    \label{fig_sensitivity_tunability}
\end{figure}

In \cref{fig_sensitivity_tunability}, we compare the \sensitivity to the \dstunability game for dataset ID 7593 of \texttt{lcbench} on three different levels: Moebius interactions showing all pure effects, Shapley interactions, summarizing higher-order interactions to main effects and interactions of order two and Shapley values representing the entire game solely in terms of main effects. What we can observe is that \dstunability and \sensitivity yield quite different explanations as \sensitivity does not blend an optimized hyperparameter configuration with the default hyperparameter configuration for evaluating the value function for a given coalition but takes the variance. Taking the variance apparently results in more pronounced interactivity structures as the performance is no longer contrasted to the default configuration.

\FloatBarrier

\section*{Ethical Statement}
We believe that \tool can support practitioners in understanding the role of hyperparameters in model performance, quantify tunability, making model-based optimization more transparent, and uncover biases in optimizers. This is particularly valuable in resource-constrained settings, where informed decisions about which hyperparameters to tune can render HPO more efficient.

\end{document}

%% file: settings.tex
\usepackage{url}
\usepackage{amsmath,amsfonts,amssymb,amsthm}
\usepackage{graphicx}

\usepackage{xspace}

\usepackage[labelfont=bf]{caption}
\usepackage{enumitem}
\usepackage{multirow}
\usepackage{placeins}
\usepackage{rotating}
\usepackage{stackengine}
\usepackage{tabularx}
\usepackage{titletoc} 
\usepackage{booktabs}

\theoremstyle{plain}
\newtheorem{theorem}{Theorem}
\newtheorem{proposition}{Proposition}

\theoremstyle{definition}
\newtheorem{definition}{Definition}

\theoremstyle{remark}

\usepackage[acronym]{glossaries}
\newacronym{SV}{SV}{Shapley Value}
\newacronym{SII}{SII}{Shapley Interaction Index}
\newacronym{k-SII}{$k$-SII}{$k$-Shapley Value}
\newacronym{SI}{SI}{Shapley Interaction}
\newacronym{MI}{MI}{Möbius Interaction}
\newacronym{STII}{STII}{Shapley Taylor Interaction Index}
\newacronym{FSII}{FSII}{Faithful Shapley Interaction Index}

\newcommand{\cX}{\mathcal{X}}
\newcommand{\cY}{\mathcal{Y}}

\renewcommand{\vec}[1]{\boldsymbol{#1}}
\newcommand{\cN}{\mathcal{N}}

\input{math_commands}

\usepackage{cleveref}
\crefname{equation}{Eq.}{Eqs.}
\crefname{figure}{Fig.}{Figs.}
\crefname{tabular}{Tab.}{Tabs.}
\crefname{section}{Sec.}{Secs.}
\crefname{appendix}{Apx.}{Apxs.}
\crefname{table}{Tab.}{Tabs.}
\crefname{theorem}{Thm.}{Thms.}

%% file: math_commands.tex
\usepackage{amsmath,amsfonts,bm}

\def\eqref#1{equation~\ref{#1}}

\def\1{\bm{1}}

\DeclareMathAlphabet{\mathsfit}{\encodingdefault}{\sfdefault}{m}{sl}
\SetMathAlphabet{\mathsfit}{bold}{\encodingdefault}{\sfdefault}{bx}{n}

\newcommand{\R}{\mathbb{R}}

\DeclareMathOperator*{\argmin}{arg\,min}

%% file: project_commands.tex
\newcommand{\edit}[1]{#1}

\newcommand{\numgames}{5\xspace}
\newcommand{\tool}{\mbox{{\sc HyperSHAP}}\xspace}
\newcommand{\ablation}{Ablation\xspace}
\newcommand{\sensitivity}{Sensitivity\xspace}
\newcommand{\tunability}{Multi-Data Tunability\xspace}
\newcommand{\dstunability}{Tunability\xspace}
\newcommand{\opthabit}{Optimizer Bias\xspace}
\newcommand{\jahs}{\texttt{JAHS-Bench-201}\xspace}
\newcommand{\rbvranger}{\texttt{rbv2\_ranger}\xspace}
\newcommand{\pdone}{\texttt{PD1}\xspace}
\newcommand{\lcbench}{\texttt{lcbench}\xspace}

\newcommand{\perf}{u}
\newcommand{\conf}{\vec{\lambda}}
\newcommand{\confs}{\vec{\Lambda}}
\newcommand{\val}{\textsc{Val}_\perf}